\documentclass{article}

    \usepackage[final, nonatbib]{neurips_2021}

\usepackage[utf8]{inputenc} % allow utf-8 input
\usepackage[T1]{fontenc}    % use 8-bit T1 fonts
\usepackage{hyperref}       % hyperlinks
\usepackage{url}            % simple URL typesetting
\usepackage{booktabs}       % professional-quality tables
\usepackage{amsfonts}       % blackboard math symbols
\usepackage{nicefrac}       % compact symbols for 1/2, etc.
\usepackage{microtype}      % microtypography
\usepackage{xcolor}         % colors
\usepackage{colortbl}

\usepackage{amsmath,amsfonts,bm}

\def\eqref#1{equation~\ref{#1}}
\def\1{\bm{1}}

\DeclareMathAlphabet{\mathsfit}{\encodingdefault}{\sfdefault}{m}{sl}
\SetMathAlphabet{\mathsfit}{bold}{\encodingdefault}{\sfdefault}{bx}{n}

\newcommand{\R}{\mathbb{R}}

\DeclareMathOperator*{\argmax}{arg\,max}

\usepackage{times}
\usepackage{latexsym}
\usepackage[pdftex]{graphicx}
\usepackage{bmpsize}
\usepackage{float} 
\usepackage{multirow}
\usepackage{graphics}
\usepackage{amsmath}
\usepackage{multicol}
\usepackage{lipsum}
\usepackage{mwe}
\usepackage{subfig}
\usepackage{algorithm,algorithmic}
\usepackage{wrapfig}

\newcommand\numberthis{\addtocounter{equation}{1}\tag{\theequation}}

\pdfoutput=1

\colorlet{lred}{red!15}
\colorlet{lblue}{blue!15}
\colorlet{lgreen}{green!20}

\title{\textsc{SalKG}: Learning From Knowledge Graph Explanations for Commonsense Reasoning}

\author{
    Aaron Chan$^{\clubsuit}$, Jiashu Xu$^{\clubsuit}$, \textbf{Boyuan Long}$^{\clubsuit}$, \\ \textbf{Soumya Sanyal}$^{\clubsuit}$, \textbf{Tanishq Gupta}$^{\diamondsuit}$\thanks{Work done while TG interned remotely at USC.} \hspace{0.5mm}, \textbf{Xiang Ren}$^{\clubsuit}$\\
    $^{\clubsuit}$University of Southern California, $^{\diamondsuit}$IIT Delhi\\
    \small{\texttt{\{chanaaro, boyuanlo, jiashuxu, soumyasa, xiangren\}@usc.edu}}, \\ \small{\texttt{Tanishq.Gupta.mt617@maths.iitd.ac.in}}
}

\begin{document}
\maketitle
\begin{abstract}
Augmenting pre-trained language models with knowledge graphs (KGs) has achieved success on various commonsense reasoning tasks.
% Since not all information in KGs may be useful, KG-augmented models often use attention to focus on certain KG features when making predictions.
% and to explain which KG features influenced the prediction most.
% Attention is supervised only by the task loss, so the model is never explicitly taught when certain KG features should be used.
% To explain such KG-augmented models' predictions on a given task instance, some prior works try to identify which parts of the KG are salient (\textit{i.e.}, important) to the model. At times, the KG may contain little to no relevant information.
% Whereas attention weights only show which KG features the model already focused on, \textit{saliency explanations} can identify which KG features the model should focus on to boost performance.
However, for a given task instance, the KG, or certain parts of the KG, may not be useful.
Although KG-augmented models often use attention to focus on specific KG components, the KG is still always used, and the attention mechanism is never explicitly taught which KG components should be used. 
% However, the attention mechanism is supervised only by the task loss, so the model is never explicitly taught which specific KG features should be used.
Meanwhile, saliency methods can measure how much a KG feature (\textit{e.g.}, graph, node, path) influences the model to make the \textit{correct} prediction, thus explaining which KG features are useful.
% Also, beyond manual inspection, it is unclear how the model can be trained with KG explanations.
% Also, beyond manual inspection, it is unclear how KG explanations should be used.
% Plus, compared to saliency methods (\textit{e.g.}, gradient-based, occlusion-based), which directly measure a feature's influence on the model's prediction, attention may less accurately reflect the model's decision process. 
% Meanwhile, it is unclear how KG explanations in general can be used to further improve the model's performance.
This paper explores how saliency explanations can be used to improve KG-augmented models' performance.
First, we propose to create coarse (\textit{Is the KG useful?}) and fine (\textit{Which nodes/paths in the KG are useful?}) saliency explanations.
% Second, to establish a performance upper bound for learning from saliency explanations, we analyze oracle KG-augmented models, whose attention is directly masked with saliency explanations of  granularity.
Second, to motivate saliency-based supervision, we analyze oracle KG-augmented models which directly use saliency explanations as extra inputs for guiding their attention.
Third, we propose \textsc{SalKG}, a framework for KG-augmented models to learn from coarse and/or fine saliency explanations.
Given saliency explanations created from a task's training set, \textsc{SalKG} jointly trains the model to predict the explanations, then solve the task by attending to KG features highlighted by the predicted explanations.
On three commonsense QA benchmarks (CSQA, OBQA, CODAH) and a range of KG-augmented models, we show that \textsc{SalKG} can yield considerable performance gains --- up to 2.76\% absolute improvement on CSQA.
\footnote{Code and data are available at: \url{https://github.com/INK-USC/SalKG}.}

\end{abstract}
% !TEX root = main.tex
\vspace{-0.4cm}
\section{Introduction} 
\label{sec:intro}
\vspace{-0.2cm}

% \begin{figure}
%   \centering
%   \includegraphics{figures/qa_exp.png}
%   \fbox{\rule[-.5cm]{0cm}{4cm} \rule[-.5cm]{4cm}{0cm}}
%   \caption{Sample figure caption.}
% \end{figure}

% \begin{figure}[h]
%     \centering
%     \includegraphics[width=0.9\textwidth]{figs/qa_exp}
%     \caption{\textbf{asdf.}}
%     \label{fig:adv_test_loss}
% \end{figure}

% KG-augmented models use both text and KG inputs to solve CSR tasks like commonsense QA. Given a question and candidate answer choice, a contextualized KG ($\mathcal{G}'$) is constructed from the full KG by using concepts in the question-answer text. However, $\mathcal{G}'$ may sometimes contain irrelevant information, which should be filtered out by the model.

% \xiang{i think 1st page will benefit a lot from an illustrative figure to show "salient KG structures" (within a contextualized KG) that can be helpful for complementing LM; maybe also shows that LM by itself cannot answer correctly.}

\begin{wrapfigure}{R}{0.45\textwidth}
    \centering
\vspace{-0.3cm}
    \includegraphics[width=0.45\textwidth]{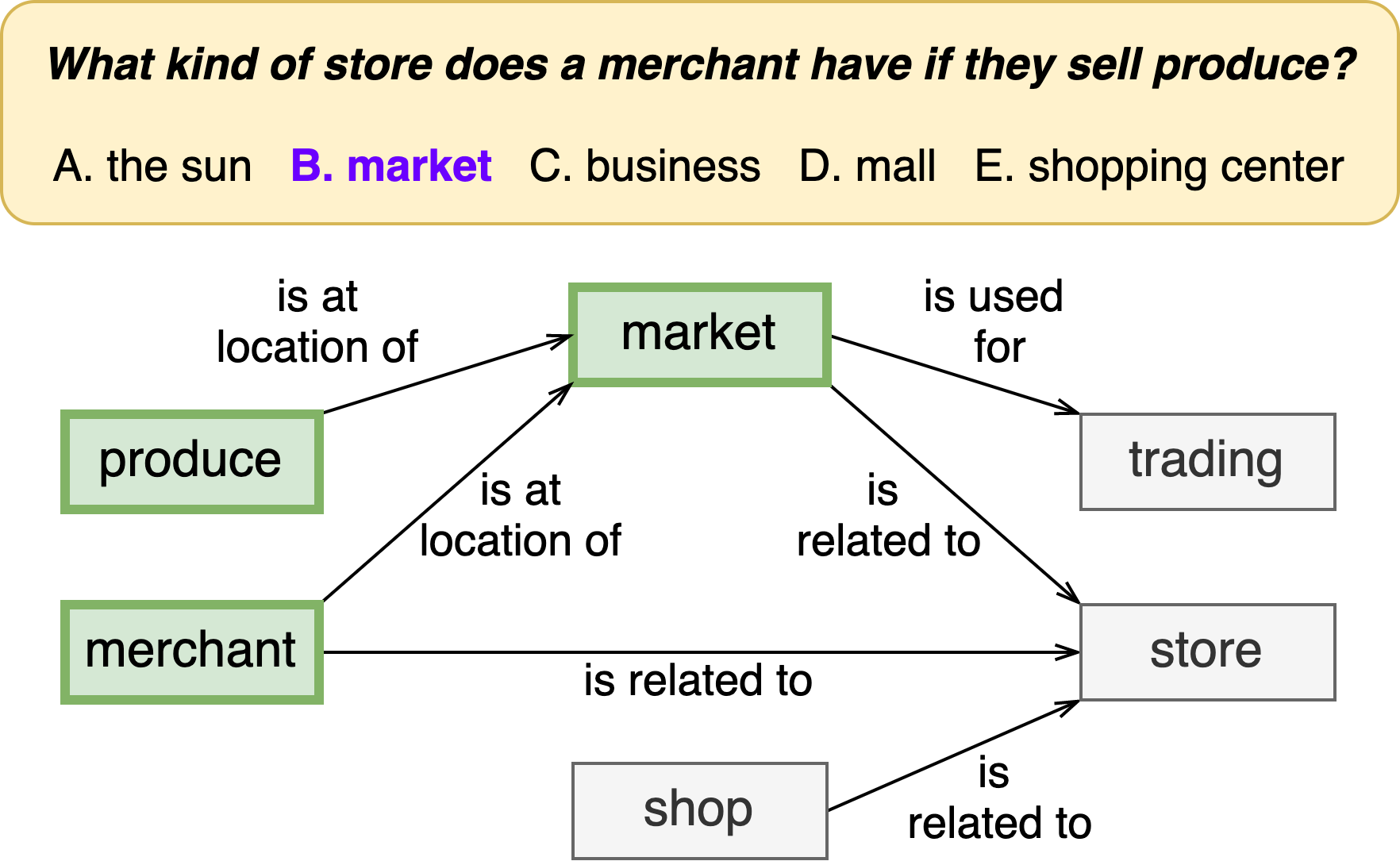}
    % \vspace{-0.6cm}
    \caption{\small \textbf{KG Saliency Explanations for Commonsense QA.} Across different questions, the KG's usefulness can vary considerably. \textit{Coarse} explanations indicate if the KG is useful overall, while \textit{fine} explanations highlight useful nodes or paths. Here, the fine explanations state that the \texttt{market}, \texttt{produce}, and \texttt{merchant} nodes are useful, while the other nodes are not.}
    \label{fig:qa}
\vspace{-0.4cm}
\end{wrapfigure}

\begin{wrapfigure}{r}{0.2\textwidth}
\vspace{-0.3cm}
    \centering
    % \vspace{-2cm}
    \includegraphics[width=0.2\textwidth]{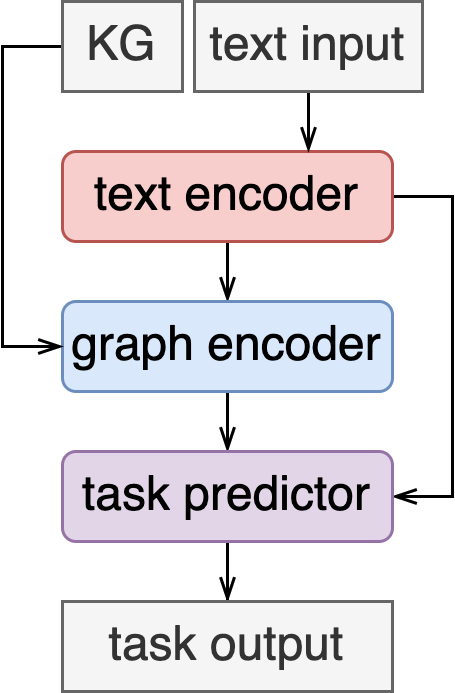}
    % \vspace{-0.6cm}
    \caption{\small \textbf{KG-Augmented Models} fuse knowledge from text and KG inputs to solve CSR tasks.}
    \label{fig:kg_model}
    \vspace{-1.2cm}
    % \vspace{-0.8cm}
\end{wrapfigure}

% Commonsense knowledge is important but not sufficiently captured in PLMs. 
Natural language processing (NLP) systems generally need common sense to function well in the real world \cite{gunning2018machine}. 
However, NLP tasks do not always provide the requisite commonsense knowledge as input. 
% Also, commonsense knowledge is seldom stated in natural language, making it hard for text encoders --- \textit{e.g.}, pre-trained language models (PLMs) \cite{devlin2018bert, liu2019roberta} --- to learn common sense from corpora alone \cite{davis2015commonsense, marcus2018deep}.
Moreover, commonsense knowledge is seldom stated in natural language, making it hard for pre-trained language models (PLMs) \cite{devlin2018bert, liu2019roberta} --- \textit{i.e.}, text encoders --- to learn common sense from corpora alone \cite{davis2015commonsense, marcus2018deep}.
% Although PLMs may capture some commonsense knowledge \cite{davison2019commonsense, petroni2019language}, it is usually insufficient for tasks that require substantial commonsense reasoning (CSR) \cite{feng2020scalable, wang2019improving}.
% KG-augmented models improve over PLMs by leveraging commonsense knowledge from KGs.
In contrast to corpora, a knowledge graph (KG) is a rich, structured source of commonsense knowledge, containing numerous facts of the form \texttt{(\begin{small}concept1, relation, concept2\end{small})}.
As a result, many methods follow the \textbf{\textit{KG-augmented model}} paradigm, which augments a text encoder with a graph encoder that reasons over the KG (Fig. \ref{fig:kg_model}).   
% \textit{KG-augmented models} consist of a PLM combined with a graph encoder that reasons over the KG.
% This interpretable predictions via multi-hop reasoning over the KG.
% and are represented in many recent commonsense reasoning methods \cite{lin2019kagnet, feng2020scalable, chen2017neural, wang2019improving}. 
%By using KG-based knowledge, 
KG-augmented models have outperformed text encoders on various commonsense reasoning (CSR) tasks, like question answering (QA) (Fig. \ref{fig:qa}) \cite{lin2019kagnet, bosselut2019dynamic, lv2020graph, yasunaga2021qa}, natural language inference (NLI) \cite{chen2017neural, wang2019improving}, and text generation \cite{liu2020kg, zhou2018commonsense}.

% KG-augmented models have outperformed text encoders on various commonsense reasoning (CSR) tasks, like question answering (QA) (Fig. \ref{fig:qa}) \cite{lin2019kagnet, bosselut2019dynamic, lv2020graph, yasunaga2021qa}, natural language inference (NLI) \cite{chen2017neural, wang2019improving, annervaz2018learning, wang2020knowledge}, and text generation \cite{liu2020kg, lin-etal-2020-commongen, lin2020differentiable}.

% \yuchen{can add KG-BART (AAAI 21) for CommonGen as an example to use KG for NLG as well.}

% \xiang{this critical paragraph feels a bit thin/weak to me now. It will be better if the discussion about why we need to explain KG-models (what are risks if not); what do we mean by explaining; why it is technically different from related work.}

% Little prior work on explaining KG-augmented models. No prior work on ~faithfully~ explaining KG-augmented models.
%
%

% Since KGs generally do not have perfect knowledge coverage, they may be useful for certain task instances but not for others.
% Plus, even if the KG is useful overall for a given instance, only some parts of the KG may be useful.
% Hence, KG-augmented models often use attention \yuchen{``attention modules/layers'' or the attention \textit{mechanism} with citation?} to focus on specific KG features during prediction.

Since KGs do not have perfect knowledge coverage, they may not contain useful knowledge for all task instances (\textit{e.g.}, if the KG in Fig. \ref{fig:qa} only consisted of the gray nodes).
% , yet existing KG-augmented models always assume the KG should be used.
Also, even if the KG is useful overall for a given task instance, only some parts of the KG may be useful (\textit{e.g.}, the green nodes in Fig. \ref{fig:qa}). 
Ideally, a KG-augmented model would know both if the KG is useful and which parts of the KG are useful.
Existing KG-augmented models always assume the KG should be used, but do often use attention~\cite{vaswani2017attention} to focus on specific KG components (\textit{e.g.}, nodes~\cite{feng2020scalable, schlichtkrull2018modeling, yan2020learning}, paths~\cite{wang2020connecting, santoro2017simple, bosselut2019dynamic}) when predicting.
Still, the attention mechanism is supervised (end-to-end) only by the task loss, so the model is never \textit{explicitly} taught which KG components should be used.
Without component-level supervision, the attention mechanism is more likely to overfit to spurious patterns. 

How can we better teach the model whether each KG feature (\textit{e.g.}, graph, node, path) is useful for solving the given task instance?
Using the task's ground truth labels, \textit{saliency methods} \cite{bastings2020elephant} can score each KG feature's influence on the model making the correct prediction.
Whereas attention weights show which KG features the model already used, saliency scores indicate which KG features the model should use.
By binarizing these scores, we are able to produce saliency explanations, which can serve as simple targets for training the model's attention mechanism.
% \xiang{somewhere later (or maybe here) should add a line to differentiate saliency vs. attention.}
For example, Fig. \ref{fig:qa} shows saliency explanations [\begin{small}\texttt{market}=1\end{small}, \begin{small}\texttt{produce}=1\end{small}, \begin{small}\texttt{trading}=0\end{small}, \begin{small}\texttt{merchant}=1\end{small}, \begin{small}\texttt{store}=0\end{small}, \begin{small}\texttt{shop}=0\end{small}], stating that \begin{small}\texttt{market}\end{small}, \begin{small}\texttt{produce}\end{small}, and \begin{small}\texttt{merchant}\end{small} are useful nodes for answering the question.  
% \yuchen{It sounds like binarizing the scores is a necessary step to produce saliency explanations, while is it true? We can still get weighted explanations w/o binarizing, right? What are the advantages of binarizing the scores ---- simplicity, etc., ?}

% \yuchen{I suggest to put the illustrate example of ``saliency expanations'' in the context of KG-augmented models here (Fig. 1) as the last part of the above paragraph. At least you could mention Fig. 1, otherwise, it is too vague to understand the concept of ``saliency explanation''.}

% Since the KG may not always contain \textit{salient} (\textit{i.e.}, relevant) information, KG-augmented models often use attention to focus on salient KG features during prediction and to explain which KG features were most salient.
% Plus, KGs naturally support explanations at different levels of granularity (\textit{e.g.}, graph, node, path).
% However, this attention mechanism is supervised only by the task loss, so the model is never explicitly taught when the KG or specific parts of the KG should be used.
% Also, beyond manual inspection, it is unclear how KG explanations should be used.
% Plus, compared to saliency methods (\textit{e.g.}, gradient-based, occlusion-based), which directly measure a feature's influence on the model's prediction, attention may less accurately reflect the model's decision process.

In this paper, we investigate how saliency explanations can be used to improve KG-augmented models' performance.
First, we propose to create \textit{coarse} (graph-level) and \textit{fine} (node-/path-level) saliency explanations.
Since KGs have features at different granularities, saliency explanations can supply a rich array of signals for learning to focus on useful KG features.
To create coarse explanations, we introduce an ensemble-based saliency method which measures the performance difference between a KG-augmented model and its corresponding non-KG-augmented model.
To create fine explanations, we can adapt any off-the-shelf saliency method, \textit{e.g.}, gradient-based \cite{denil2014extraction} or occlusion-based \cite{li2016understanding}.
Second, to demonstrate the potential of saliency-based supervision, we analyze the performance of \textit{oracle} KG-augmented models, whose attention weights are directly masked with coarse and/or fine saliency explanations.

% \xiang{make a new paragraph; smooth out the transition.}
Third, as motivated by our oracle model analysis, we propose the \textit{Learning from \textbf{Sal}iency Explanations of \textbf{KG}-Augmented Models} (\textbf{\textsc{SalKG}}) framework.
Given coarse and/or fine explanations created from thse task's training set, \textsc{SalKG} jointly trains the model to predict the explanations, then solve the task by attending to KG features highlighted in the predicted explanations.
Using saliency explanations to regularize the attention mechanism can help the model generalize better to unseen instances, especially when coarse and fine explanations are used together as complementary learning signals.
Indeed, on three standard commonsense QA benchmarks (CSQA, OBQA, CODAH) and a range of KG-augmented models, we show that \textsc{SalKG} can achieve considerable performance gains.

\vspace{-0.2cm}
\section{Preliminaries}
\label{sec:background}
\vspace{-0.2cm}

% CSR is the ability to perceive, understand, or judge things, using basic knowledge that is shared by almost all humans but rarely stated explicitly \cite{gunning2018machine}.
% % In this work, we consider NLP tasks that are designed to require CSR (e.g., QA) \cite{talmor2019commonsenseqa, mihaylov2018can}.
% To solve NLP-based CSR tasks, we use KG-augmented models (Fig. \ref{fig:qa}).
% \xiang{can compress down a bit.}

% \xiang{I feel we need one sentence to connect CSR and KG-augmented models to start this section? (e.g., why it's important to study KG-augmented models for CSR.)}
% \textcolor{red}{Xiang: I feel we need one sentence to connect CSR and KG-augmented models to start this section? (e.g., why it's important to study KG-augmented models for CSR.)}
Since KGs abundantly provide structured commonsense knowledge, KG-augmented models are often helpful for solving CSR tasks.
% KG-augmented models can be used for any KG-related CSR task, \textit{e.g.}, natural language inference \cite{wang2019improving, chen2017neural}, storytelling \cite{guan2019story, hsu2020knowledge}, and dialogue generation \cite{zhou2018commonsense}. 
CSR tasks are generally formulated as multi-choice QA (discriminative) tasks \cite{talmor2019commonsenseqa, mihaylov2018can, khot2020qasc}, but sometimes framed as open-ended response (generative) \cite{liu2020kg, lin-etal-2020-commongen} tasks.
% Many CSR tasks, \textit{e.g.}, natural language inference \cite{wang2019improving, chen2017neural}, have been formulated in the multi-choice QA setting, while some CSR tasks are also formulated as open-ended generation. 
% In this paper, we focus on multi-choice QA, which has been heavily studied in CSR \cite{sap2020commonsense, storks2019recent}.
Given that multi-choice QA has been more extensively studied, we consider CSR in terms of multi-choice QA.
Here, we present the multi-choice QA problem setting (Fig. \ref{fig:qa}) and the structure of KG-augmented models (Fig. \ref{fig:kg_model}).
% \aaron{Add running QA example and refer to Fig. 1-2 when explaining stuff}

% Second, the question-answer text ($x$) and $\mathcal{G}$ are encoded by the PLM ($f_{\text{text}}$) and graph encoder ($f_{\text{graph}}$), respectively.

% \xiang{add an illustrative figure to faciliate the introduction.}

\textbf{Problem Definition~}
Given a question $q$ and set of answer choices $A = \{a_i\}$, a multi-choice QA model aims to predict a plausibility score $\rho(q, a_i)$ for each $(q, a_i)$ pair, so that the predicted answer $\hat{a} = \argmax_{a_i \in A} \hspace{0.5mm} \rho(q, a_i)$ matches the target answer $a^*$.
Let $q \oplus a_i$ be the text statement formed from $(q, a_i)$, where $\oplus$ denotes concatenation.
For example, in Fig. \ref{fig:qa}, the text statement for $q \oplus a^*$ would be: \textit{What kind of store does a merchant have if they sell produce? market}.
We abbreviate $q \oplus a_i$ as $x_i$ and its plausibility score as $\rho(x_i)$.

\textbf{KG-Augmented Models~}
KG-augmented models use additional supervision from knowledge graphs to solve the multi-choice QA task. They encode the text and KG inputs individually as embeddings, then fuse the two embeddings together to use for prediction.
A KG is denoted as $\tilde{\mathcal{G}}=(\tilde{\mathcal{V}}, \tilde{\mathcal{R}}, \tilde{\mathcal{E}})$, where $\tilde{\mathcal{V}}$, $\tilde{\mathcal{R}}$, and $\tilde{\mathcal{E}}$ are the KG's nodes (concepts), relations, and edges (facts), respectively. 
An \textbf{\textit{edge}} is a directed triple of the form $e = (c_1,r,c_2) \in \tilde{\mathcal{E}}$, in which $c_1, c_2 \in \tilde{\mathcal{V}}$ are \textbf{\textit{nodes}}, and $r \in \tilde{\mathcal{R}}$ is the \textbf{\textit{relation}} between $c_1$ and $c_2$. 
A \textbf{\textit{path}} is a connected sequence of edges in the KG. 
% \xiang{1) discuss and cite instances of CSR tasks; 2) help explain terminologies via the example in Fig. 1}
When answering a question, the model does not use the entire KG, since most information in $\tilde{\mathcal{G}}$ is irrelevant to $x_i$. Instead, the model uses a smaller, \textbf{\textit{contextualized KG}} $\mathcal{G}_i =(\mathcal{V}_i,\mathcal{R}_i,\mathcal{E}_i)$, which is built from $\tilde{\mathcal{G}}$ using $x_i$.
$\mathcal{G}_i$ can be constructed heuristically by extracting edges from $\tilde{\mathcal{G}}$ \cite{lin2019kagnet, ma2019towards}, generating edges with a PLM \cite{bosselut2019dynamic}, or both \cite{wang2020connecting, yan2020learning}.
% Usually, $\mathcal{G}$ is a subgraph heuristically extracted from $\mathcal{G}$ by $\mathcal{V} \subseteq \mathcal{V}$ as the concepts mentioned in $x$, $\mathcal{R} \subseteq \mathcal{R}$ as the relations between concepts in $\mathcal{V}$, and $\mathcal{E} \subseteq \mathcal{E}$ as the edges involving $\mathcal{V}$ and $\mathcal{R}$ \cite{lin2019kagnet, feng2020scalable}.
% If $\mathcal{G}$ does not provide enough relevant information to extract a good $\mathcal{G}$, then new edges are sometimes added to $\mathcal{G}$ using a PLM-based generator \cite{bosselut2019comet, wang2020connecting, yan2020learning}.
In this paper, we consider KG-augmented models where $\mathcal{G}_i$ is built by heuristically by extracting edges from $\tilde{\mathcal{G}}$ (see Sec. \ref{sec:app_contextualized_kg} for more details), since most KG-augmented models follow this paradigm.
If $x_i$ and $\mathcal{G}_i$ are not discussed in the context of other answer choices, then we further simplify $x_i$'s and $\mathcal{G}_i$'s notation as $x$ and $\mathcal{G}$, respectively.
Since the model never uses the \textit{full} KG at once, we use ``KG'' to refer to $\mathcal{G}$ in the rest of the paper.

As in prior works \cite{lin2019kagnet, bosselut2019dynamic}, a KG-augmented model $\mathcal{F}_{\text{KG}}$ has three main components: \textbf{\textit{text encoder}} $f_\text{text}$, \textbf{\textit{graph encoder}} $f_\text{graph}$, and \textbf{\textit{task predictor}} $f_\text{task}$ (Fig. \ref{fig:kg_model}).
Meanwhile, its corresponding non-KG-augmented model $\mathcal{F}_{\text{No-KG}}$ has no graph encoder but has a slightly different task predictor $\bar{f}_\text{task}$ which only takes $\mathbf{x}$ as input.
In both $\mathcal{F}_{\text{KG}}$ and $\mathcal{F}_{\text{No-KG}}$, the task predictor outputs $\rho(x)$.
Let $\mathbf{x}$ and $\mathbf{g}$ be the embeddings of $x$ and $\mathcal{G}$, respectively. Then, the workflows of $\mathcal{F}_{\text{KG}}$ and $\mathcal{F}_{\text{No-KG}}$ are defined below:
% in Eq. \ref{eq:1}-\ref{eq:4}.
\begin{equation*}
% \vspace{0.02cm}
    \mathbf{x} = f_\text{text}(x); \quad \mathbf{g} = f_\text{graph}(\mathcal{G}, \mathbf{x}); \quad \mathcal{F}_{\text{KG}}(x, \mathcal{G}) = f_\text{task}(\mathbf{x} \oplus \mathbf{g}); \quad \mathcal{F}_{\text{No-KG}}(x) = \bar{f}_\text{task}(\mathbf{x}).
\end{equation*}
% \vspace{0.05cm}
% \begin{wrapfigure}{R}{0.32\textwidth}
% \vspace{-0.7cm}
% \begin{align}
%     \label{eq:1}
%     \mathbf{x} &= f_\text{text}(x) \\
%     \label{eq:2}
%     \mathbf{g} &= f_\text{graph}(\mathcal{G}, \mathbf{x}) \\
%     \label{eq:3}
%     \mathcal{F}_{\text{KG}}(x, \mathcal{G}) &= f_\text{task}(\mathbf{x} \oplus
%     \mathbf{g}) \\
%     \label{eq:4}
%     \mathcal{F}_{\text{No-KG}}(x) &= \bar{f}_\text{task}(\mathbf{x})
% \end{align}
% \vspace{-0.2cm}
% \end{wrapfigure}

Typically, $f_\text{text}$ is a PLM \cite{devlin2018bert, liu2019roberta}, $f_\text{graph}$ is a graph neural network (GNN) \cite{feng2020scalable, schlichtkrull2018modeling} or edge/path aggregation model \cite{lin2019kagnet, bosselut2019dynamic, santoro2017simple}, and $f_\text{task}$ and $\bar{f}_\text{task}$ are multilayer perceptrons (MLPs).
In general, $f_\text{graph}$ reasons over $\mathcal{G}$ by encoding either nodes or paths, then using soft attention to pool the encoded nodes/paths into $\mathbf{g}$.
Let $\mathcal{L}_{\text{task}}$ be the task loss for training $\mathcal{F}_{\text{KG}}$ and $\mathcal{F}_{\text{No-KG}}$. For multi-choice QA, $\mathcal{L}_{\text{task}}$ is cross-entropy loss, with respect to the distribution over $A$.
For brevity, when comparing different models, we may also refer to $\mathcal{F}_{\text{KG}}$ and $\mathcal{F}_{\text{No-KG}}$ as KG and No-KG, respectively.

\vspace{-0.3cm}
\section{Creating KG Saliency Explanations} 
\label{sec:explanations}
\vspace{-0.2cm}
% \xiang{will be great to have illustration figure for this section; to show the output of the two saliency over one concrete example.}

% \xiang{
% I feel we need to be more formal in this part:

% 1) concepts, notations, and formulation for all the new things we're introducing here. 

% 2) Also, we want to make it more clear that the conceptual idea is general and what we implemented in our experiments is some specific instances of them (i.e., what you have in Sec 3.1).

% 3) We do need one illustrative figures to help people understand the high-level ideas, plus one figure to depict the technical approach (i.e., Sec 3.2).
% }

\begin{wraptable}{r}{0.25\textwidth}
% \begin{table}[t]
% \vspace{-1.1cm}
\vspace{-0.5cm}
\centering
\scalebox{0.75}{
\begin{tabular}{lcc}
    \toprule  
    \textbf{Explanation Setting} & \textbf{Unit} \\
    \midrule
    {Coarse} & KG \\
    {Fine (MHGRN)} & Node \\
    {Fine (PathGen)} & Path \\
    {Fine (RN)} & Path \\
    % \midrule
    \bottomrule
\end{tabular}
}
% \vspace{-0.1cm}
\caption{\small \textbf{KG unit types} used for different explanation modes (Sec. \ref{sec:explanations}) and graph encoders (Sec. \ref{sec:oracle_eval}).
% \xiang{may add one more column with examples?}
}
\label{tab:units}
\vspace{-0.1cm}
% \end{table}
\end{wraptable}

Now, we show how to create coarse and fine saliency explanations, which tell us if the KG or certain parts of the KG are useful. 
% \yuchen{``are useful'' for a particular/given instance (instead of overall)}
These explanations can be used as extra inputs to mask oracle models' attention (Sec. \ref{sec:oracle}) or as extra supervision to regularize \textsc{SalKG} models' attention (Sec. \ref{sec:salkg}).
We first abstractly define a \textbf{\textit{unit}} as either $\mathcal{G}$ itself or a component of $\mathcal{G}$.
A unit can be a graph, node, path, \textit{etc.}, and we categorize units as \textbf{\textit{coarse}} (the entire graph $\mathcal{G}$) or \textbf{\textit{fine}} (a node or path within $\mathcal{G}$) (Table \ref{tab:units}).
%  shows the unit type corresponding to different explanation modes and KG-augmented models.
Given a model and task instance $(x, \mathcal{G})$, we define an \textbf{\textit{explanation}} as a \textit{binary} indicator of whether a unit $u$ of $\mathcal{G}$ is useful for the model's prediction on $(x, \mathcal{G})$.
If $u$ is useful, then $u$ should strongly influence the model to solve the instance correctly.
By making explanations binary, we can easily use explanations as masks or learning targets (since binary labels are easier to predict than real-valued scores) for attention weights.
% \yuchen{Again, why does it have to be \textit{binary}? Could add some words.}
% \xiang{missing a more formal introduction of what is ``useful" to the model's instance prediction.}
% Such explanations are sometimes called extractive explanations \cite{wiegreffe2020measuring}. 
% Although KG-augmented models take both text and KG inputs, in this work, we limit our analysis to explanations of KG inputs.
% While text-based saliency explanations have been studied in prior works \cite{sundararajan2017axiomatic, li2016understanding}, we only consider KG-based explanations in this paper. 
% Thus, \textit{coarse saliency} scores the importance of $\mathcal{G}$ as a whole, while \textit{fine saliency} scores the importance of individual components within the $\mathcal{G}$. 
% Also, we define a \textit{saliency unit} as a KG input being scored (\textit{e.g.}, graph, node, path).
% Using coarse or fine saliency explanations created from the training set, \textsc{SalKG} (Sec. \ref{sec:method}) trains KG-augmented models to focus on the task-relevant saliency units.

%%%

\vspace{-0.2cm}
\subsection{Coarse Saliency Explanations}
\label{sec:explanations_coarse}
\vspace{-0.2cm}
% \xiang{include citation to related work and discuss the relevance?}
% Coarse saliency units are graphs --- namely, KGs. 
% Since $\mathcal{G}$ does not have perfect knowledge coverage, it may provide mostly useful information for certain instances but mostly noisy information for others.
Since $\mathcal{G}$ may not always be useful, a KG-augmented model should ideally know when to use $\mathcal{G}$.
Here, the unit $u$ is the graph $\mathcal{G}$.
Given instance $(x, \mathcal{G})$, a coarse saliency explanation $y_\text{c}(x, \mathcal{G}) \in \{0, 1\}$ indicates if $\mathcal{G}$ helps the model solve the instance.
By default, $\mathcal{F}_{\text{KG}}$ assumes $\mathcal{G}$ is used, so we propose an ensemble-based saliency formulation for $y_\text{c}(x, \mathcal{G})$.
That is, we define $y_\text{c}(x, \mathcal{G})$ as stating if $\mathcal{F}_{\text{KG}}$ (\textit{i.e.}, uses $\mathcal{G}$) or $\mathcal{F}_{\text{No-KG}}$ (\textit{i.e.}, does not use $\mathcal{G}$) should be used to solve $(x, \mathcal{G})$.
Under this formulation, each $(x, \mathcal{G})$ has coarse units $\mathcal{G}$ and \texttt{None}, where \texttt{None} means ``$\mathcal{G}$ is not used''.  
% \yuchen{No-G is thus an empty set, right? Then, I think it's redundant to include a G in the notation. Maybe use [G, None]? 
% Also, previously, you said $[\cdot,\cdot]$ means \textit{concatenation}, but it seems that it doesn't mean that here? I guess \{G, None\} would be a better choice for describing a set. 
% }
% Recall that $\mathcal{F}_{\text{KG}}$ and $\mathcal{F}_{\text{No-KG}}$ have the same architecture, besides $\mathcal{F}_{\text{No-KG}}$ not having a graph encoder $f_\text{graph}$ but having task predictor $\tilde{f}_\text{task}$ instead of $f_\text{task}$ (Sec. \ref{sec:background}). 
% Given a task instance, coarse saliency involves deciding between using the KG (\textit{i.e.}, use KG-augmented model's prediction) or not using the KG (\textit{i.e.}, use PLM's prediction), which depends on how much relevant information the KG provides for this instance. 
% \xiang{missing an accessible explanation to lay readers about the usefulness/motivation of such explanation --- maybe a small subsection for this?}

% \xiang{Should first briefly introduce the high-level idea of attribution/saliency computation (e.g., given different features); and then say how you extend that to our setting.}
To get $y_\text{c}(x, \mathcal{G})$, we begin by computing coarse saliency score $s_{\text{c}}(x, \mathcal{G}) \in \R$, which we define as the performance difference between $\mathcal{F}_{\text{KG}}$ and $\mathcal{F}_{\text{No-KG}}$.
For QA input $x_i = q \oplus a_i$ and its KG $\mathcal{G}_i$, let $p_{\text{KG}}(x_i, \mathcal{G}_i)$ and $p_{\text{No-KG}}(x_i)$ be the confidence probabilities for $x_i$ predicted by $\mathcal{F}_{\text{KG}}$ and $\mathcal{F}_{\text{No-KG}}$, respectively.

\begin{wrapfigure}{r}{0.48\textwidth}
% \vspace{-1cm}
\vspace{-0.5cm}

\begin{align*}
    s_{\text{c}}&(x_i, \mathcal{G}_i) \\ = \numberthis \label{eq:1}
    &\begin{cases}
        p_{\text{KG}}(x_i, \mathcal{G}_i) - p_{\text{No-KG}}(x_i), \hspace{-1.5mm} &a_i = a^*, \\
        p_{\text{No-KG}}(x_i) - p_{\text{KG}}(x_i, \mathcal{G}_i), \hspace{-1.5mm} &a_i \neq a^*.
    \end{cases}
\end{align*}
% \vspace{-0.7cm}

% \begin{align*}
%     y_{\text{c}}&(x_i, \mathcal{G}_i) \\ = \numberthis \label{eq:6}
%     &\begin{cases}
%         1, \hspace{-1.5mm} & s_{\text{c}}(x_i, \mathcal{G}_i) > T \\
%         0, \hspace{-1.5mm} & s_{\text{c}}(x_i, \mathcal{G}_i) \leq T
%     \end{cases}
% \end{align*}

\vspace{-0.3cm}
\end{wrapfigure}

Ideally, a QA model should predict higher probabilities for answer choices $a_i$ that are correct, and vice versa.
To capture this notion, we define $s_{\text{c}}(x_i, \mathcal{G}_i)$ in Eq. \ref{eq:1}, where $a^*$ denotes the correct answer. Note that $s_{\text{c}}(x_i, \mathcal{G}_i)$ is positive if $p_{\text{KG}}(x_i, \mathcal{G}_i)$ is higher than $p_{\text{No-KG}}(x_i)$ for correct choices and lower for incorrect choices.
% Then, using threshold $T$ we discretize $s_{\text{c}}(x_i, \mathcal{G}_i)$ into $y_{\text{c}}(x_i, \mathcal{G}_i)$ using Eq. \ref{eq:6}.
We obtain $y_{\text{c}}(x_i, \mathcal{G}_i)$ by binarizing $s_{\text{c}}(x_i, \mathcal{G}_i)$ to $0$ or $1$ based on whether it is greater than or less than a threshold $T$, respectively. 
If $y_\text{c}(x_i, \mathcal{G}_i) = 1$, then the KG is useful, and vice versa.
%If $s_{\text{c}}(x_i, \mathcal{G}_i) > T$, then $y_\text{c}(x_i, \mathcal{G}_i) = 1$ (\textit{i.e.}, $\mathcal{G}_i$ is useful); otherwise, $y_\text{c}(x_i, \mathcal{G}_i) = 0$ (\textit{i.e.}, $\mathcal{G}_i$ is not useful).
% Then, with threshold $T$, we discretize $s_c(x, \mathcal{G})$ into $y_{\text{c}}(x, \mathcal{G})$: If $s_{\text{c}}(x, \mathcal{G}) > T$, then $y_\text{c}(x, \mathcal{G}) = 1$ (\textit{i.e.}, $\mathcal{G}$ is useful); otherwise, $y_\text{c}(x, \mathcal{G}) = 0$ (\textit{i.e.}, $\mathcal{G}$ is not useful). 
% \xiang{touch a bit on how is T set or being studied?}
% See Appendix \ref{sec:app_coarse_sal} for more details about $\Phi$ and $T$.
See the appendix for more details about why we use ensemble-based saliency for coarse explanations (Sec. \ref{sec:app_coarse}) and how we tune $T$ (Sec. \ref{sec:app_threshold}).

\vspace{-0.2cm}
\subsection{Fine Saliency Explanations}
\label{sec:explanations_fine}
\vspace{-0.2cm}
% For a given $(x, \mathcal{G})$, some units in $\mathcal{G}$ may be useful, while others are noisy.
% Ideally, a model should only use the useful units in $\mathcal{G}$ to predict $(x, \mathcal{G})$.
Even if $\mathcal{G}$ is useful, not every part of $\mathcal{G}$ may be useful. Hence, fine saliency explanations can identify which parts of a KG are actually useful. For a given instance $(x, \mathcal{G})$, we denote the fine saliency explanation for a fine unit $u$ in $\mathcal{G}$ as $y_\text{f}(u; x, \mathcal{G}) \in \{0, 1\}$. Fine units can be nodes, paths, \textit{etc.} in the KG.
%For fine unit $u$ in $\mathcal{G}$, a fine saliency explanation $y_\text{f}(u; x, \mathcal{G}) \in \{0, 1\}$ indicates if $u$ helps the model solve $(x, \mathcal{G})$.
% Fine saliency units are either nodes or paths. 
If a graph encoder $f_\text{graph}$ encodes a certain type of unit, it is natural to define $y_\text{f}(u; x, \mathcal{G})$ with respect to such units.
For example, MHGRN \cite{feng2020scalable} encodes $\mathcal{G}$'s nodes, so we define MHGRN's fine saliency explanations with respect to nodes.
% Thus, we define $f_\text{graph}$'s units as those being updated/pooled into its graph embedding $\mathbf{g}$. 
% We refer to graph encoders using node and path units as node-based and path-based graph encoders, respectively. 
% For instance $(x, \mathcal{G})$, fine saliency assumes the model uses at least a part of $\mathcal{G}$ and indicates whether each saliency unit $u \in \mathcal{G}$ should be used for the model's prediction.
% To get $y_\text{f}(u; x, \mathcal{G})$, we first compute fine saliency score $s_{\text{f}}(u; x, \mathcal{G}) \in \R$, then discretize $s_{\text{f}}(u; x, \mathcal{G})$ into $y_\text{f}(u; x, \mathcal{G})$. 
Similar to coarse saliency explanations, to obtain $y_\text{f}(u; x, \mathcal{G})$, we first compute fine saliency score $s_{\text{f}}(u; x, \mathcal{G}) \in \R$, and then binarize it. For a QA input $x_i = q \oplus a_i$ and its KG $\mathcal{G}_i$, let $u_{ij}$ be the $j^{th}$ fine unit in $\mathcal{G}_i$ and $p_{\text{KG}}(x_i, \mathcal{G}_i)$ denote $\mathcal{F}_{\text{KG}}$'s predicted probability for $x_i$. There are many existing saliency methods (\textit{a.k.a.} attribution methods) \cite{denil2014extraction,sundararajan2017axiomatic,li2016understanding} for calculating the importance score of an input, with respect to a model and a given label. While $s_\text{f}(u_{ij}; x_i, \mathcal{G}_i)$ can be computed via any saliency method, we use gradient-based and occlusion-based methods, since they are the most common types of saliency methods \cite{bastings2020elephant}.

Let $\phi(u_{ij}; x_i, \mathcal{G}_i)$ denote the raw saliency score given by some saliency method. Gradient-based methods measure an input's saliency via the gradient of the model's output with respect to the input.   
We use the \textit{gradient$\times$input} (Grad) method \cite{denil2014extraction}, where $\phi(u_{ij}; x_i, \mathcal{G}_i)$ is the dot product of $u_{ij}$'s embedding and the gradients of $p_{\text{KG}}(x_i, \mathcal{G}_i)$ with respect to $u_{ij}$.
Occlusion-based methods measure an input's saliency as how the model's output is affected by erasing that input.
% Erasing useful inputs should have greater effect, and vice versa.
We use the \textit{leave-one-out} (Occl) method \cite{li2016understanding}, where $\phi(u_{ij}; x_i, \mathcal{G}_i)$ is the decrease in $p_{\text{KG}}(x_i, \mathcal{G}_i)$ if $u_{ij}$ is removed from $\mathcal{G}_i$, \textit{i.e.}, $\phi(u_{ij}; x_i, \mathcal{G}_i)$ = $p_{\text{KG}}(x_i, \mathcal{G}_i)$ - $p_{\text{KG}}(x_i, \mathcal{G}_i \setminus u_{ij})$.

\begin{wrapfigure}{R}{0.37\textwidth}
\vspace{-0.5cm}

\begin{align*}
    s_{\text{f}}&(u_{ij}; x_i, \mathcal{G}_i) \\ = \numberthis \label{eq:2}
    &\begin{cases}
        \phi(u_{ij}; x_i, \mathcal{G}_i), \hspace{-1.5mm} &a_i = a^* \\
        -\phi(u_{ij}; x_i, \mathcal{G}_i), \hspace{-1.5mm} &a_i \neq a^*
    \end{cases}
\end{align*}

\vspace{-0.2cm}
\end{wrapfigure}

% \yuchen{I would use $y_\text{f}(u; x, \mathcal{G})$ instead as the $u$ is the only key variable here.}

% This can be done by computing a fine saliency score $s_f(u; x, \mathcal{G})$ for each $u$. 
% While $s_f(u; x, \mathcal{G})$ can be computed using any feature attribution method, we consider gradient-based and occlusion-based methods in this paper.

% \xiang{having paragraph titles to help navigate this part.}

% After computing $s_{\text{f}}(\mathbf{u})$ for each $\mathbf{u}$ in $\mathcal{G}_i'$, we create fine saliency explanations by binarizing the top-$k$-scoring saliency units in $\mathcal{G}_i'$ as useful (positive) and the rest as not useful (negative). Let $y_{\text{f}}(\mathbf{u})$ denote the binarized version of $s_{\text{f}}(\mathbf{u})$. 
% \paragraph{Discretizing Fine Saliency Scores}
Intuitively, a unit is more useful if it increases the probability of correct answer choice $a^*$, and vice versa.
Thus, we define the saliency score $s_\text{f}(u_{ij}; x_i, \mathcal{G}_i)$ for unit $u_{ij}$ as Eq. \ref{eq:2}.
%After computing $s_{\text{f}}(u_{ij}; x_i, \mathcal{G}_i)$ for all $u_{ij}$ in $\mathcal{G}_i$,
Next, we binarize the saliency scores to get $y_\text{f}(u_{ij}; x_i, \mathcal{G}_i)$, by selecting the top-$k$\%-scoring units in $\mathcal{G}_i$ and setting $y_\text{f}(u_{ij}; x_i, \mathcal{G}_i) = 1$ (\textit{i.e.}, $u_{ij}$ is useful) for these units.
For all other units in $\mathcal{G}$, we set $y_\text{f}(u_{ij}; x_i, \mathcal{G}_i) = 0$ (\textit{i.e.}, $u_{ij}$ is not useful).
% See Appendix \ref{sec:app_fine_sal} for more details about the saliency methods and $k$.
See the appendix for more details about the fine saliency methods (Sec. \ref{sec:app_fine}) and tuning threshold $k$ (Sec. \ref{sec:app_threshold}). 
% \aaron{Like in Eq. 5 of Sec. 3.1, explain how saliency score signs are flipped for units from incorrect answer choices. Can include a more general discussion of this in the first paragraph of Sec. 3.}
% !TEX root = main.tex
\vspace{-0.1cm}
\section{\textsc{Oracle}: Using KG Saliency Explanations as Inputs}
\label{sec:oracle}
\vspace{-0.2cm}

% \xiang{You need a 1st subsection to first formalize and introduce the whole procedure of the analysis for target saliency. This should include: the high-level goal of the analysis, the specific analysis questions to look at; the exp/eval setup; the model variants to compare, etc.}

In this section, we analyze KG saliency explanations' potential to improve KG-augmented models' performance.
Recall that creating saliency explanations requires the task's ground truth labels (Sec. \ref{sec:explanations}), so directly using test set explanations is infeasible.
Still, before exploring ways to leverage training set explanations (Sec. \ref{sec:salkg}), we first establish upper bounds on how much models can benefit from saliency explanations.
Here, we study three key questions: \textbf{(1)} \textit{Does the model improve when provided oracle access to coarse/fine explanations?} \textbf{(2)} \textit{Are coarse and fine explanations complementary?} \textbf{(3)} \textit{How do gradient-based explanations compare to occlusion-based explanations?}
% To do so, we measure the model's improvement when it is provided oracle access to KG explanations.

\vspace{-0.1cm}
\subsection{\textsc{Oracle} Models}
\label{sec:oracle_models}
\vspace{-0.1cm}

\textsc{Oracle} models are KG-augmented models with oracle access to saliency explanations.
An \textsc{Oracle} model uses ground truth labels to create explanations (even at inference time), and then uses the explanations as extra inputs to perform hard attention over the units.
%$\mathcal{F}_{\text{KG}}$ (plus $\mathcal{F}_{\text{No-KG}}$, for coarse explanations) and 
% In general, we denote saliency-explanation-based attention weights as \textbf{\textit{saliency weights}}.
We define the model attention weights that are modified based on saliency explanations as \textbf{\textit{saliency weights}}.
Below, we introduce the \textsc{Oracle}-Coarse, \textsc{Oracle}-Fine, and \textsc{Oracle}-Hybrid models, shown in Fig. \ref{fig:oracle_salkg}a-c.

% As a pilot study, we first consider the class of \textsc{Oracle} models. 
% Unlike \textsc{SalKG} models (Sec. \ref{sec:salkg}), \textsc{Oracle} models directly use ground truth explanations at test time. 
% Thus, \textsc{Oracle} demonstrates \textsc{SalKG}'s potential performance gains by serving as an upper bound.
% Like base models, \textsc{Oracle} models are trained only with the task loss, so their objective is $\mathcal{L}_{\text{S}^*} = \mathcal{L}_{\text{task}}$.
% Below, we discuss the coarse, fine, and hybrid variants of \textsc{Oracle} models.

% To use KG-augmented models for multi-choice QA, we set $x = (q, a_i)$, build $\mathcal{G}_i'$ using concepts mentioned in $(q, a_i)$, and compute $(q, a_i)$'s probability as $p_{\text{KG}}(q, a_i) = \mathcal{F}_{\text{KG}}(x, \mathcal{G}_i')$.

% To use PLMs for multi-choice QA, we instead compute $(q, a_i)$'s probability as $p_{\text{No-KG}}(q, a_i) = \mathcal{F}_{\text{No-KG}}(x)$. Also, let $\Phi$ be some evaluation metric for our CSR task.

% Also, let $\mathcal{F}_{\text{No-KG}}(x) = f_\text{score}(\mathbf{x})$ denote the task output for a PLM that does not use $\mathcal{G}$.

\textbf{\textsc{Oracle}-Coarse~}
\textsc{Oracle}-Coarse ($\mathcal{F}_{\text{c}}^*$) uses coarse explanations to do hard attention over $\mathcal{F}_{\text{KG}}$'s and $\mathcal{F}_{\text{No-KG}}$'s predictions.
First, $\mathcal{F}_{\text{KG}}$ and $\mathcal{F}_{\text{No-KG}}$ are trained separately, then frozen.
% As usual, $\mathcal{F}_{\text{KG}}$ uses soft attention over nodes/paths in $\mathcal{G}$ to compute graph embedding $\mathbf{g}$ (Sec. \ref{sec:background}).
% $\mathcal{F}_{\text{KG}}$'s soft attention weights are \textit{not} saliency weights, since they are not affected by saliency explanations.
Next, for each instance $(x, \mathcal{G})$, they are used to create a coarse explanation $y_{\text{c}}(x, \mathcal{G}) \in \{0, 1\}$.
%Second, for each instance $(x, \mathcal{G})$, we use $\mathcal{F}_{\text{KG}}$ and $\mathcal{F}_{\text{No-KG}}$ to create coarse explanation $y_{\text{c}}(x, \mathcal{G}) \in \{0, 1\}$.
Then, $\mathcal{F}_{\text{c}}^*$ is defined as an ensemble model that performs hard attention over coarse units ($\mathcal{G}$ and \texttt{None}) by weighting $\mathcal{F}_{\text{KG}}$'s prediction with $y_{\text{c}}(x, \mathcal{G})$ and $\mathcal{F}_{\text{No-KG}}$'s prediction with $1 - y_{\text{c}}(x, \mathcal{G})$ (Table \ref{tab:oracle_models}; Fig. \ref{fig:oracle_salkg}a).
In other words, $y_{\text{c}}(x, \mathcal{G})$ and $1 - y_{\text{c}}(x, \mathcal{G})$ are the saliency weights for $\mathcal{F}_{\text{c}}^*$.

% Hence, $\mathcal{F}_{\text{c}}^*$'s prediction is: $\mathcal{F}_{\text{c}}^*(x, \mathcal{G}) = y_{\text{c}}(x, \mathcal{G}) \mathcal{F}_{\text{KG}}(x, \mathcal{G}) + (1 - y_{\text{c}}(x, \mathcal{G})) \mathcal{F}_{\text{No-KG}}(x)$.
% \begin{equation*}
%     \mathcal{F}_{\text{S,c}}^*(x, \mathcal{G}) = y_{\text{c}}(x) \mathcal{F}_{\text{KG}}(x, \mathcal{G}) + (1 - y_{\text{c}}(x)) \mathcal{F}_{\text{No-KG}}(x)
% \end{equation*}

% In Sec. \ref{sec:explanations_coarse}, we discussed how KG information may be helpful for some instances, but distracting for others. We also showed how binary coarse saliency targets $y_{\text{c}}$ are created to explain when the KG is helpful to the model. 
% Given instance $x$, $y_{\text{c}} = 1$ if KG is salient, and $y_{\text{c}} = 0$ if KG is not salient. Ideally, our model would use the KG-augmented model if the KG is salient, but use the PLM if the KG is not salient.
% For each instance, \textsc{SalKG}-Coarse-Oracle decides between 

\textbf{\textsc{Oracle}-Fine~}
\textsc{Oracle}-Fine ($\mathcal{F}_{\text{f}}^*$) has the same architecture as $\mathcal{F}_{\text{KG}}$ and uses fine explanations to do hard attention over fine units (\textit{i.e.}, nodes or paths in $\mathcal{G}$).
% $\mathcal{F}_{\text{f}}^*$ has the same architecture as $\mathcal{F}_{\text{KG}}$.
First, $\mathcal{F}_{\text{KG}}$ is trained, then frozen.
% Again, $\mathcal{F}_{\text{KG}}$ uses soft attention over fine units to compute $\mathbf{g}$.
As usual, $\mathcal{F}_{\text{KG}}$ uses soft attention over fine units in $\mathcal{G}$ to compute graph embedding $\mathbf{g}$ (Sec. \ref{sec:background}).
Then, for each fine unit $u$ in $\mathcal{G}$, $\mathcal{F}_{\text{KG}}$ is used to create fine explanation $y_{\text{f}}(u; x, \mathcal{G}) \in \{0, 1\}$.
Let $\hat{y}_{\text{f}}(u; x, \mathcal{G}) \in [0, 1]$ denote $\mathcal{F}_{\text{f}}^*$'s soft attention weight for $u$.
We train $\mathcal{F}_{\text{f}}^*$ the same way as $\mathcal{F}_{\text{KG}}$, except each $\hat{y}_{\text{f}}(u; x, \mathcal{G})$ is (hard attention) masked with $y_{\text{f}}(u; x, \mathcal{G})$, \textit{i.e.}, $\hat{y}_{\text{f}}(u; x, \mathcal{G}) \leftarrow \hat{y}_{\text{f}}(u; x, \mathcal{G}) \odot y_{\text{f}}(u; x, \mathcal{G})$, where $\odot$ denotes element-wise multiplication (Table \ref{tab:oracle_models}; Fig. \ref{fig:oracle_salkg}b).
% Recall that $\mathcal{F}_{\text{KG}}$'s graph encoder $f_\text{graph}$ uses attention to pool $\mathcal{G}$'s units into graph embedding $\mathbf{g}$ (Sec. \ref{sec:background}).
% $f_\text{graph}$'s attention mechanism predicts a weight for each unit $u$, indicating $u$'s importance to the model's prediction.
% Ideally, salient units would have higher attention weights than non-salient units.
% Third, $\mathcal{F}_{\text{f}}^*$ is trained the same way as $\mathcal{F}_{\text{KG}}$, but masks the attention weight for each $u$ with $y_{\text{f}}(u; x, \mathcal{G})$ (Fig. \ref{fig:oracle_salkg}b).
% \xiang{unclear concept!}
This means only units with $y_{\text{f}}(u; x, \mathcal{G}) = 1$ will have $\hat{y}_{\text{f}}(u; x, \mathcal{G}) > 0$ and thus be able to influence $\mathcal{F}_{\text{f}}^*$'s prediction.
% Note that masking is performed prior to the attention's softmax, so that the final attention output is a distribution over units in $\mathcal{G}$.
% Besides the attention masking, $\mathcal{F}_{\text{f}}^*$ has the same architecture as $\mathcal{F}_{\text{KG}}$.
Let $y_{\text{f}}(x, \mathcal{G})$ and $\hat{y}_{\text{f}}(x, \mathcal{G})$ denote the explanations and soft attention weights, respectively, for all units in the graph.
%$\hat{y}_{\text{f}}(x, \mathcal{G}) = [\hat{y}_{\text{f}}(u; x, \mathcal{G})]_{u \in \mathcal{G}}$ and $y_{\text{f}}(x, \mathcal{G}) = [y_{\text{f}}(u; x, \mathcal{G})]_{u \in \mathcal{G}}$.
Then, $\hat{y}_{\text{f}}(x, \mathcal{G}) \odot y_{\text{f}}(x, \mathcal{G})$ are the saliency weights for $\mathcal{F}_{\text{f}}^*$.

% In Sec. \ref{sec:explanations_fine}, we discussed how each KG saliency unit (e.g., node, path) can either be helpful or distracting for a given instance. We also showed how a binary fine saliency target $y_{\text{f}}$ is created to explain if a saliency unit is helpful to the model.
% Given unit $u$, $y_{f} = 1$ if $u$ is salient, and $y_{f} = 0$ if $u$ is not salient. Ideally, our model would use focus more on units for which $y_{f} = 1$.

\textbf{\textsc{Oracle}-Hybrid~}
\textsc{Oracle}-Hybrid ($\mathcal{F}_{\text{h}}^*$) unifies \textsc{Oracle}-Coarse and \textsc{Oracle}-Fine as a single model, thus leveraging the coarse-fine hierarchy inherent in KG saliency explanations.
% The \textsc{Oracle}-Hybrid model $\mathcal{F}_{\text{h}}^*$ uses both coarse and fine explanations, thus leveraging the hierarchy inherent in KG saliency explanations.
First, $\mathcal{F}_{\text{f}}^*$ (which uses fine explanations) and $\mathcal{F}_{\text{No-KG}}$ are separately trained, then frozen.
Then, for each $(x, \mathcal{G})$, $\mathcal{F}_{\text{f}}^*$ and $\mathcal{F}_{\text{No-KG}}$ are used to create $y_{\text{h}}(x, \mathcal{G}) \in \{0, 1\}$, which we define as the coarse explanation for $\mathcal{F}_{\text{f}}^*$ and $\mathcal{F}_{\text{No-KG}}$.
$y_{\text{h}}(x, \mathcal{G})$ is computed the same way as $y_{\text{c}}(x, \mathcal{G})$, besides replacing $\mathcal{F}_{\text{KG}}$ with $\mathcal{F}_{\text{f}}^*$.
% Third, $\mathcal{F}_{\text{h}}^*$ is an ensemble that uses $y_{\text{h}}(x, \mathcal{G})$ to choose between $\mathcal{F}_{\text{f}}^*$'s and $\mathcal{F}_{\text{No-KG}}$'s prediction (Table \ref{tab:oracle_models}; Fig. \ref{fig:oracle_salkg}c).
Finally, similar to $\mathcal{F}_{\text{c}}^*$, $\mathcal{F}_{\text{h}}^*$ is an ensemble that performs hard attention over coarse units by weighting $\mathcal{F}_{\text{f}}^*$'s prediction with $y_{\text{h}}(x, \mathcal{G})$ and $\mathcal{F}_{\text{No-KG}}$'s prediction with $1 - y_{\text{h}}(x, \mathcal{G})$ (Table \ref{tab:oracle_models}; Fig. \ref{fig:oracle_salkg}c).
That is, $y_{\text{h}}(x, \mathcal{G})$ and $1 - y_{\text{h}}(x, \mathcal{G})$ are the saliency weights for $\mathcal{F}_{\text{h}}^*$.
% Ideally, the coarse/fine explanations would provide complementary signals for training $\mathcal{F}_{\text{h}}^*$.

% , while $\mathcal{F}_{\text{f}}^*$ itself uses $y_{\text{f}}(u; x, \mathcal{G})$ to decide whether to use each unit $u \in \mathcal{G}$ (Fig. \ref{fig:oracle_salkg}b).
% Thus, $\mathcal{F}_{\text{h}}^*$'s prediction is: $\mathcal{F}_{\text{h}}^*(x, \mathcal{G}) = y_{\text{h}}(x, \mathcal{G}) \mathcal{F}_{\text{f}}^*(x, \mathcal{G}) + (1 - y_{\text{h}}(x, \mathcal{G})) \mathcal{F}_{\text{No-KG}}(x)$.
% \begin{equation*}
%     \mathcal{F}_{\text{S,h}}^*(x, \mathcal{G}) = y_{\text{c}}(x) \mathcal{F}_{\text{S,f}}^*(x, \mathcal{G}) + (1 - y_{\text{c}}(x)) \mathcal{F}_{\text{No-KG}}(x)
% \end{equation*}
% \soumya{It is not clear how fine saliency is used in this model}

% \begin{wraptable}{R}{\textwidth}
\begin{table}[t]
% \vspace{-1.7cm}
\centering
\renewcommand\arraystretch{1.5}
\scalebox{0.73}{
\begin{tabular}{lcc}
    \toprule  
    \textbf{Model} & \textbf{Output} & \textbf{Saliency Weights} \\ 
    % & \textbf{Saliency Loss ($\mathcal{L}_{\text{sal}}$)} \\
    \midrule
    {\textsc{Oracle}-Coarse} 
        & $\mathcal{F}_{\text{c}}^*(x, \mathcal{G}) = y_{\text{c}}(x, \mathcal{G}) \mathcal{F}_{\text{KG}}(x, \mathcal{G}) + (1 - y_{\text{c}}(x, \mathcal{G})) \mathcal{F}_{\text{No-KG}}(x)$
        & $[y_{\text{c}}(x, \mathcal{G}), 1 - y_{\text{c}}(x, \mathcal{G})]$ \\
        % & N/A \\
    {\textsc{Oracle}-Fine}
        & $\mathcal{F}_{\text{f}}^*(x, \mathcal{G}) \sim \mathcal{F}_{\text{KG}}(x, \mathcal{G})$
        & $\hat{y}_{\text{f}}(x, \mathcal{G}) \odot y_{\text{f}}(x, \mathcal{G})$ \\
        % & N/A \\
    {\textsc{Oracle}-Hybrid}
        & $\mathcal{F}_{\text{h}}^*(x, \mathcal{G}) = y_{\text{h}}(x, \mathcal{G}) \mathcal{F}_{\text{f}}^*(x, \mathcal{G}) + (1 - y_{\text{h}}(x, \mathcal{G})) \mathcal{F}_{\text{No-KG}}(x)$
        & $[y_{\text{h}}(x, \mathcal{G}), 1 - y_{\text{h}}(x, \mathcal{G})]$ \\
        % & N/A \\
    \bottomrule
\end{tabular}
}
\vspace{0.2cm}
\caption{\small \textbf{Comparison of \textsc{Oracle} Models.} For each \textsc{Oracle} Model, we show its output and saliency weights. Note that the explanations are given (not predicted), so there is no $\mathcal{L}_{\text{sal}}$. 
While $\mathcal{F}_{\text{c}}^*$ and $\mathcal{F}_{\text{h}}^*$ are both ensembles of $\mathcal{F}_{\text{KG}}$ and $\mathcal{F}_{\text{No-KG}}$, $\mathcal{F}_{\text{f}}^*$ has the same architecture as $\mathcal{F}_{\text{KG}}$ (denoted by $\sim$) besides the attention masking.}

\label{tab:oracle_models}
\vspace{-0.5cm}
\end{table}

\vspace{-0.1cm}
\subsection{Evaluation Protocol}
\label{sec:oracle_eval}
\vspace{-0.1cm}

% \paragraph{Datasets}
We use the CSQA \cite{talmor2019commonsenseqa} and OBQA \cite{mihaylov2018can} multi-choice QA datasets. 
For CSQA, we use the accepted in-house data split from \cite{lin2019kagnet}, as the official test labels are not public. 
As in prior works, we use the ConceptNet \cite{speer2017conceptnet} KG for both datasets. 
We report accuracy, the standard metric for multi-choice QA.
% For all models (except \textsc{Oracle}-Coarse, which is not trained), we report the mean accuracy over three seeds.
For $\mathcal{F}_{\text{No-KG}}$ and $\mathcal{F}_{\text{KG}}$, we pick the best model over three seeds, then use them to create explanations for \textsc{Oracle} models.
We use thresholds $T=0.01$ and $k=10$ for coarse and fine explanations, respectively. 
% Thus, for fair comparison, we report performance only for the best (single-seed) baselines and the \textsc{Oracle} models built from them.
% \paragraph{KG-Augmented Models}
For text encoders, we use BERT(-Base) \cite{devlin2018bert} and RoBERTa(-Large) \cite{liu2019roberta}. For graph encoders, we use MHGRN \cite{feng2020scalable}, PathGen \cite{wang2020connecting}, and Relation Network (RN) \cite{santoro2017simple, lin2019kagnet}. MHGRN has node units, while PathGen and RN have path units.
% \subsection{Baseline Models}
% \label{sec:oracle_baselines}
% \xiang{move and merge.}
As \textit{\textbf{baseline models}}, we use $\mathcal{F}_{\text{No-KG}}$, $\mathcal{F}_{\text{KG}}$, and $\mathcal{F}_{\text{No-KG}} + \mathcal{F}_{\text{KG}}$, where $\mathcal{F}_{\text{No-KG}} + \mathcal{F}_{\text{KG}}$ is an ensemble whose prediction is the mean of $\mathcal{F}_{\text{No-KG}}$'s and $\mathcal{F}_{\text{KG}}$'s predictions.
\textsc{Oracle} and baseline models are trained only with task loss $\mathcal{L}_{\text{task}}$.
% , so their learning objective is $\mathcal{L}_{\text{base}} = \mathcal{L}_{\text{task}}$. 
% \xiang{some important details are skipped.}
% As described in Sec. \ref{sec:explanations}, $\mathcal{F}_{\text{No-KG}}$ and $\mathcal{F}_{\text{KG}}$ are also used to create coarse/fine saliency explanations. 
% We also call these explanations \textit{saliency targets}, since they serve as learning targets for the \textsc{SalKG} model.

\begin{table*}[t]
% \vspace{-1.1cm}
\centering
% \vspace{-0.2cm}
\scalebox{0.61}{
\begin{tabular}{lcccccccccccc}
    \toprule &
    \multicolumn{6}{c}{ \textbf{CSQA Test Accuracy (\%)}} & \multicolumn{6}{c}{ \textbf{OBQA Test Accuracy (\%)}}\\
    \cmidrule(lr){2-7} \cmidrule(lr){8-13}
    & \multicolumn{2}{c}{ \textbf{MHGRN}} & \multicolumn{2}{c}{ \textbf{PathGen}} & \multicolumn{2}{c}{ \textbf{RN}} & \multicolumn{2}{c}{ \textbf{MHGRN}} & \multicolumn{2}{c}{ \textbf{PathGen}} & \multicolumn{2}{c}{ \textbf{RN}} \\
    \cmidrule(lr){2-3}
    \cmidrule(lr){4-5}
    \cmidrule(lr){6-7}
    \cmidrule(lr){8-9}
    \cmidrule(lr){10-11}
    \cmidrule(lr){12-13}
    \textbf{Model} & BERT & RoBERTa & BERT & RoBERTa & BERT & RoBERTa & BERT & RoBERTa & BERT & RoBERTa & BERT & RoBERTa \\
    \midrule
    {No-KG} & 55.44 & 70.59 & 55.44 & 70.59  & 55.44 & 70.59  & 53.60 & 68.40  & 53.60 & 68.40  & 53.60 & 68.40  \\
    {KG} & 56.57 & 73.33 & 56.65 & 72.04 & 55.60 & 71.07 & 53.20 & 69.80 & 55.00 & 67.80 & 58.60 & 70.20 \\
    {No-KG + KG} & 56.57 & 71.39 & 57.45 & 73.00 & 56.73 & 68.49 & 55.60 & 70.60 & 54.40 & 70.6 & 53.40 & 69.60 \\
    % {No-KG + KG} & - & - & - & - & - & - \\
    % {KG} & 58.77 ($\pm$0.49) & 73.14 ($\pm$0.78) & 56.54 ($\pm$0.73) & 72.58 ($\pm$0.57) & 56.46 ($\pm$1.22) & 71.37 ($\pm$1.20) \\
    % {\textsc{SalKG}-Coarse} & 56.89 ($\pm$0.00) & 74.05 ($\pm$0.14) & 58.23 ($\pm$0.37) & 72.79 ($\pm$0.12) & 57.59 ($\pm$0.33) & 72.44 ($\pm$0.16) \\
    % {\textsc{SalKG}-Fine (Grad)} & 53.26 ($\pm$2.24) & - & 56.35 ($\pm$1.91) & - & 56.14 ($\pm$1.97) & - \\
    % {\textsc{SalKG}-Fine (Occl)} & 56.78 ($\pm$2.14) & 73.49 ($\pm$0.56) & 57.64 ($\pm$2.12) & 71.39 ($\pm$1.54) & 56.86 ($\pm$0.41) & 71.58 ($\pm$1.10) \\
    % {\textsc{SalKG}-Hybrid (Grad)} & - & - & - & - & - & - \\
    % {\textsc{SalKG}-Hybrid (Occl)} & - & - & - & - & - & - \\
    %{$\max(\text{No-KG}, \text{KG})$} & 56.57 & 73.33 & 56.65 & 72.04 & 55.60 & 71.07 \\
    \midrule
    {\textsc{Oracle}-Coarse} & 66.16 & 81.39 & 68.57 & 80.10 & 67.28 & 79.69 & 70.60 & 79.40 & 65.00 & 76.60 & 69.00 & 79.00 \\
    % {\textsc{Oracle}-Coarse (Grad)} & - & - & - & - & - & - \\
    % {\textsc{Oracle}-Coarse (Occl)} & - & - & - & - & - & - \\
    \midrule
    {\textsc{Oracle}-Fine (Grad)} & 74.86 & 76.15 & 79.61 & 87.35 & 81.39 & 83.24 & 67.60 & 72.60 & 73.80 & 73.40 & 68.00 & 62.80 \\
    {\textsc{Oracle}-Fine (Occl)} & 91.06 & 87.99 & 79.61 & 75.34 & 73.73 & 68.41 & 77.00 & 71.20 & 83.60 & 62.60 & 55.60 & 61.40 \\
    \midrule
    {\textsc{Oracle}-Hybrid (Grad)} & 85.50 & 84.21 & \textbf{90.49} & 92.83 & \textbf{92.26} & 93.56 & 80.80 & 84.80 & 85.60 & \textbf{92.80} & \textbf{85.40} & \textbf{86.80} \\
    {\textsc{Oracle}-Hybrid (Occl)} & \textbf{95.89} & \textbf{98.63} & 88.96 & \textbf{96.78} & 85.25 & \textbf{95.25} & \textbf{87.00} & \textbf{89.60} & \textbf{92.80} & 90.60 & 67.40 & 80.60 \\
    % {\textsc{Oracle}-Hybrid (Grad)} & - & - & - & - & - & - \\
    % {\textsc{Oracle}-Hybrid (Occl)} & - & - & - & - & - & - \\
    % {\textsc{SalKG}-Coarse (Ensemble)} & \textbf{57.37} ($\pm$0.08) & \textbf{74.03} ($\pm$0.09) & \textbf{58.23} ($\pm$0.82) & \textbf{72.93} ($\pm$0.29) & \textbf{58.15} ($\pm$0.05) & \textbf{73.73} ($\pm$0.45) \\
    % {\textsc{SalKG}-Fine (Unnormalized top-10)} & 55.44 ($\pm$1.81) & 73.46 ($\pm$0.69) & 53.34 ($\pm$1.75) & 70.99 ($\pm$0.85) & 56.05 ($\pm$0.54) & 69.59 ($\pm$1.26) \\
    % {\textsc{SalKG}-Coarse-Target (Ensemble)} & 66.16 & 82.60 & 68.57 & 80.10 & 67.45 & 79.69 \\
    % {\textsc{SalKG}-Coarse-Target} & - & 81.92 ($\pm$0.58) & - & - & - & - \\
    % {\textsc{SalKG}-Fine-Target (Unnormalized top-10)} & 66.85 ($\pm$5.33) & 77.14 ($\pm$3.14) & 81.57 ($\pm$0.36) & 82.06 ($\pm$4.12) & 79.10 ($\pm$0.74) & 79.85 ($\pm$10.47) \\
    % {\textsc{SalKG}-Fine-Target} & 67.15^* ($\pm$1.58) & 79.45 ($\pm$1.30) & - & - & - & - \\
    \bottomrule 
\end{tabular}
}
\vspace{-0.1cm}
\caption{\small \textbf{\textsc{Oracle} Performance on CSQA and OBQA}
}
\label{tab:oracle}
\vspace{-0.2cm}
\end{table*}

\vspace{-0.1cm}
\subsection{Analysis}
\label{sec:oracle_analysis}
\vspace{-0.1cm}
% \aaron{Will work on this subsection later.}
% \textcolor{blue}{\textbf{Aaron: Will polish this subsection later.}}
In Table \ref{tab:oracle}, we show CSQA and OBQA performance for the baseline and \textsc{Oracle} models.
%$\mathcal{F}_{\text{No-KG}}$ (No-KG), $\mathcal{F}_{\text{KG}}$ (KG), $\mathcal{F}_{\text{No-KG}} + \mathcal{F}_{\text{KG}}$ (No-KG + KG), and \textsc{Oracle} models.
We analyze these results via the three questions below.

% In this section, we analyze the performance of the \textsc{Oracle} models compared to the PLM (No-KG) and KG-augmented PLM (KG) baseline models. The results are reported in Tables \ref{tab:csqa_target} and \ref{tab:obqa_target} for CSQA and OBQA respectively.

\textbf{\textit{Does the model improve when provided oracle access to coarse/fine explanations?}~}
Yes.
\textsc{Oracle}-Coarse beats all baselines, while \textsc{Oracle}-Fine beats all baselines except on OBQA RN+RoBERTa.
%One reason for OBQA RN+RoBERTa's performance drop is that $k=10$ may not be an optimal threshold for this setting.
%While we do not tune thresholds for \textsc{Oracle}, we do later for \textsc{SalKG} and find that this can affect performance.
These results motivate us to develop a framework for models to improve performance by learning from coarse/fine explanations.
Also, on average, \textsc{Oracle}-Fine outperforms \textsc{Oracle}-Coarse, which suggests that fine explanations may often provide richer signal than their coarse counterparts.
Indeed, fine explanations indicate the saliency of every unit in the KG, while coarse explanations only indicate the saliency of the KG as a whole.

% We observe that \textsc{Oracle}-Coarse, which uses the ground truth coarse saliency explanations, consistently outperform the No-KG and KG models on all settings. This gives us the motivation to develop a model framework using which we can learn to generate the ground truth explanations from the training explanations.
% \textsc{Oracle}-Fine models also consistently outperform the baseline models. Moreover, we see that on average, \textsc{Oracle}-Fine is also better than \textsc{Oracle}-Coarse model. This suggests that fine saliency explanations have a richer information than the coarse counterparts.

\textbf{\textit{Are coarse and fine explanations complementary?}~}
Yes.
Across all settings, \textsc{Oracle}-Hybrid performs significantly better than \textsc{Oracle}-Coarse and \textsc{Oracle}-Fine.
This suggests that coarse and fine explanations are complementary and that it is effective to leverage both hierarchically.

% Lastly, we report the performance of the \textsc{Oracle}-Hybrid models which use both coarse and fine explanations. We find that these models are significantly better than any of the baselines or the individual coarse/fine models. It suggests that both the coarse and fine saliencies contain complementary information and our proposed way of combining these explanations hierarchically is indeed effectively leveraging that.

\textbf{\textit{How do gradient-based explanations compare to occlusion-based explanations?}~}
Overall, Grad and Occl perform similarly.
Grad performs better on some settings (\textit{e.g.}, MHGRN), while Occl performs better on others (\textit{e.g.}, RN).
See Table \ref{tab:ablation} and Sec. \ref{sec:app_grad_vs_occl} for more Grad \textit{vs.} Occl experiments.
% We will further compare Grad and Occl in later experiments.

% Next, we use two variants of fine saliency explanations - Gradient (grad) and Occlusion (occl) to evaluate \textsc{Oracle}-Fine models. While we find that both methods improve upon the coarse saliency model, there is no clear trend between the two variants. Hence, we continue exploring both the explanation sources for rest of the experiments.

In our \textsc{Oracle} pilot study, KG-augmented models achieve large performance gains when given explanations as input.
This suggests that, if oracle explanations can somehow be \textit{predicted} accurately during inference without using ground truth labels, then KG-augmented models can still achieve improvements without directly using explanations as input.
This motivates us to train KG-augmented models with explanation-based supervision via \textsc{SalKG}, which we describe in Sec. \ref{sec:salkg}.

% \yuchen{I would suggest to add a short summary here as the conclusion, then add some words lead to the motivation/support evidence for further studying SalKG. The analysis here suggests that if we are able to predict the explanations correctly (as in the oracle models), it is very promising to improve the model performance. }

% !TEX root = main.tex
\vspace{-0.2cm}
\section{\textsc{SalKG}: Using KG Saliency Explanations as Supervision} 
\label{sec:salkg}
\vspace{-0.2cm}

\definecolor{blue-violet}{rgb}{0.54, 0.17, 0.89}
\definecolor{hanpurple}{rgb}{0.32, 0.09, 0.98}
\begin{figure}[tb!]
\begin{center}
    \centering
% \vspace{-0.4cm}
    \includegraphics[width=\textwidth]{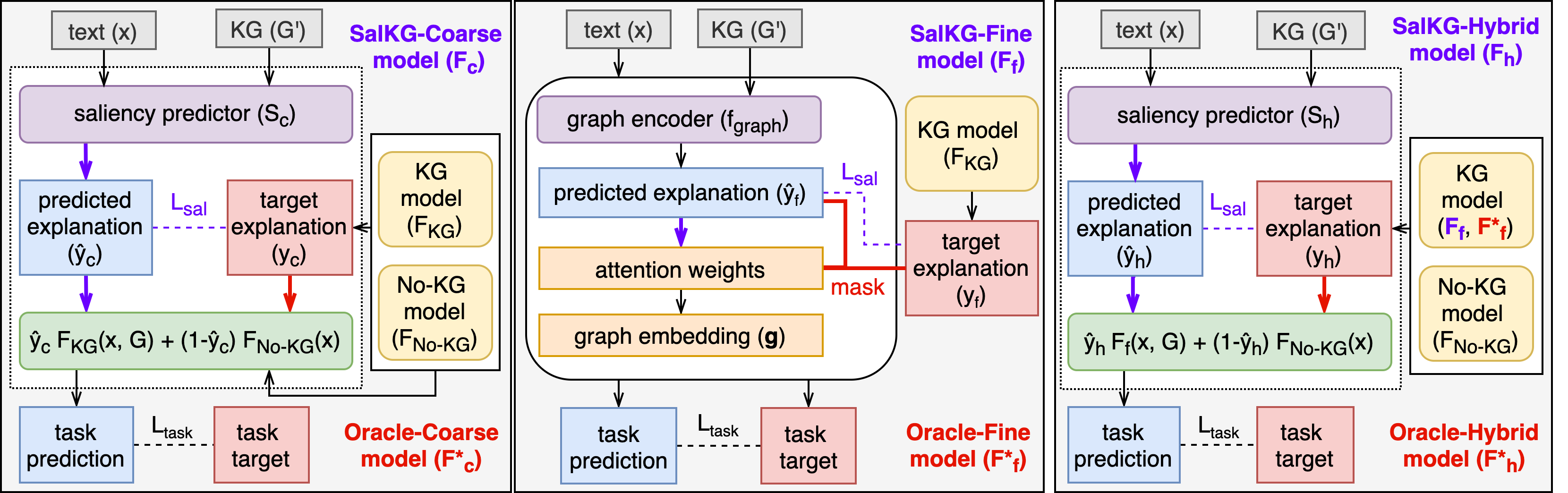}
    % \vspace{-0.6cm}
    \caption{\small \textbf{Schematics for \textsc{Oracle} and \textsc{SalKG} Models.} \textcolor{red}{Red} arrows indicate the \textsc{Oracle} pipeline, where the target explanation is provided as input. \textcolor{hanpurple}{Purple} arrows indicate the \textsc{SalKG} pipeline, where the target explanation is used as supervision for the predicted explanation. In \textsc{SalKG}-Coarse and \textsc{SalKG}-Hybrid, the saliency predictor has the same architecture as $\mathcal{F}_{\text{KG}}$. Meanwhile, \textsc{Oracle}-Fine and \textsc{SalKG}-Fine (shown as white module, with text encoder and task predictor omitted) both have the same architecture as $\mathcal{F}_{\text{KG}}$.}
    \label{fig:oracle_salkg}
\vspace{-0.5cm}
\end{center}
\end{figure}

Based on the analysis from Sec. \ref{sec:oracle_analysis}, we propose the \textsc{SalKG} framework for KG-augmented models to learn from coarse/fine saliency explanations.
Whereas \textsc{Oracle} models (Sec. \ref{sec:oracle_models}) use explanations directly as extra inputs, \textsc{SalKG} models only use them as extra supervision during the training phase.
% Like \textsc{Oracle} models (Sec. \ref{sec:oracle_models}) \textsc{SalKG} model uses $\mathcal{F}_{\text{KG}}$ (plus $\mathcal{F}_{\text{No-KG}}$, for coarse explanations) and ground truth labels to create explanations, but then uses the explanations as extra supervision instead of as extra inputs.
With explanations created from the training set via $\mathcal{F}_{\text{KG}}$ and $\mathcal{F}_{\text{No-KG}}$, \textsc{SalKG} models are jointly trained to predict the explanations (via saliency loss $\mathcal{L}_{\text{sal}}$) and use the predicted explanations to solve the task (via task loss $\mathcal{L}_{\text{task}}$).
Thus, \textsc{SalKG} models have the following objective: $\mathcal{L}_{\text{S}} = \mathcal{L}_{\text{task}} + \lambda \mathcal{L}_{\text{sal}}$, where $\lambda \geq 0$ is a loss weighting parameter.
This multitask objective not only encourages \textsc{SalKG} models to focus on useful KG units for solving the task, but also to learn more general graph/node/path representations. 
% the design there is to force the model to predict oracle explanations + making accurate QA predictions based on the predicted explanations.
% A \textsc{SalKG} model has the same architecture as its corresponding $\mathcal{F}_{\text{KG}}$ base model.
% , but augmented with a saliency predictor $\mathcal{S}_{\text{KG}}$.
Below, we present \textsc{SalKG}-Coarse, \textsc{SalKG}-Fine, and \textsc{SalKG}-Hybrid models.
% , and define $\mathcal{L}_{\text{sal}}$ for each model variant.
% , which are analogous to \textsc{Oracle} models.

% At a high level, the \textsc{SalKG} model $\mathcal{F}^*_{\text{KG}}$ consists of two components: a saliency predictor $\mathcal{S}_{\text{KG}}$ and a task predictor $\mathcal{T}_{\text{KG}}$.
% Because creating explanations requires ground truth labels for the CSR task, explanations are not available to the model at test time. Thus, using saliency explanations created from the training set, $\mathcal{S}_{\text{KG}}$ is trained to predict which KG inputs are likely to be salient for a given task instance.
% Then, $\mathcal{T}_{\text{KG}}$ uses $\mathcal{S}_{\text{KG}}$'s saliency predictions to output a prediction for the CSR task.
% The saliency model has essentially the same architecture as $\mathcal{F}_{\text{KG}}$, but is also trained to upweight KG inputs labeled as salient by the explanations. 
% Below, we describe how $\mathcal{F}^*_{\text{KG}}$ is implemented for coarse (\textsc{SalKG}-Coarse) and fine (\textsc{SalKG}-Fine) saliency.

% $\mathcal{F}^*_{\text{KG}}$

% We train $\mathcal{F}^*_{\text{KG}}$ with the downstream task loss and a saliency prediction loss. For both coarse and fine saliency, the saliency predictor is a self-attention model. 

% \xiang{1) current writing is too brief and hard for new readers to follow the details; 2) make reference to Figures throughout your introduction of the following methods, to make them accessible to new readers.}

\textbf{\textsc{SalKG}-Coarse~}
Unlike \textsc{Oracle}-Coarse, \textsc{SalKG}-Coarse ($\mathcal{F}_{\text{c}}$) is not given oracle coarse explanation $y_\text{c}(x, \mathcal{G})$ as input. Instead, a saliency predictor $\mathcal{S}_{\text{c}}$ (with the same architecture as $\mathcal{F}_{\text{KG}}$) is trained to predict the oracle coarse explanation. $\mathcal{S}_{\text{c}}$ predicts coarse explanation as probability $\hat{y}_{\text{c}}(x, \mathcal{G}) \in [0, 1]$. $\mathcal{F}_{\text{c}}$'s output is an ensemble that does soft attention over coarse units by weighting $\mathcal{F}_{\text{KG}}$'s and $\mathcal{F}_{\text{No-KG}}$'s predictions with saliency weights $\hat{y}_{\text{c}}(x, \mathcal{G})$ and $1 - \hat{y}_{\text{c}}(x, \mathcal{G})$, respectively (Table \ref{tab:salkg_models}; Fig. \ref{fig:oracle_salkg}a). Here, $\mathcal{L}_{\text{sal}}(\hat{y}_{\text{c}}(x, \mathcal{G}), y_{\text{c}}(x, \mathcal{G}))$ is the cross-entropy loss.

\textbf{\textsc{SalKG}-Fine~}
% Let $\mathcal{F}_{\text{f}}$ denote the \textsc{SalKG}-Fine model, which has the same architecture as $\mathcal{F}_{\text{KG}}$.
Similarly, \textsc{SalKG}-Fine ($\mathcal{F}_{\text{f}}$) is not given oracle fine explanation $y_{\text{f}}(u; x, \mathcal{G})$ as input, although both have the same architecture as $\mathcal{F}_{\text{KG}}$.
Instead, for each fine unit $u$, $\mathcal{F}_{\text{f}}$'s attention mechanism is trained to predict $y_{\text{f}}(u; x, \mathcal{G})$ as soft attention weight $\hat{y}_{\text{f}}(u; x, \mathcal{G}) \in [0, 1]$ (Table \ref{tab:salkg_models}; Fig. \ref{fig:oracle_salkg}b).
% Recall that $\hat{y}_{\text{f}}(u; x, \mathcal{G}) \in [0, 1]$ is the soft attention weight for fine unit $u$ in $\mathcal{G}$.
% While \textsc{Oracle}-Fine directly masks each $\hat{y}_{\text{f}}(u; x, \mathcal{G})$ with $y_{\text{f}}(u; x, \mathcal{G})$, $\mathcal{F}_{\text{f}}$'s attention mechanism is trained such that $\hat{y}_{\text{f}}(u; x, \mathcal{G})$ approximates $y_{\text{f}}(u; x, \mathcal{G})$ (Table \ref{tab:salkg_models}; Fig. \ref{fig:oracle_salkg}b).
As before, $\hat{y}_{\text{f}}(x, \mathcal{G}) = [\hat{y}_{\text{f}}(u; x, \mathcal{G})]_{u \in \mathcal{G}}$ are the soft attention weights for $(x, \mathcal{G})$, while $y_{\text{f}}(x, \mathcal{G}) = [y_{\text{f}}(u; x, \mathcal{G})]_{u \in \mathcal{G}}$ are the fine explanations for $(x, \mathcal{G})$.
Then, $\hat{y}_{\text{f}}(x, \mathcal{G})$ are the saliency weights for $\mathcal{F}_{\text{f}}$, trained with KL divergence loss $\mathcal{L}_{\text{sal}}(\hat{y}_{\text{f}}(x, \mathcal{G}), y_{\text{f}}(x, \mathcal{G}))$.

% Note that $\mathcal{F}_{\text{f}}$ has the same architecture as $\mathcal{F}_{\text{f}}^*$ and $\mathcal{F}_{\text{KG}}$.

% In state-of-the-art graph encoders, all saliency unit embeddings within a KG are pooled into a single graph embedding via multi-head self-attention. Thus, we train the \textsc{SalKG}-Fine model to upweight salient units by regularizing its graph encoder's attention weights to approximate the fine saliency targets. For \textsc{SalKG}-Fine, $\mathcal{F}^*_{\text{KG}}$ has the same architecture as $\mathcal{F}_{\text{KG}}$. Here, $\mathcal{S}_{\text{KG}}$ is $\mathcal{F}_{\text{KG}}$'s self-attention mechanism, which predicts each unit's saliency target, and $\mathcal{T}_{\text{KG}}$ is $\mathcal{F}_{\text{KG}}$'s task predictor MLP, which directly predicts the CSR task labels as $\mathcal{T}_{\text{KG}}(x, \mathcal{G})$. Thus, $\mathcal{F}^*_{\text{KG}}$ is again trained using the following objective: $\mathcal{L} = \mathcal{L}_{\text{task}} + \lambda \mathcal{L}_{\text{sal}}$, where $\mathcal{L}_{\text{sal}}$ is still cross entropy loss.

% \begin{wrapfigure}{R}{0.4\textwidth}
%     \centering
% \vspace{-0.7cm}
%     \includegraphics[width=0.4\textwidth]{figures/salkg_fine.png}
%     % \vspace{-0.6cm}
%     \caption{\small \textbf{\textsc{SalKG}-Fine Model}}
%     \label{fig:fine_model}
% \vspace{-1cm}
% \end{wrapfigure}

\textbf{\textsc{SalKG}-Hybrid~}
Similar to the other \textsc{SalKG} variants, \textsc{SalKG}-Hybrid ($\mathcal{F}_{\text{h}}$) does not use any oracle explanations. 
% First, $\mathcal{F}_{\text{f}}$ and $\mathcal{F}_{\text{No-KG}}$ are separately trained, then frozen. 
Like in \textsc{SalKG}-Coarse, a saliency predictor $\mathcal{S}_{\text{h}}$ is trained to predict oracle coarse explanation $y_h(x, \mathcal{G})$ (Sec. \ref{sec:oracle_models}). Predicted coarse explanation probabilities $\hat{y}_{\text{h}}(x, \mathcal{G}) \in [0, 1]$ are then used as soft attention over coarse units by weighting $\mathcal{F}_{\text{f}}$'s and $\mathcal{F}_{\text{No-KG}}$'s predictions with weights $\hat{y}_{\text{h}}(x, \mathcal{G})$ and $1 - \hat{y}_{\text{h}}(x, \mathcal{G})$, respectively (Table \ref{tab:salkg_models}; Fig. \ref{fig:oracle_salkg}c). Here, $\mathcal{L}_{\text{sal}}(\hat{y}_{\text{h}}(x, \mathcal{G}), y_{\text{h}}(x, \mathcal{G}))$ is cross-entropy loss.

\begin{table}[t]
\vspace{-0.4cm}
\centering
\renewcommand\arraystretch{1.5}
\scalebox{0.73}{
\begin{tabular}{lccc}
    \toprule  
    \textbf{Model} & \textbf{Output} & \textbf{Saliency Weights} & \textbf{Saliency Loss ($\mathcal{L}_{\text{sal}}$)} \\
    \midrule
    {\textsc{SalKG}-Coarse} 
        & $\mathcal{F}_{\text{c}}(x, \mathcal{G}) = \hat{y}_{\text{c}}(x, \mathcal{G}) \hspace{0.5mm} \mathcal{F}_{\text{KG}}(x, \mathcal{G}) + (1 - \hat{y}_{\text{c}}(x, \mathcal{G})) \hspace{0.5mm} \mathcal{F}_{\text{No-KG}}(x)$
        % & $\alpha(x, \mathcal{G})$ 
        & $[\hat{y}_{\text{c}}(x, \mathcal{G}), 1 - \hat{y}_{\text{c}}(x, \mathcal{G})]$
        & CE$(\hat{y}_{\text{c}}(x, \mathcal{G}), y_{\text{c}}(x, \mathcal{G}))$ \\
    {\textsc{SalKG}-Fine} 
        & $\mathcal{F}_{\text{f}}(x, \mathcal{G}) \sim \mathcal{F}_{\text{KG}}(x, \mathcal{G})$
        & $\hat{y}_{\text{f}}(x, \mathcal{G})$
        & KL$(\hat{y}_{\text{f}}(x, \mathcal{G}), y_{\text{f}}(x, \mathcal{G}))$ \\
    {\textsc{SalKG}-Hybrid} 
        & $\mathcal{F}_{\text{h}}(x, \mathcal{G}) = \hat{y}_{\text{h}}(x, \mathcal{G}) \mathcal{F}_{\text{f}}(x, \mathcal{G}) + (1 - \hat{y}_{\text{h}}(x, \mathcal{G})) \mathcal{F}_{\text{No-KG}}(x)$ 
        & $[\hat{y}_{\text{h}}(x, \mathcal{G}), 1 - \hat{y}_{\text{h}}(x, \mathcal{G})]$ 
        & CE$(\hat{y}_{\text{h}}(x, \mathcal{G}), y_{\text{h}}(x, \mathcal{G}))$ \\
    \bottomrule
\end{tabular}
}
\vspace{0.1cm}
\caption{\small \textbf{Comparison of \textsc{SalKG} Models.} For each \textsc{SalKG} Model, we show its output, saliency weights, and $\mathcal{L}_{\text{sal}}$. While $\mathcal{F}_{\text{c}}$ and $\mathcal{F}_{\text{h}}$ are both ensembles, $\mathcal{F}_{\text{f}}$ has the same architecture as $\mathcal{F}_{\text{KG}}$ (denoted by $\sim$). ``CE'' denotes cross-entropy loss, while ``KL'' denotes KL divergence loss.}

\label{tab:salkg_models}
\vspace{-0.6cm}
\end{table}
% \end{wraptable}
% !TEX root = main.tex
\vspace{-0.2cm}
\section{Experiments} 
\label{sec:exp}
\vspace{-0.2cm}

% \yuchen{
% Some quick comments on Sec 6:
% 1) It seems that Table 5 and 6 can be summarized to a single table by only showing RoBERTa results, and leave the BERT results in the appendix.
% 2) Also, would be great to see some qualitative analysis with case studies for readers who are curious what saliency explanations produced by SalKG look like.
% }

\subsection{Evaluation Protocol}
\label{sec:exp_eval}
We evaluate \textsc{SalKG} models on the CSQA \cite{talmor2019commonsenseqa}, OBQA \cite{mihaylov2018can}, and CODAH \cite{chen2019codah} multi-choice QA datasets (Sec. \ref{sec:app_datasets}). 
%Our experiments show that \textsc{SalKG} models can outperform various baseline models.
In addition to the baselines in Sec. \ref{sec:oracle_eval}, we consider two new baselines, \textsc{Random} and \textsc{Heuristic}, which help show that coarse/fine saliency explanations provide strong learning signal for KG-augmented models to focus on useful KG features.
We follow the same evaluation protocol in Sec. \ref{sec:oracle_eval}, except we now also report mean and standard deviation performance over multiple seeds. 
See Sec. \ref{sec:app_eval_protocol} for a more detailed description of the evaluation protocol.

\textbf{\textsc{Random}~}
\textsc{Random} is a variant of \textsc{SalKG} where each unit's explanation is random.
\textsc{Random}-Coarse is like \textsc{SalKG}-Coarse, but with each $y_\text{c}(x, \mathcal{G})$ uniformly sampled from $\{0, 1\}$. 
\textsc{Random}-Fine is like \textsc{SalKG}-Fine, but randomly picking $k$\% of units in $\mathcal{G}$ to set $y_\text{f}(u; x, \mathcal{G}) = 1$.
%, while setting $y_\text{f}(u; x, \mathcal{G}) = 0$ for all other units.
\textsc{Random}-Hybrid is like \textsc{SalKG}-Hybrid, but with each $y_\text{h}(x, \mathcal{G})$ uniformly sampled from $\{0, 1\}$ as well as using \textsc{Random}-Fine instead of \textsc{SalKG}-Fine.

\textbf{\textsc{Heuristic}~}
Each $\mathcal{G}$ has three node types: question nodes (\textit{i.e.}, nodes in $q$), answer nodes (\textit{i.e.}, nodes in $a_i$), and intermediate nodes (\textit{i.e.}, other nodes) \cite{lin2019kagnet}.
Let QA nodes be nodes in $q$ or $a_i$.
% We hypothesize that question and answer nodes (\textit{i.e.}, QA nodes) are most relevant to answering questions.
\textsc{Heuristic} is a variant of \textsc{SalKG} where each unit's explanation is based on the presence of QA nodes in $\mathcal{G}$.
Let $\bar{N}$ be the mean number of QA nodes per KG (in train set), and let $N(\mathcal{G})$ be the number of QA nodes in $\mathcal{G}$.
\textsc{Heuristic}-Coarse is like \textsc{SalKG}-Coarse, except $y_\text{c}(x, \mathcal{G}) = 1$ if and only if $N(\mathcal{G}) > \bar{N}$.
\textsc{Heuristic}-Fine is like \textsc{SalKG}-Fine, but how $y_\text{f}(u; x, \mathcal{G})$ is set depends on whether the fine units are nodes or paths.
For node units, $y_\text{f}(u; x, \mathcal{G}) = 1$ if and only if $u$ is a QA node.
For path units, $y_\text{f}(u; x, \mathcal{G}) = 1$ if and only if $u$ consists only of QA nodes.
\textsc{Heuristic}-Hybrid is like \textsc{SalKG}-Hybrid, but with $y_\text{h}(x, \mathcal{G}) = 1$ if and only if $N(\mathcal{G}) > \bar{N}$, while \textsc{Heuristic}-Fine is used instead of \textsc{SalKG}-Fine.

\vspace{-0.2cm}
\subsection{Main Results}
\label{sec:exp_main}
\vspace{-0.2cm}

% \textbf{Main Results~}
Table \ref{tab:csqa} shows performance on CSQA, while Table \ref{tab:obqa_codah} shows performance on OBQA and CODAH.
Best performance is highlighted in \colorbox{lgreen}{green}, second-best performance is highlighted in \colorbox{lblue}{blue}, and best non-\textsc{SalKG} performance is highlighted in \colorbox{lred}{red} (if it is not already green or blue).
For \textsc{SalKG} (unlike \textsc{Oracle}), we find that Occl usually outperforms Grad, so we only report Occl performance in Tables \ref{tab:csqa}-\ref{tab:obqa_codah}. 
For a comparison of Grad and Occl on \textsc{SalKG}, see Table \ref{tab:ablation} and Sec. \ref{sec:app_grad_vs_occl}.
Being an ensemble, No-KG + KG tends to beat both No-KG and KG if both have similar performance. Otherwise, No-KG + KG's performance is in between No-KG's and KG's.

\begin{table*}[t]
% \vspace{-1.1cm}
\centering
\vspace{-0.2cm}
\scalebox{0.70}{
\begin{tabular}{lcccccc}
    \toprule &
    \multicolumn{6}{c}{ \textbf{CSQA Test Accuracy (\%)}}\\
    \cmidrule(lr){2-7}
    & \multicolumn{2}{c}{ \textbf{MHGRN}} & \multicolumn{2}{c}{ \textbf{PathGen}} & \multicolumn{2}{c}{ \textbf{RN}} \\
    \cmidrule(lr){2-3}
    \cmidrule(lr){4-5}
    \cmidrule(lr){6-7}
    \textbf{Model} & BERT & RoBERTa & BERT & RoBERTa & BERT & RoBERTa \\
    \midrule
    % {No-KG} & 55.44 & 70.59 & 55.44 & 70.59 & 55.44 & 70.59 \\
    % {KG} & 56.57 & 73.33 & 56.65 & 72.04 & 55.60 & 71.07 \\
    {No-KG} & 53.13 ($\pm$2.34) & 69.65 ($\pm$1.06) & 53.13 ($\pm$2.34) & 69.65 ($\pm$1.06) & 53.13 ($\pm$2.34) & 69.65 ($\pm$1.06) \\
    {KG} & \cellcolor{lred}57.48 ($\pm$0.89) & \cellcolor{lred}73.14 ($\pm$0.78) & 56.54 ($\pm$0.73) & \cellcolor{lred}72.58 ($\pm$0.57) & \cellcolor{lred}56.46 ($\pm$1.22) & \cellcolor{lred}71.37 ($\pm$1.20) \\
    {No-KG + KG} & 56.14 ($\pm$2.28) & 72.15 ($\pm$0.67) & \cellcolor{lred}57.29 ($\pm$1.30) & 72.44 ($\pm$0.72) & 55.98 ($\pm$1.98) & 71.15 ($\pm$0.81) \\
    % {No-KG + KG} & - & - & - & - & - & - \\
    %{$\max(\text{No-KG}, \text{KG})$} & 56.57 & 73.33 & 56.65 & 72.04 & 55.60 & 71.07 \\
    % {KG} & 58.77 ($\pm$0.49) & 73.14 ($\pm$0.78) & 56.54 ($\pm$0.73) & 72.58 ($\pm$0.57) & 56.46 ($\pm$1.22) & 71.37 ($\pm$1.20) \\
    \midrule
    {\textsc{Random}-Coarse} & 55.04 ($\pm$1.44) & 71.06 ($\pm$1.09) & 55.09 ($\pm$1.08) & 71.15 ($\pm$1.06) & 55.15 ($\pm$1.23) & 69.06 ($\pm$2.96) \\
    % {\textsc{Random}-Fine} & 44.75 ($\pm$0.91) & 71.36 ($\pm$0.58) & 56.54 ($\pm$2.02) & 69.59 ($\pm$3.63) & 50.85 ($\pm$1.07) & 68.79 ($\pm$0.17) \\
    {\textsc{Random}-Fine} & 54.69 ($\pm$2.54) & 73.09 ($\pm$1.06) & 54.66 ($\pm$0.97) & 71.26 ($\pm$3.19) & 49.88 ($\pm$1.75) & 69.08 ($\pm$1.95) \\
    {\textsc{Random}-Hybrid} & 52.43 ($\pm$2.60) & 71.93 ($\pm$0.77) & 55.24 ($\pm$0.58) & 71.35 ($\pm$0.34) & 54.36 ($\pm$0.35) & 70.12 ($\pm$0.35) \\
    % {\textsc{SalKG}-Coarse (Grad)} & - & - & - & - & - & - \\
    % {\textsc{SalKG}-Coarse (Occl)} & - & - & - & - & - & - \\
    \midrule
    % {\textsc{SalKG}-Fine (Grad)} & \textcolor{red}{55.44} ($\pm$1.22) & \textcolor{red}{72.95} ($\pm$1.44) & \textcolor{red}{57.10} ($\pm$0.81) & \textcolor{red}{70.10} ($\pm$0.28) & \textcolor{red}{56.14} ($\pm$1.97) & 72.12 ($\pm$0.14) \\
    {\textsc{Heuristic}-Coarse} & 55.55 ($\pm$2.29) & 72.15 ($\pm$0.84) & 56.92 ($\pm$0.18) & 72.57 ($\pm$0.49) & 56.42 ($\pm$1.11) & 71.18 ($\pm$0.77) \\
    {\textsc{Heuristic}-Fine} & 52.54 ($\pm$1.67) & 71.50 ($\pm$1.01) & 54.00 ($\pm$1.89) & 71.11 ($\pm$0.93) & 52.04 ($\pm$2.13) & 65.08 ($\pm$3.67) \\
    {\textsc{Heuristic}-Hybrid} & 56.35 ($\pm$0.81) & 72.58 ($\pm$0.32) & 56.83 ($\pm$0.48) & 71.33 ($\pm$0.87) & 54.38 ($\pm$3.30) & 65.07 ($\pm$2.02) \\
    \midrule
    % {\textsc{SalKG}-Hybrid (Grad)} & 59.07 ($\pm$0.56) & \textcolor{red}{72.79} ($\pm$0.20) & 57.53 ($\pm$0.43) & \textcolor{red}{71.39} ($\pm$0.14) & 57.29 ($\pm$0.29) & 71.98 ($\pm$ 0.28) \\
    {\textsc{SalKG}-Coarse} & \cellcolor{lblue}{57.98 ($\pm$0.90)} & \cellcolor{lgreen}{73.64 ($\pm$1.05)} & \cellcolor{lblue}{57.75 ($\pm$0.77)} & \cellcolor{lgreen}73.07 ($\pm$0.25) & \cellcolor{lblue}57.50 ($\pm$1.25) & \cellcolor{lblue}73.11 ($\pm$1.13) \\
    
    % {\textsc{SalKG}-Fine} & 56.78 ($\pm$2.14) & \cellcolor{lblue}73.65 ($\pm$0.21) & 57.64 ($\pm$2.12) & 71.39 ($\pm$1.54) & 56.86 ($\pm$0.41) & 71.58 ($\pm$1.10) \\
    
    {\textsc{SalKG}-Fine} & 54.36 ($\pm$2.34) & 70.00 ($\pm$0.81) & 54.39 ($\pm$2.03) & 72.12 ($\pm$0.91) & 54.30 ($\pm$1.41) & 71.64 ($\pm$1.51) \\
    
    % {\textsc{SalKG}-Fine (Occl Top-10\%)} & \textcolor{red}{56.78} ($\pm$2.14) & 73.65 ($\pm$0.21) & 57.64 ($\pm$2.12) & \textcolor{red}{71.39} ($\pm$1.54) & 56.86 ($\pm$0.41) & 71.58 ($\pm$1.10) \\
    
    % {\textsc{SalKG}-Fine (Occl Top-30\%)} & - & - & - & 69.54 & - & - \\
    
    % {\textsc{SalKG}-Fine (Occl Top-50\%)} & - & - & - & 70.99 & - & - \\
    
    {\textsc{SalKG}-Hybrid} & \cellcolor{lgreen}58.70 ($\pm$0.65) & \cellcolor{lblue}73.37 ($\pm$0.12) & \cellcolor{lgreen}59.87 ($\pm$0.42) & \cellcolor{lblue}72.67 ($\pm$0.65) & \cellcolor{lgreen}58.78 ($\pm$0.14) & \cellcolor{lgreen}74.13 ($\pm$0.71) \\
    
    % {\textsc{SalKG}-Hybrid (Occl Top-10\%)} & 59.12 ($\pm$0.28) & \textcolor{red}{73.41} ($\pm$0.16) & 60.35 ($\pm$0.32) & \textcolor{red}{72.58} ($\pm$0.23) & 58.80 ($\pm$0.19) & 74.64 ($\pm$0.09) \\
    
    % {\textsc{SalKG}-Hybrid (Occl Top-30\%)} & - & - & - & - & - & - \\
    
    % {\textsc{SalKG}-Hybrid (Occl Top-50\%)} & - & - & - & - & - & - \\
    
    % \midrule

    % {Best \textsc{Oracle}} & 95.89 & 98.63 & 90.49 & 96.78 & 92.26 & 95.25 \\

    \bottomrule 
\end{tabular}
}
% \vspace{-1mm}
\caption{\small \textbf{\textsc{SalKG} Performance on CSQA}
% \xiang{ seems that “Occl” is mostly better than “Grad” --- so we could prioritize tuning Occl methods; and at the end we can consider moving parts of the Grad vs. Occl results to an ablation study (on some settings)}
}
\label{tab:csqa}
% \vspace{-0.3cm}
\end{table*}

Across all datasets, we find that \textsc{SalKG}-Hybrid and \textsc{SalKG}-Coarse are consistently the two best models.
On CSQA, \textsc{SalKG}-Hybrid has the highest performance on BERT+MHGRN, BERT+PathGen, BERT+RN, and RoBERTa+RN, while \textsc{SalKG}-Coarse is the best on RoBERTa+MHGRN and RoBERTa+PathGen.
In particular, on RoBERTa+RN, BERT+RN, and BERT+PathGen, \textsc{SalKG}-Hybrid beats $\max$(No-KG, KG, No-KG + KG) by large margins of 2.76\%, 2.58\%, and 2.32\%, respectively.
Meanwhile, OBQA and CODAH, \textsc{SalKG} is not as dominant but still yields improvements overall.
On OBQA, \textsc{SalKG}-Coarse is the best on RoBERTa+RN (beating $\max$(No-KG, KG, No-KG + KG) by 1.89\%) and RoBERTa+PathGen, while \textsc{SalKG}-Hybrid performs best on RoBERTa+MHGRN.
On CODAH, \textsc{SalKG}-Coarse gets the best performance on both RoBERTa+MHGRN (beating $\max$(No-KG, KG, No-KG + KG) by 1.71\%) and RoBERTa+PathGen.
\textsc{SalKG}-Coarse outperforming \textsc{SalKG}-Hybrid on OBQA and CODAH indicates that local KG supervision from fine explanations may not be as useful for these two datasets.
On the other hand, \textsc{SalKG}-Fine is consistently weaker than \textsc{SalKG}-Hybrid and \textsc{SalKG}-Coarse, but still shows slight improvement for RoBERTa+RN on CSQA.
These results show that learning from KG saliency explanations is generally effective for improving KG-augmented models' performance, especially in CSQA when both coarse and fine explanations are used to provide complementary learning signals for \textsc{SalKG}-Hybrid.
Furthermore, across all datasets, we find that \textsc{SalKG} outperforms \textsc{Random} and \textsc{Heuristic} on every setting.
This is evidence that explanations created from saliency methods can provide better learning signal than those created randomly or from simple heuristics.

\begin{table*}[t]
% \vspace{-1.1cm}
\centering
% \vspace{-0.2cm}
\scalebox{0.70}{
\begin{tabular}{lccccc}
    \toprule &
    \multicolumn{3}{c}{ \textbf{OBQA Test Accuracy (\%)}} & \multicolumn{2}{c}{ \textbf{CODAH Test Accuracy (\%)}}\\
    \cmidrule(lr){2-4} \cmidrule(lr){5-6}
    \textbf{Model (RoBERTa)} & \textbf{MHGRN} & \textbf{PathGen} & \textbf{RN} & \textbf{MHGRN} & \textbf{PathGen} \\
    \midrule
    {No-KG} & 68.73 ($\pm$0.31) & 68.73 ($\pm$0.31) & 68.73 ($\pm$0.31) & 83.96 ($\pm$0.79) & 83.96 ($\pm$0.79) \\
    {KG} & 68.87 ($\pm$2.16) & 68.40 ($\pm$1.59) & 66.80 ($\pm$4.73) & 84.02 ($\pm$1.27) & 84.02 ($\pm$1.62) \\
    {No-KG + KG} & 68.53 ($\pm$0.95) & \cellcolor{lred}69.67 ($\pm$1.45) & \cellcolor{lred}69.40 ($\pm$0.35) & \cellcolor{lred}{84.08 ($\pm$1.46)} & \cellcolor{lblue}84.69 ($\pm$1.48) \\
   
    \midrule

    {\textsc{Random}-Coarse} & 68.11 ($\pm$1.12) & 67.18 ($\pm$4.13) & 65.02 ($\pm$2.57) & 83.48 ($\pm$0.91) & 84.68 ($\pm$1.65) \\
    {\textsc{Random}-Fine} & 57.60 ($\pm$5.33) & 55.13 ($\pm$7.00) & 48.53 ($\pm$4.82) & 74.77 ($\pm$6.90) & 80.48 ($\pm$1.23) \\
    {\textsc{Random}-Hybrid} & 68.33 ($\pm$0.40) & 69.53 ($\pm$0.31) & 69.27 ($\pm$0.12) & 83.86 ($\pm$0.69) & 83.75 ($\pm$0.60) \\

    \midrule
    
    {\textsc{Heuristic}-Coarse} & \cellcolor{lred}69.24 ($\pm$2.47) & 65.58 ($\pm$6.08) & 64.29 ($\pm$3.06) & 82.64 ($\pm$0.10) & 82.52 ($\pm$0.18) \\
    {\textsc{Heuristic}-Fine} & 57.27 ($\pm$3.76) & 51.80 ($\pm$2.95) & 50.53 ($\pm$3.51) & 82.25 ($\pm$1.43) & 82.55 ($\pm$2.03) \\
    {\textsc{Heuristic}-Hybrid} & 68.47 ($\pm$0.23) & 68.40 ($\pm$0.00) & 68.60 ($\pm$0.20) & 82.16 ($\pm$2.11) & 82.73 ($\pm$1.51) \\
    
    \midrule

    {\textsc{SalKG}-Coarse} & \cellcolor{lblue}69.93 ($\pm$0.56) & \cellcolor{lgreen}70.02 ($\pm$0.55) & \cellcolor{lgreen}71.29 ($\pm$0.57) & \cellcolor{lgreen}85.79 ($\pm$1.83) & \cellcolor{lgreen}85.43 ($\pm$1.88) \\
    
    % different top-k for OBQA PathGen and RN
    % {\textsc{SalKG}-Fine} & 65.07 ($\pm$1.70) & 63.27 ($\pm$3.00) & 64.13 ($\pm$5.06) & 84.38 ($\pm$1.02) & 84.38 ($\pm$0.92) \\
    {\textsc{SalKG}-Fine} & 64.82 ($\pm$0.97) & 51.51 ($\pm$0.87) & 62.29 ($\pm$0.85) & 84.08 ($\pm$1.14) & 83.36 ($\pm$0.81) \\

    {\textsc{SalKG}-Hybrid} & \cellcolor{lgreen}70.20 ($\pm$0.69) & \cellcolor{lblue}69.80 ($\pm$0.49) & \cellcolor{lblue}70.47 ($\pm$0.91) & \cellcolor{lblue}85.17 ($\pm$0.54) &  84.42 ($\pm$0.64) \\
    
    \bottomrule 
\end{tabular}
}
% \vspace{-1mm}
\caption{\small \textbf{\textsc{SalKG} Performance on OBQA and CODAH 
% \xiang{we need significance testing vs. 2nd best methods}
}
}
\label{tab:obqa_codah}
\vspace{-0.3cm}
\end{table*}

\begin{wraptable}{R}{0.45\textwidth}
% \begin{table}[t]
% \vspace{-1.1cm}
\centering
\vspace{-0.4cm}

\scalebox{0.70}{
\begin{tabular}{lc}
    \toprule
    \textbf{Model (RoBERTa)} & \textbf{CSQA Test Accuracy (\%)} \\
   
    \midrule
    
    {RN \cite{santoro2017simple}} & 70.08 ($\pm$0.21) \\
    {RN + Link Prediction \cite{wang2020connecting}} & 69.33 ($\pm$0.98) \\
    {RGCN \cite{schlichtkrull2018modeling}} & 68.41 ($\pm$0.66) \\
    {GAT \cite{velivckovic2017graph}} & 71.20 ($\pm$0.72) \\
    {GN \cite{battaglia2018relational}} & 71.12 ($\pm$0.45) \\
    {GconAttn \cite{wang2019improving}} & 69.88 ($\pm$0.47) \\
    {MHGRN \cite{feng2020scalable}} & 71.11 ($\pm$0.81) \\
    {PathGen \cite{wang2020connecting}} & 72.68 ($\pm$0.42) \\

    \midrule

    {\textsc{SalKG}-Coarse (MHGRN)} & \textbf{74.01} ($\pm$0.14) \\
    
    {\textsc{SalKG}-Fine (MHGRN)} & 72.68 ($\pm$1.46) \\

    {\textsc{SalKG}-Hybrid (MHGRN)} & \textbf{73.87} ($\pm$0.48) \\
    
    {\textsc{SalKG}-Coarse (PathGen)} & \textbf{72.76} ($\pm$0.12) \\
    
    {\textsc{SalKG}-Fine (PathGen)} & 71.21 ($\pm$1.31) \\

    {\textsc{SalKG}-Hybrid (PathGen)} & \textbf{73.03} ($\pm$0.84) \\
    
    \bottomrule 
\end{tabular}
}
% \vspace{0.2cm}
\caption{\small \textbf{Comparison of \textsc{SalKG} to Published CSQA Baselines.
} \textsc{SalKG} models that outperform all baselines are shown in \textbf{bold}.
}
\label{tab:csqa_pub}
\vspace{-0.6cm}
% \end{table}
\end{wraptable}

\paragraph{Comparison to Published CSQA Baselines}
To further demonstrate that \textsc{SalKG} models perform competitively, we also compare \textsc{SalKG} (using MHGRN and PathGen) to the many KG-augmented model baseline results published in \cite{feng2020scalable, wang2020connecting, yan2020learning}, for the CSQA in-house split.
The baselines we consider are RN \cite{santoro2017simple}, RN + Link Prediction \cite{feng2020scalable}, RGCN \cite{schlichtkrull2018modeling}, GAT \cite{velivckovic2017graph}, GN \cite{battaglia2018relational}, GconAttn \cite{wang2019improving}, MHGRN \cite{feng2020scalable}, and PathGen \cite{wang2020connecting}.
For the non-\textsc{SalKG} versions of MHGRN, PathGen, and RN, we quote the published results.
Since these published results average over four seeds (instead of three), we report \textsc{SalKG} results over four seeds in Table \ref{tab:csqa_pub}.
We find that most of the listed \textsc{SalKG} variants can outperform all of the baselines. 
For MHGRN, \textsc{SalKG}-Coarse (MHGRN) performs the best overall, \textsc{SalKG}-Hybrid (MHGRN) beats vanilla MHGRN, and \textsc{SalKG}-Fine (MHGRN) is on par with vanilla MHGRN. 
For PathGen, \textsc{SalKG}-Hybrid (PathGen) and \textsc{SalKG}-Coarse (PathGen) both slightly outperform vanilla PathGen, while \textsc{SalKG}-Fine (PathGen) performs worse.

\paragraph{CSQA Leaderboard Submission}
In addition to our experiments on the CSQA in-house split, we evaluated \textsc{SalKG} on the CSQA official split by submitting \textsc{SalKG} to the CSQA leaderboard. Since the best models on the CSQA leaderboard use the ALBERT \cite{lan2019albert} text encoder, and PathGen was the highest graph encoder on the leaderboard out of the three we experimented with, we trained \textsc{SalKG}-Hybrid (ALBERT+PathGen), which achieved a test accuracy of 75.9\%. For reference, a previously submitted ALBERT+PathGen achieved a test accuracy of 75.6\% on the CSQA leaderboard. This result suggests that the proposed \textsc{SalKG} training procedure can yield some improvements over baselines that do not use explanation-based regularization.

\paragraph{\textit{Why does \textsc{SalKG}-Fine perform poorly?}}
In general, \textsc{SalKG}-Fine does not perform as well as \textsc{SalKG}-Coarse and \textsc{SalKG}-Hybrid. Often, \textsc{SalKG}-Fine is noticeably worse than KG and No-KG. Recall that the KG model and \textsc{SalKG}-Fine model both assume that the KG should always be used to solve the given instance. Still, the success of \textsc{SalKG}-Coarse shows that the KG sometimes may not be useful. But why does \textsc{SalKG}-Fine almost always perform worse than the KG model?

We believe it is because \textsc{SalKG}-Fine is more committed to the flawed assumption of universal KG usefulness. Whereas the KG model is trained to solve the task always using the KG as context, SalKG-Fine is trained to both solve the task always using the KG as context (\textit{i.e.}, global KG supervision) and attend to specific parts of the KG (\textit{i.e.}, local KG supervision). Since \textsc{SalKG}-Fine is trained with both global and local KG supervision, it is much more likely to overfit, as the KG is not actually useful for all instances. That is, for training instances where the KG should not be used, \textsc{SalKG}-Fine is pushed to not only use the KG, but also to attend to specific parts of the KG. This leads to a SalKG-Fine model that does not generalize well to test instances where the KG is not useful.

To address this issue, we proposed the \textsc{SalKG}-Hybrid model, which is designed to take the best of both \textsc{SalKG}-Coarse and \textsc{SalKG}-Fine. For a given instance, \textsc{SalKG}-Hybrid uses its \textsc{SalKG}-Coarse component to predict whether the KG is useful, then uses its \textsc{SalKG}-Fine component to attend to the useful parts of the KG only if the KG is predicted to be useful.
Indeed, we find that \textsc{SalKG}-Hybrid performs much better than \textsc{SalKG}-Fine and is the best model overall on CSQA. These results support our hypothesis about why \textsc{SalKG}-Fine performs relatively poorly.

\vspace{-0.1cm}
\subsection{Ablation Studies}
\label{sec:experiments_ablation}
\vspace{-0.1cm}

\begin{wraptable}{R}{0.45\textwidth}
% \begin{table}[t]
\vspace{-1cm}
\centering
\scalebox{0.70}{
\begin{tabular}{lcc}
    \toprule &
    \multicolumn{2}{c}{ \textbf{CSQA Dev Accuracy (\%)}}\\
    \cmidrule(lr){2-3}
    \textbf{Model (BERT)} & \multicolumn{1}{c}{ \textbf{MHGRN}} & \multicolumn{1}{c}{ \textbf{PathGen}}\\
    \midrule
    % {No-KG} & 55.44 & 55.44 \\
    % {KG} & 56.57 & 56.65 \\
    % {No-KG} & 57.25 & 57.25 \\
    % {KG} & 58.80 & 61.26 \\
    % {No-KG + KG} & 59.21 & 59.38 \\
    {\textsc{SalKG}-Coarse} & \textbf{59.49} ($\pm$0.05) & \textbf{60.72} ($\pm$0.58) \\
    {- w/ Grad} & 56.84 ($\pm$2.27) & 56.18 ($\pm$2.31) \\
    {- w/ Occl} & 57.60 ($\pm$0.74) & 56.32 ($\pm$1.66) \\
    % {\textsc{SalKG}-Coarse (Grad)} & 56.84 ($\pm$2.27) & 56.18 ($\pm$2.31) \\
    % {\textsc{SalKG}-Coarse (Occl)} & 57.60 ($\pm$0.74) & 56.32 ($\pm$1.66) \\
    
    % \midrule
    % {\textsc{SalKG}-Coarse} & \textbf{59.49} ($\pm$0.05) & \textbf{60.72} ($\pm$0.58) \\
    % {- w/o Pos. Upweight} & 56.73 ($\pm$0.28) & 60.33 ($\pm$0.45) \\
    
    % {\textsc{SalKG}-Fine (Grad)} & \textcolor{red}{56.40} ($\pm$0.96) & \textcolor{red}{57.25} ($\pm$2.65) \\
    \midrule
    {\textsc{SalKG}-Fine (Occl)} & \textbf{57.28} ($\pm$0.95) & \textbf{59.13} ($\pm$2.35) \\
    {- w/ Grad} & 56.05 ($\pm$1.03) & 58.80 ($\pm$1.08) \\
    
    \midrule
    {\textsc{SalKG}-Hybrid (Occl)} & 59.92 ($\pm$0.31) & \textbf{60.88} ($\pm$0.05) \\
    {- w/ Grad} & \textbf{60.17} ($\pm$0.21) & 59.71 ($\pm$0.08) \\
    
    \midrule
    {\textsc{SalKG}-Fine (Occl)} & \textbf{57.28} ($\pm$0.95) & \textbf{59.13} ($\pm$2.35) \\
    % \midrule
    % {\textsc{SalKG}-Coarse (Random)} & 58.21 ($\pm$0.61) & 58.83 ($\pm$0.19) \\
    % {\textsc{SalKG}-Coarse (Heuristic)} & 58.42 ($\pm$0.64) & 59.30 ($\pm$0.08) \\
    % {\textsc{SalKG}-Fine (Random)} & 49.28 ($\pm$2.02) & 52.72 ($\pm$0.98) \\
    % {\textsc{SalKG}-Fine (Heuristic)} & 55.99 ($\pm$1.19) & 55.39 ($\pm$0.61) \\
    {- w/ Random Prune} & 50.61 ($\pm$0.68) & 54.10 ($\pm$2.13) \\
    {- w/ Heuristic Prune} & 50.72 ($\pm$0.46) & 50.53 ($\pm$0.74) \\
    \midrule
    % {\textsc{SalKG}-Coarse (Grad)} & - & 61.12 ($\pm$0.09) \\
    % {\textsc{SalKG}-Coarse (Goccl)} & 59.30 ($\pm$0.16) & 61.29 ($\pm$0.29) \\
    % {\textsc{SalKG}-Fine (Grad Pipeline)} & - & 54.68 ($\pm$1.44) \\
    % {\textsc{SalKG}-Fine (Occl Pipeline)} & - & - \\
    % \midrule
    
    % {\textsc{SalKG}-Coarse (Pos. Weight)} & 56.73 ($\pm$0.28) & 60.33 ($\pm$0.45) \\
    % \midrule
    {\textsc{SalKG}-Fine (Occl)} & \textbf{57.28} ($\pm$0.95) & \textbf{59.13} ($\pm$2.35) \\
    {- w/ BCE Sal. Loss} & 50.83 ($\pm$1.75) & 55.15 ($\pm$2.58) \\
    % {No-KG} & 55.44 & 55.44 \\
    % {KG} & 56.57 & 56.65 \\
    % {\textsc{SalKG}-Coarse} & - & 58.23 ($\pm$0.82) \\
    % {\textsc{SalKG}-Fine} & - & 53.34 ($\pm$1.75) \\
    % \midrule
    % {\textsc{SalKG}-Coarse (Random)} & - & - \\
    % {\textsc{SalKG}-Coarse (Grad)} & - & - \\
    % {\textsc{SalKG}-Coarse (Occl)} & - & - \\
    % {\textsc{SalKG}-Fine (Random)} & - & - \\
    % \midrule
    % {\textsc{SalKG}-Coarse (Pipeline)} & - & 56.14 ($\pm$0.44)\\
    % {\textsc{SalKG}-Fine (Pipeline)} & - & 52.81 ($\pm$2.36)\\
    % \midrule
    % {\textsc{SalKG}-Coarse (No Pos. Weight)} & - & 55.44 ($\pm$0.00)\\
    % \midrule
    % {\textsc{SalKG}-Fine (BCE)} & - & - \\
    
    \bottomrule
    % {\textsc{SalKG}-Coarse (Self-Attention)} & - & 58.37 ($\pm$0.09)\\
    % \midrule

\end{tabular}
    }
% \vspace{-0.1cm}
\caption{\small \textbf{Ablation Studies.} Best model in \textbf{bold}.
}
\label{tab:ablation}
\vspace{-0.5cm}
% \end{table}
\end{wraptable}

In Table \ref{tab:ablation}, we validate our \textsc{SalKG} design choices with ablation studies. We report dev accuracy for BERT+MHGRN and BERT+PathGen on CSQA.

% In the top section, we again show the baseline models' performance. In the middle section, we show the performance of \textsc{SalKG}-Coarse and its ablation variants. In the bottom section, we show the performance of \textsc{SalKG}-Fine and its ablation variants. Each ablation variant is denoted with its ablation name in parentheses.

% \paragraph{Joint Training of $\mathcal{S}_{\text{KG}}$ and $\mathcal{T}_{\text{KG}}$}
% % For both \textsc{SalKG}-Coarse and \textsc{SalKG}-Fine, we train $\mathcal{S}_{\text{KG}}$ and $\mathcal{T}_{\text{KG}}$ jointly by default. 
% For \textsc{SalKG}-Coarse, we train $\mathcal{S}_{\text{KG}}$ and $\mathcal{T}_{\text{KG}}$ jointly by default.
% For \textsc{SalKG}-Coarse, $\mathcal{T}_{\text{KG}}$ is just a weighted sum function, so joint training means $\mathcal{S}_{\text{KG}}$ is trained using the CSR task loss with respect to $\mathcal{T}_{\text{KG}}$'s output. 
% Alternatively, we could train $\mathcal{S}_{\text{KG}}$ and $\mathcal{T}_{\text{KG}}$ sequentially in a pipeline, such that $\mathcal{S}_{\text{KG}}$ and $\mathcal{T}_{\text{KG}}$ do not share any learning signal. In Table \ref{tab:ablation}, we see that \textsc{SalKG}-Coarse greatly outperforms \textsc{SalKG}-Coarse (Pipeline) by 2.09\%, while \textsc{SalKG}-Fine slightly outperforms \textsc{SalKG}-Fine (Pipeline) by 0.53\%.
% This suggests that $\mathcal{S}_{\text{KG}}$ is best trained using both the CSR task loss and saliency prediction loss.

\textbf{\textit{Are ensemble-based coarse explanations effective?}~}
By default, \textsc{SalKG}-Coarse uses our proposed ensemble-based coarse explanations (Sec. \ref{sec:explanations_coarse}).
Alternatively, we consider using Grad and Occl to create coarse explanations.
For Grad, we compute $\phi$ the same way as in Sec. \ref{sec:explanations_fine}, except using graph embedding $\mathbf{g}$ instead of node/path embeddings.
Since a zero vector would have zero gradient, this is equivalent to comparing $\mathbf{g}$ to a zero vector baseline.
For Occl, we compute $\phi$ as the decrease in $p_{\text{KG}}$ if $\mathbf{g}$ is replaced with a zero vector.
For both Grad and Occl, we set $s_{\text{c}} = \phi$.
In Table \ref{tab:ablation}, we see that our default \textsc{SalKG}-Coarse significantly outperforms \textsc{SalKG}-Coarse with both Grad and Occl.
In Sec. \ref{sec:app_coarse}, we further discuss why Grad and Occl are ill-suited for creating coarse explanations.

\textbf{\textit{For \textsc{SalKG}, is Occl better than Grad?}~}
In Tables \ref{tab:csqa}-\ref{tab:obqa_codah}, we report \textsc{SalKG}-Fine and \textsc{SalKG}-Hybrid performance with Occl fine explanations.
In Table \ref{tab:ablation}, we compare Occl and Grad on \textsc{SalKG}-Fine and \textsc{SalKG}-Hybrid.
Overall, Occl slightly outperforms Grad, although Grad beats Occl on MHGRN for \textsc{SalKG}-Hybrid.
Their relative performance could also depend on the choice of top-$k$\%, which we plan to explore later.
In Sec. \ref{sec:app_grad_vs_occl}, we further compare Occl and Grad on other settings.

\textbf{\textit{How does} \textsc{SalKG}-Fine\textit{'s soft KG pruning compare to hard KG pruning?}~}
\textsc{SalKG}-Fine does soft pruning of unhelpful fine units via soft attention.
We compare \textsc{SalKG}-Fine to two baselines where the KG is filtered via hard pruning, which cannot be easily incorporated into end-to-end training.
For \textsc{Random} Prune and \textsc{Heuristic} Prune, we respectively create \textsc{Random} and \textsc{Heuristic} explanations, then hard prune all negative units from the KG.
The KG-augmented model then uses the pruned KG as its KG input.
In Table \ref{tab:ablation}, we see that \textsc{SalKG}-Fine significantly outperforms the two baselines, showing the benefits of jointly training the model on saliency and QA prediction.

% \paragraph{Selection of salient units} 
% \paragraph{\textsc{SalKG}-Fine: Pruning $\mathcal{G}$ via Fine Explanations}
% For both Coarse and Fine variants of \textsc{SalKG}, we analyze the effect of choosing the most salient units versus choosing at random or using some simple heuristic. \textsc{SalKG}-Coarse (Random) randomly selects whether to use a KG to augment the PLM or not. While \textsc{SalKG}-Coarse (Heuristic) selects to use a KG if the number of nodes in the KG is greater than the average number of nodes in a KG across all training instances. This simple heuristic is based on a hypothesis that larger KGs probably contain more useful information. Similarly, \textsc{SalKG}-Fine (Random) just randomly selects top-$10\%$ units as salient units. \textsc{SalKG}-Fine (Heuristic) selects the nodes of the KG with positive saliency scores for MHGRN (or correspondingly, positive paths for PathGen). We observe that our \textsc{SalKG} models outperform all the random and heuristic selection algorithms by strong margins. Thus, this shows that our strategies are better than the basic random and heuristic strategies.

% \paragraph{Fine Saliency Loss}
% \paragraph{\textsc{SalKG}-Fine: Saliency Loss Function}
\textbf{\textit{Is it effective to train} \textsc{SalKG}-Fine \textit{with KL divergence?}~}
We train \textsc{SalKG}-Fine's explanation predictor (\textit{i.e.}, attention mechanism) using KL divergence as the saliency loss.
Thus, within a KG, the distribution over attention weights constitutes a single prediction.
Alternatively, we could treat each attention weight as a separate prediction and train the attention mechanism using binary cross entropy (BCE) loss.
In Table \ref{tab:ablation}, we find that using KL divergence yields much higher performance than using BCE loss.
This suggests that the attention weights should not be trained separately, as each attention weight is highly dependent on other attention weights in the same KG.

% In \textsc{SalKG}-Fine, we implement $\mathcal{S}_{\text{KG}}$ as a binary classifier and train $\mathcal{S}_{\text{KG}}$ using binary cross entropy (BCE) loss. By doing so, we treat each saliency unit as an individual saliency instance. Besides, BCE loss, we also tried training $\mathcal{S}_{\text{KG}}$ with KL divergence. Here, the model aims to minimize the KL divergence between the $\mathcal{S}_{\text{KG}}$'s predicted attention weight distribution and the saliency target distribution, which means all saliency units in the KG together form a single saliency instance. 
% We see that \textsc{SalKG}-Fine (KL Divergence) only beats \textsc{SalKG}-Fine by 0.14\%, which is much lower than both models' reported standard deviations. Thus, we conclude that \textsc{SalKG}-Fine (KL Divergence) and \textsc{SalKG}-Fine have comparable performance.

% \paragraph{Choice of Coarse saliency explanation} In this, we explore the effect of using some other saliency \soumya{Get more clarity}

% Ablation 3: It helps to jointly use coarse and fine saliency in pruned QA. (?)
% [QA --> saliency + pruned QA] (coarse)
% [QA --> saliency + pruned QA] (fine)
% [QA --> saliency + pruned QA] (coarse + fine)

\vspace{-0.2cm}
\subsection{Case Studies}
\label{sec:case_studies}
\vspace{-0.2cm}
We visualize coarse/fine explanations created from BERT+PathGen on CSQA, with 1-hop or 2-hop paths as fine units.
For coarse explanations, we show examples of positive (\textit{i.e.}, useful) and negative KGs.
Since KGs are too large to show here, we uniformly sample three paths per KG.
For the positive KG example, the question is \textit{James loved to play violin. He did it in his spare time because he found it what?}, the answer choice is \textit{relaxing}, and the target answer is \textit{relaxing}.
Its paths are: 
\textbf{(1)} \begin{small}\texttt{play} --[is related to]--> \texttt{x} <--[is used for]-- \texttt{relaxing} \end{small}, 
\textbf{(2)} \begin{small}\texttt{violin} --[is used for]--> \texttt{x} --[is used for]--> \texttt{relaxing} \end{small}, and
\textbf{(3)} \begin{small}\texttt{time} <--[has subevent]-- \texttt{x} --[has subevent]--> \texttt{relax} \end{small}.
For the negative KG example, the question is \textit{Where do soldiers not deployed eat their food?}, the answer choice is \textit{neighbor's house}, and the target answer is \textit{military base}.
Its paths are: 
\textbf{(1)} \begin{small}\texttt{soldier} <--[is related to]-- \texttt{x} <--[is related to]-- \texttt{house} \end{small}, 
\textbf{(2)} \begin{small}\texttt{eat} --[is related to]--> \texttt{x} --[is at location of]--> \texttt{house} \end{small}, and
\textbf{(3)} \begin{small}\texttt{food} <--[is related to]-- \texttt{x} --[is at location of]--> \texttt{house} \end{small}.
For fine explanations, we show examples of positive and negative paths from the same KG.
Here, the question is \textit{Where can you find a bar before traveling a long distance?}, the answer choice is \textit{airport}, and the target answer is \textit{airport}.
The positive path is: \begin{small}\texttt{bar} --[is at location]--> \texttt{airport} \end{small}.
The negative path is: \begin{small}\texttt{travel} <--[is used for]-- \texttt{x} --[is at location]-- \texttt{airport} \end{small}.
We can roughly see that the positive KGs/paths are useful for predicting the correct answer, and vice versa.
% In the second example, the question is \textit{What do you feel for a someone when you comfort friend?}, the answer choice is \textit{feeling bad}, and target answer is \textit{care}.
% The positive path is: \begin{small}\texttt{comfort} <--[is the antonym of]-- \texttt{x} --[is the antonym of]--> \texttt{feel} \end{small}.
% The negative path is: \begin{small}\texttt{comfort} --[is at location of]--> \texttt{x} --[is related to]--> \texttt{feeling} \end{small}.
% We can roughly see that positive KG/paths help predict the correct answer choice, and vice versa.
However, as shown in \cite{raman2020learning}, the model's judgment of KG/path usefulness may not always align with human judgment.
See Sec. \ref{sec:app_case_qual} for more illustrative examples of coarse/fine explanations.

% !TEX root = main.tex
\vspace{-0.2cm}
\section{Related Work} 
\label{sec:related}
\vspace{-0.2cm}

\textbf{Creating Model Explanations~}
% \paragraph{Text-Based Explanations}
Many methods aim to explain PLMs' predictions by highlighting important tokens in the model's text input.
% Although some of these works focus on abstractive (free-text) explanations \cite{rajani2019explain, strout2019human, zhao2020lirex}, most aim to provide extractive explanations which highlight salient tokens in the model's text input.
Such methods are usually gradient-based \cite{sundararajan2017axiomatic, li2015visualizing, denil2014extraction}, attention-based \cite{mohankumar2020towards, tutek2020staying, ghaeini2018interpreting, lee2017interactive}, or occlusion-based \cite{deyoung2019eraser, poerner2018evaluating, kadar2017representation, li2016understanding}.
% How feature attribution methods should be chosen remains an open question and the subject of much recent debate \cite{bastings2020elephant, wiegreffe2019attention, serrano2019attention, jain2019attention}.
% \paragraph{Graph-Based Explanations}
% There are also methods proposing extractive explanations for graph encoders, especially GNNs.
Similarly, for graph encoders, a number of works use post-hoc optimization to identify important nodes \cite{huang2020graphlime, ying2019gnnexplainer} or subgraphs \cite{ying2019gnnexplainer} in the graph input. 
% Some GNNs use attention for pooling, which naturally highlights nodes with higher attention weights \cite{lee2019self, lee2018graph}.
% More sophisticated approaches use post-hoc optimization to identify salient nodes \cite{huang2020graphlime, ying2019gnnexplainer} or subgraphs \cite{ying2019gnnexplainer}.
% Meanwhile, KG-augmented models take both text and graph inputs. 
Meanwhile, KG-augmented models' attention weights can be used to explain which parts of the KG are important \cite{lin2019kagnet, feng2020scalable, liu2020commonsense, wang2020connecting, yan2020learning}. 
These KG explanations can be interpreted as identifying knowledge in the KG that is complementary to the knowledge encoded in the PLM. 
% However, there is little work on how the KG-augmented models can learn from such explanations.
% However, there is little work on how such KG explanations should be used. 
% For KG-augmented models, \textsc{SalKG} focuses more on how KG explanations are used rather than created.

% While \textsc{SalKG} also uses saliency methods (e.g., Grad, Occl) to create explanations, our study is limited to explanations regarding KG-augmented models' graph inputs.

\textbf{Learning From Model Explanations~}
Besides manual inspection, explanations can be used in various ways, like extra supervision or regularization \cite{pruthi2020evaluating, hase2020leakage, narang2020wt5, andreas2017learning}, pruned inputs \cite{jain2020learning, bastings2019interpretable, lei2016rationalizing}, additional inputs \cite{hase2021can, co2018guiding}, and intermediate variables \cite{wiegreffe2020measuring, zhou2020towards, rajani2019explain}.
The most similar work to ours is \cite{pruthi2020evaluating}, which proposed training a student model to mimic a teacher model's predictions by regularizing the student model's attention via text explanations created from the teacher model.
However, \cite{pruthi2020evaluating} aims to evaluate explanations, while our goal is to improve performance via explanations.
% Still, methods for learning from explanations have largely focused on domains like text and images, as opposed to graphs. 
To the best of our knowledge, \textsc{SalKG} is the first to supervise KG-augmented models with KG explanations.
% \aaron{Fix description of \cite{pruthi2020evaluating}}

See Sec. \ref{sec:app_related_work} for a more comprehensive overview of the related literature.

% !TEX root = main.tex
\vspace{-0.2cm}
\section{Conclusion}
\label{sec:conclusion}
\vspace{-0.2cm}

In this paper, we proposed creating coarse and fine explanations for KG-augmented models, then using these explanations as extra inputs (\textsc{Oracle}) or supervision (\textsc{SalKG}). 
% and learn from investigated the usage of KG explanations as supervision for KG-augmented models.
% We proposed \textsc{SalKG}, a simple framework for learning from KG explanations of both coarse and fine granularity.
Across three commonsense QA benchmarks, \textsc{SalKG} achieves strong performance, especially when both coarse and fine explanations are used.
% In future work, we plan to explore several directions for improving \textsc{SalKG}-Fine's design as well as ideas for combining \textsc{SalKG}-Coarse and \textsc{SalKG}-Fine into a hybrid framework.
In future work, we plan to explore incorporating active learning into \textsc{SalKG}, so that models can also leverage explanation-based feedback from humans about KG saliency.
% \xiang{can talk about looping in humans to solicit feedback on the salient features of the KG as future work?}
% !TEX root = main.tex
\vspace{-0.2cm}
\section{Acknowledgments}
\label{sec:acknowledgments}
\vspace{-0.2cm}

This research is supported in part by the Office of the Director of National Intelligence (ODNI), Intelligence Advanced Research Projects Activity (IARPA), via Contract No. 2019-19051600007, the DARPA MCS program under Contract No. N660011924033, the Defense Advanced Research Projects Agency with award W911NF-19-20271, NSF IIS 2048211, NSF SMA 1829268, and gift awards from Google, Amazon, JP Morgan, and Sony. We would like to thank all of our collaborators at the USC INK Research Lab for their constructive feedback on this work.

% \section*{Acknowledgements}

\bibliography{references}

\begin{thebibliography}{10}

\bibitem{andreas2017learning}
Jacob Andreas, Dan Klein, and Sergey Levine.
\newblock Learning with latent language.
\newblock {\em arXiv preprint arXiv:1711.00482}, 2017.

\bibitem{bastings2020elephant}
Jasmijn Bastings and Katja Filippova.
\newblock The elephant in the interpretability room: Why use attention as
  explanation when we have saliency methods?
\newblock {\em arXiv preprint arXiv:2010.05607}, 2020.

\bibitem{bastings2019interpretable}
Joost Bastings, Wilker Aziz, and Ivan Titov.
\newblock Interpretable neural predictions with differentiable binary
  variables.
\newblock {\em arXiv preprint arXiv:1905.08160}, 2019.

\bibitem{battaglia2018relational}
Peter~W Battaglia, Jessica~B Hamrick, Victor Bapst, Alvaro Sanchez-Gonzalez,
  Vinicius Zambaldi, Mateusz Malinowski, Andrea Tacchetti, David Raposo, Adam
  Santoro, Ryan Faulkner, et~al.
\newblock Relational inductive biases, deep learning, and graph networks.
\newblock {\em arXiv preprint arXiv:1806.01261}, 2018.

\bibitem{bosselut2019dynamic}
Antoine Bosselut and Yejin Choi.
\newblock Dynamic knowledge graph construction for zero-shot commonsense
  question answering.
\newblock {\em arXiv preprint arXiv:1911.03876}, 2019.

\bibitem{chen2019codah}
Michael Chen, Mike D{'}Arcy, Alisa Liu, Jared Fernandez, and Doug Downey.
\newblock {CODAH}: An adversarially-authored question answering dataset for
  common sense.
\newblock In {\em Proceedings of the 3rd Workshop on Evaluating Vector Space
  Representations for {NLP}}, pages 63--69, Minneapolis, USA, June 2019.
  Association for Computational Linguistics.

\bibitem{chen2017neural}
Qian Chen, Xiaodan Zhu, Zhen-Hua Ling, Diana Inkpen, and Si~Wei.
\newblock Neural natural language inference models enhanced with external
  knowledge.
\newblock {\em arXiv preprint arXiv:1711.04289}, 2017.

\bibitem{co2018guiding}
John~D Co-Reyes, Abhishek Gupta, Suvansh Sanjeev, Nick Altieri, Jacob Andreas,
  John DeNero, Pieter Abbeel, and Sergey Levine.
\newblock Guiding policies with language via meta-learning.
\newblock {\em arXiv preprint arXiv:1811.07882}, 2018.

\bibitem{davis2015commonsense}
Ernest Davis and Gary Marcus.
\newblock Commonsense reasoning and commonsense knowledge in artificial
  intelligence.
\newblock {\em Communications of the ACM}, 58(9):92--103, 2015.

\bibitem{denil2014extraction}
Misha Denil, Alban Demiraj, and Nando De~Freitas.
\newblock Extraction of salient sentences from labelled documents.
\newblock {\em arXiv preprint arXiv:1412.6815}, 2014.

\bibitem{devlin2018bert}
Jacob Devlin, Ming-Wei Chang, Kenton Lee, and Kristina Toutanova.
\newblock {BERT}: Pre-training of deep bidirectional transformers for language
  understanding.
\newblock In {\em Proceedings of NAACL)}, pages 4171--4186, Minneapolis,
  Minnesota, June 2019. Association for Computational Linguistics.

\bibitem{deyoung2019eraser}
Jay DeYoung, Sarthak Jain, Nazneen~Fatema Rajani, Eric Lehman, Caiming Xiong,
  Richard Socher, and Byron~C Wallace.
\newblock Eraser: A benchmark to evaluate rationalized nlp models.
\newblock {\em arXiv preprint arXiv:1911.03429}, 2019.

\bibitem{feng2020scalable}
Yanlin Feng, Xinyue Chen, Bill~Yuchen Lin, Peifeng Wang, Jun Yan, and Xiang
  Ren.
\newblock Scalable multi-hop relational reasoning for knowledge-aware question
  answering.
\newblock {\em arXiv preprint arXiv:2005.00646}, 2020.

\bibitem{ghaeini2018interpreting}
Reza Ghaeini, Xiaoli~Z Fern, and Prasad Tadepalli.
\newblock Interpreting recurrent and attention-based neural models: a case
  study on natural language inference.
\newblock {\em arXiv preprint arXiv:1808.03894}, 2018.

\bibitem{gunning2018machine}
David Gunning.
\newblock Machine common sense concept paper.
\newblock {\em arXiv preprint arXiv:1810.07528}, 2018.

\bibitem{hase2021can}
Peter Hase and Mohit Bansal.
\newblock When can models learn from explanations? a formal framework for
  understanding the roles of explanation data.
\newblock {\em arXiv preprint arXiv:2102.02201}, 2021.

\bibitem{hase2020leakage}
Peter Hase, Shiyue Zhang, Harry Xie, and Mohit Bansal.
\newblock Leakage-adjusted simulatability: Can models generate non-trivial
  explanations of their behavior in natural language?
\newblock {\em arXiv preprint arXiv:2010.04119}, 2020.

\bibitem{honnibal-johnson:2015:EMNLP}
Matthew Honnibal and Mark Johnson.
\newblock An improved non-monotonic transition system for dependency parsing.
\newblock In {\em Proceedings of the 2015 Conference on Empirical Methods in
  Natural Language Processing}, pages 1373--1378, Lisbon, Portugal, September
  2015. Association for Computational Linguistics.

\bibitem{huang2020graphlime}
Qiang Huang, Makoto Yamada, Yuan Tian, Dinesh Singh, Dawei Yin, and Yi~Chang.
\newblock Graphlime: Local interpretable model explanations for graph neural
  networks.
\newblock {\em arXiv preprint arXiv:2001.06216}, 2020.

\bibitem{jain2019attention}
Sarthak Jain and Byron~C Wallace.
\newblock Attention is not explanation.
\newblock {\em arXiv preprint arXiv:1902.10186}, 2019.

\bibitem{jain2020learning}
Sarthak Jain, Sarah Wiegreffe, Yuval Pinter, and Byron~C Wallace.
\newblock Learning to faithfully rationalize by construction.
\newblock {\em arXiv preprint arXiv:2005.00115}, 2020.

\bibitem{kadar2017representation}
Akos K{\'a}d{\'a}r, Grzegorz Chrupa{\l}a, and Afra Alishahi.
\newblock Representation of linguistic form and function in recurrent neural
  networks.
\newblock {\em Computational Linguistics}, 43(4):761--780, 2017.

\bibitem{khot2020qasc}
Tushar Khot, Peter Clark, Michal Guerquin, Peter Jansen, and Ashish Sabharwal.
\newblock {QASC:} {A} dataset for question answering via sentence composition.
\newblock In {\em The Thirty-Fourth {AAAI} Conference on Artificial
  Intelligence, {AAAI} 2020, The Thirty-Second Innovative Applications of
  Artificial Intelligence Conference, {IAAI} 2020, The Tenth {AAAI} Symposium
  on Educational Advances in Artificial Intelligence, {EAAI} 2020, New York,
  NY, USA, February 7-12, 2020}, pages 8082--8090. {AAAI} Press, 2020.

\bibitem{lan2019albert}
Zhenzhong Lan, Mingda Chen, Sebastian Goodman, Kevin Gimpel, Piyush Sharma, and
  Radu Soricut.
\newblock Albert: A lite bert for self-supervised learning of language
  representations.
\newblock {\em arXiv preprint arXiv:1909.11942}, 2019.

\bibitem{lee2017interactive}
Jaesong Lee, Joong-Hwi Shin, and Jun-Seok Kim.
\newblock Interactive visualization and manipulation of attention-based neural
  machine translation.
\newblock In {\em Proceedings of the 2017 Conference on Empirical Methods in
  Natural Language Processing: System Demonstrations}, pages 121--126, 2017.

\bibitem{lee2018graph}
John~Boaz Lee, Ryan Rossi, and Xiangnan Kong.
\newblock Graph classification using structural attention.
\newblock In {\em Proceedings of the 24th ACM SIGKDD International Conference
  on Knowledge Discovery \& Data Mining}, pages 1666--1674, 2018.

\bibitem{lee2019self}
Junhyun Lee, Inyeop Lee, and Jaewoo Kang.
\newblock Self-attention graph pooling.
\newblock In {\em International Conference on Machine Learning}, pages
  3734--3743. PMLR, 2019.

\bibitem{lei2016rationalizing}
Tao Lei, Regina Barzilay, and Tommi Jaakkola.
\newblock Rationalizing neural predictions.
\newblock {\em arXiv preprint arXiv:1606.04155}, 2016.

\bibitem{li2015visualizing}
Jiwei Li, Xinlei Chen, Eduard Hovy, and Dan Jurafsky.
\newblock Visualizing and understanding neural models in nlp.
\newblock {\em arXiv preprint arXiv:1506.01066}, 2015.

\bibitem{li2016understanding}
Jiwei Li, Will Monroe, and Dan Jurafsky.
\newblock Understanding neural networks through representation erasure.
\newblock {\em arXiv preprint arXiv:1612.08220}, 2016.

\bibitem{lin2019kagnet}
Bill~Yuchen Lin, Xinyue Chen, Jamin Chen, and Xiang Ren.
\newblock {K}ag{N}et: Knowledge-aware graph networks for commonsense reasoning.
\newblock In {\em Proceedings of EMNLP-IJCNLP}, pages 2829--2839, Hong Kong,
  China, November 2019. Association for Computational Linguistics.

\bibitem{lin-etal-2020-commongen}
Bill~Yuchen Lin, Wangchunshu Zhou, Ming Shen, Pei Zhou, Chandra Bhagavatula,
  Yejin Choi, and Xiang Ren.
\newblock {C}ommon{G}en: A constrained text generation challenge for generative
  commonsense reasoning.
\newblock In {\em Findings of the Association for Computational Linguistics:
  EMNLP 2020}, pages 1823--1840, Online, November 2020. Association for
  Computational Linguistics.

\bibitem{liu2020kg}
Ye~Liu, Yao Wan, Lifang He, Hao Peng, and Philip~S Yu.
\newblock Kg-bart: Knowledge graph-augmented bart for generative commonsense
  reasoning.
\newblock {\em arXiv preprint arXiv:2009.12677}, 2020.

\bibitem{liu2020commonsense}
Ye~Liu, Tao Yang, Zeyu You, Wei Fan, and Philip~S Yu.
\newblock Commonsense evidence generation and injection in reading
  comprehension.
\newblock {\em arXiv preprint arXiv:2005.05240}, 2020.

\bibitem{liu2019roberta}
Yinhan Liu, Myle Ott, Naman Goyal, Jingfei Du, Mandar Joshi, Danqi Chen, Omer
  Levy, Mike Lewis, Luke Zettlemoyer, and Veselin Stoyanov.
\newblock Roberta: A robustly optimized bert pretraining approach.
\newblock {\em arXiv preprint arXiv:1907.11692}, 2019.

\bibitem{lv2020graph}
Shangwen Lv, Daya Guo, Jingjing Xu, Duyu Tang, Nan Duan, Ming Gong, Linjun
  Shou, Daxin Jiang, Guihong Cao, and Songlin Hu.
\newblock Graph-based reasoning over heterogeneous external knowledge for
  commonsense question answering.
\newblock In {\em Proceedings of the AAAI Conference on Artificial
  Intelligence}, volume~34, pages 8449--8456, 2020.

\bibitem{ma2019towards}
Kaixin Ma, Jonathan Francis, Quanyang Lu, Eric Nyberg, and Alessandro
  Oltramari.
\newblock Towards generalizable neuro-symbolic systems for commonsense question
  answering.
\newblock In {\em Proceedings of the First Workshop on Commonsense Inference in
  Natural Language Processing}, pages 22--32, Hong Kong, China, November 2019.
  Association for Computational Linguistics.

\bibitem{marcus2018deep}
Gary Marcus.
\newblock Deep learning: A critical appraisal.
\newblock {\em arXiv preprint arXiv:1801.00631}, 2018.

\bibitem{mihaylov2018can}
Todor Mihaylov, Peter Clark, Tushar Khot, and Ashish Sabharwal.
\newblock Can a suit of armor conduct electricity? a new dataset for open book
  question answering.
\newblock In {\em Proceedings of the 2018 Conference on Empirical Methods in
  Natural Language Processing}, pages 2381--2391, Brussels, Belgium,
  October-November 2018. Association for Computational Linguistics.

\bibitem{mohankumar2020towards}
Akash~Kumar Mohankumar, Preksha Nema, Sharan Narasimhan, Mitesh~M Khapra,
  Balaji~Vasan Srinivasan, and Balaraman Ravindran.
\newblock Towards transparent and explainable attention models.
\newblock {\em arXiv preprint arXiv:2004.14243}, 2020.

\bibitem{narang2020wt5}
Sharan Narang, Colin Raffel, Katherine Lee, Adam Roberts, Noah Fiedel, and
  Karishma Malkan.
\newblock Wt5?! training text-to-text models to explain their predictions.
\newblock {\em arXiv preprint arXiv:2004.14546}, 2020.

\bibitem{poerner2018evaluating}
Nina Poerner, Benjamin Roth, and Hinrich Sch{\"u}tze.
\newblock Evaluating neural network explanation methods using hybrid documents
  and morphological agreement.
\newblock {\em arXiv preprint arXiv:1801.06422}, 2018.

\bibitem{pruthi2020evaluating}
Danish Pruthi, Bhuwan Dhingra, Livio~Baldini Soares, Michael Collins, Zachary~C
  Lipton, Graham Neubig, and William~W Cohen.
\newblock Evaluating explanations: How much do explanations from the teacher
  aid students?
\newblock {\em arXiv preprint arXiv:2012.00893}, 2020.

\bibitem{rajani2019explain}
Nazneen~Fatema Rajani, Bryan McCann, Caiming Xiong, and Richard Socher.
\newblock Explain yourself! leveraging language models for commonsense
  reasoning.
\newblock {\em arXiv preprint arXiv:1906.02361}, 2019.

\bibitem{raman2020learning}
Mrigank Raman, Aaron Chan, Siddhant Agarwal, Peifeng Wang, Hansen Wang,
  Sungchul Kim, Ryan Rossi, Handong Zhao, Nedim Lipka, and Xiang Ren.
\newblock Learning to deceive knowledge graph augmented models via targeted
  perturbation.
\newblock {\em arXiv preprint arXiv:2010.12872}, 2020.

\bibitem{santoro2017simple}
Adam Santoro, David Raposo, David~G Barrett, Mateusz Malinowski, Razvan
  Pascanu, Peter Battaglia, and Timothy Lillicrap.
\newblock A simple neural network module for relational reasoning.
\newblock In {\em Advances in neural information processing systems}, pages
  4967--4976, 2017.

\bibitem{schlichtkrull2018modeling}
Michael Schlichtkrull, Thomas~N Kipf, Peter Bloem, Rianne Van Den~Berg, Ivan
  Titov, and Max Welling.
\newblock Modeling relational data with graph convolutional networks.
\newblock In {\em European Semantic Web Conference}, pages 593--607. Springer,
  2018.

\bibitem{serrano2019attention}
Sofia Serrano and Noah~A Smith.
\newblock Is attention interpretable?
\newblock {\em arXiv preprint arXiv:1906.03731}, 2019.

\bibitem{speer2017conceptnet}
Robyn Speer, Joshua Chin, and Catherine Havasi.
\newblock Conceptnet 5.5: an open multilingual graph of general knowledge.
\newblock In {\em Proceedings of AAAI}, pages 4444--4451, 2017.

\bibitem{strout2019human}
Julia Strout, Ye~Zhang, and Raymond~J Mooney.
\newblock Do human rationales improve machine explanations?
\newblock {\em arXiv preprint arXiv:1905.13714}, 2019.

\bibitem{sundararajan2017axiomatic}
Mukund Sundararajan, Ankur Taly, and Qiqi Yan.
\newblock Axiomatic attribution for deep networks.
\newblock In {\em International Conference on Machine Learning}, pages
  3319--3328. PMLR, 2017.

\bibitem{talmor2019commonsenseqa}
Alon Talmor, Jonathan Herzig, Nicholas Lourie, and Jonathan Berant.
\newblock {C}ommonsense{QA}: A question answering challenge targeting
  commonsense knowledge.
\newblock In {\em Proceedings of the 2019 Conference of the North {A}merican
  Chapter of the Association for Computational Linguistics: Human Language
  Technologies, Volume 1 (Long and Short Papers)}, pages 4149--4158,
  Minneapolis, Minnesota, June 2019. Association for Computational Linguistics.

\bibitem{tutek2020staying}
Martin Tutek and Jan {\v{S}}najder.
\newblock Staying true to your word:(how) can attention become explanation?
\newblock {\em arXiv preprint arXiv:2005.09379}, 2020.

\bibitem{vaswani2017attention}
Ashish Vaswani, Noam Shazeer, Niki Parmar, Jakob Uszkoreit, Llion Jones,
  Aidan~N Gomez, Lukasz Kaiser, and Illia Polosukhin.
\newblock Attention is all you need.
\newblock {\em arXiv preprint arXiv:1706.03762}, 2017.

\bibitem{velivckovic2017graph}
Petar Veli{\v{c}}kovi{\'c}, Guillem Cucurull, Arantxa Casanova, Adriana Romero,
  Pietro Lio, and Yoshua Bengio.
\newblock Graph attention networks.
\newblock {\em arXiv preprint arXiv:1710.10903}, 2017.

\bibitem{wang2020connecting}
Peifeng Wang, Nanyun Peng, Pedro Szekely, and Xiang Ren.
\newblock Connecting the dots: A knowledgeable path generator for commonsense
  question answering.
\newblock {\em arXiv preprint arXiv:2005.00691}, 2020.

\bibitem{wang2019improving}
Xiaoyan Wang, Pavan Kapanipathi, Ryan Musa, Mo~Yu, Kartik Talamadupula, Ibrahim
  Abdelaziz, Maria Chang, Achille Fokoue, Bassem Makni, Nicholas Mattei, et~al.
\newblock Improving natural language inference using external knowledge in the
  science questions domain.
\newblock In {\em Proceedings of the AAAI Conference on Artificial
  Intelligence}, volume~33, pages 7208--7215, 2019.

\bibitem{wiegreffe2020measuring}
Sarah Wiegreffe, Ana Marasovic, and Noah~A Smith.
\newblock Measuring association between labels and free-text rationales.
\newblock {\em arXiv preprint arXiv:2010.12762}, 2020.

\bibitem{wiegreffe2019attention}
Sarah Wiegreffe and Yuval Pinter.
\newblock Attention is not not explanation.
\newblock {\em arXiv preprint arXiv:1908.04626}, 2019.

\bibitem{yan2020learning}
Jun Yan, Mrigank Raman, Aaron Chan, Tianyu Zhang, Ryan Rossi, Handong Zhao,
  Sungchul Kim, Nedim Lipka, and Xiang Ren.
\newblock Learning contextualized knowledge structures for commonsense
  reasoning.
\newblock {\em arXiv preprint arXiv:2010.12873}, 2020.

\bibitem{yasunaga2021qa}
Michihiro Yasunaga, Hongyu Ren, Antoine Bosselut, Percy Liang, and Jure
  Leskovec.
\newblock Qa-gnn: Reasoning with language models and knowledge graphs for
  question answering.
\newblock {\em arXiv preprint arXiv:2104.06378}, 2021.

\bibitem{ying2019gnnexplainer}
Zhitao Ying, Dylan Bourgeois, Jiaxuan You, Marinka Zitnik, and Jure Leskovec.
\newblock Gnnexplainer: Generating explanations for graph neural networks.
\newblock In {\em Advances in neural information processing systems}, pages
  9244--9255, 2019.

\bibitem{zellers2018swag}
Rowan Zellers, Yonatan Bisk, Roy Schwartz, and Yejin Choi.
\newblock Swag: A large-scale adversarial dataset for grounded commonsense
  inference.
\newblock {\em arXiv preprint arXiv:1808.05326}, 2018.

\bibitem{zhao2020lirex}
Xinyan Zhao and VG~Vydiswaran.
\newblock Lirex: Augmenting language inference with relevant explanation.
\newblock {\em arXiv preprint arXiv:2012.09157}, 2020.

\bibitem{zhou2018commonsense}
Hao Zhou, Tom Young, Minlie Huang, Haizhou Zhao, Jingfang Xu, and Xiaoyan Zhu.
\newblock Commonsense knowledge aware conversation generation with graph
  attention.
\newblock In {\em IJCAI}, pages 4623--4629, 2018.

\bibitem{zhou2020towards}
Wangchunshu Zhou, Jinyi Hu, Hanlin Zhang, Xiaodan Liang, Maosong Sun, Chenyan
  Xiong, and Jian Tang.
\newblock Towards interpretable natural language understanding with
  explanations as latent variables.
\newblock {\em arXiv preprint arXiv:2011.05268}, 2020.

\end{thebibliography}
\bibliographystyle{plain}

\newpage
\appendix
% !TEX root = main.tex
\section{Appendix} 
\label{sec:appendix}

\subsection{Construction of the Contextualized KG}
\label{sec:app_contextualized_kg}
In Sec. \ref{sec:background}, we defined the full KG as $\tilde{\mathcal{G}} = (\tilde{\mathcal{V}}, \tilde{\mathcal{R}}, \tilde{\mathcal{E}})$, where $\tilde{\mathcal{V}}$, $\tilde{\mathcal{R}}$, and $\tilde{\mathcal{E}}$ are all of the KG's nodes (concepts), relations, and edges (facts), respectively. For each instance, we assume access to $\tilde{\mathcal{G}}$ but do not use the entire KG in practice. Given a question $q$ and an answer choice $a_i$ for some instance, we construct the contextualized KG, $\tilde{\mathcal{G}_i} = (\mathcal{V}_i, \mathcal{R}_i, \mathcal{E}_i)$ by heuristically extracting edges from $\tilde{\mathcal{G}}$, following the approach taken by most prior KG-augmented model works \cite{feng2020scalable, wang2020connecting, lin2019kagnet}.

$\tilde{\mathcal{G}_i} = (\mathcal{V}_i, \mathcal{R}_i, \mathcal{E}_i)$ is built differently for node-based models and path-based models, and we describe both types of contextualized KG construction procedures below. Note that these procedures are not designed by us, but simply follow what was proposed and shown to work well in the KG-augmented models' original papers \cite{feng2020scalable, wang2020connecting}. Thus, we do not experiment with different contextualized KG construction procedures, since it is out of the scope of our work.

Let us define the KG nodes mentioned in $q$ and $a_i$ as QA nodes. For example, for the question \textit{What would you put in a teakettle?} and answer choice \textit{water}, the QA nodes would be \begin{small}\texttt{put}\end{small}, \begin{small}\texttt{teakettle}\end{small}, and \begin{small}\texttt{water}\end{small}. We ground raw mentions of QA nodes to the KG via spaCy-based lemmatization and stop-word filtering \cite{honnibal-johnson:2015:EMNLP}.

For node-based models (MHGRN \cite{feng2020scalable}), we select $\mathcal{V}_i \subseteq \tilde{\mathcal{V}}$ as the QA nodes and all nodes in the QA nodes' 1-hop KG neighborhood. Next, we choose $\mathcal{R}_i \subseteq \tilde{\mathcal{R}}$ as all of the relations between concepts in $\mathcal{V}_i$. Finally, we take $\mathcal{E}_i \subseteq \tilde{\mathcal{E}}$ as all of the edges involving $\mathcal{V}_i$ and $\mathcal{R}_i$.

For path-based models (PathGen \cite{wang2020connecting}, RN \cite{feng2020scalable, battaglia2018relational}), we select $\tilde{\mathcal{G}_i}$ as all 2-hop paths between all question-answer node pairs. Thus, $\mathcal{V}_i \subseteq \tilde{\mathcal{V}}$ consists of the QA nodes as well as all intermediate nodes in the 2-hop paths. Meanwhile, $\mathcal{R}_i \subseteq \tilde{\mathcal{R}}$ and $\mathcal{E}_i \subseteq \tilde{\mathcal{E}}$ consist of all relations and edges within the 2-hop paths. When reasoning over the 2-hop paths, the model does not actually use the intermediate nodes, perhaps in order to keep the path more general \cite{feng2020scalable, wang2020connecting}.

\subsection{Alternative Formulation of Coarse Saliency Explanations}
\label{sec:app_coarse}
\textsc{SalKG}-Coarse uses coarse explanations, which state whether $\mathcal{G}$ or \texttt{None} (\textit{i.e.}, no $\mathcal{G}$) should be used for the given task instance.
By default, \textsc{SalKG}-Coarse uses our proposed ensemble-based coarse explanations (Sec. \ref{sec:explanations_coarse}). 
In this case, the coarse explanations decide between $\mathcal{G}$ and \texttt{None} at the \textit{prediction} level.
That is, the coarse explanations correspond to saliency weights which perform attention over $\mathcal{F}_{\text{KG}}$'s and $\mathcal{F}_{\text{No-KG}}$'s predictions.

% In Sec. \ref{sec:explanations}, we stated that the discretization of real-valued saliency scores into binary saliency targets depends on the values of certain threshold parameters. In all of our experiments, we use the same thresholds. For \textsc{SalKG}-Coarse on multi-choice QA, the threshold $T$ depends on the base PLM's ($\mathcal{F}_{\text{No-KG}}$) predicted answer choice probabilities for the given question instance. See Appendix \ref{sec:app_coarse_sal} for more details about how $T$ is computed. 

\paragraph{Graph Embedding Based Explanations}
In Sec. \ref{sec:experiments_ablation}, we also considered applying coarse explanations at the graph embedding level.
In this case, using $\mathcal{G}$ corresponds to using graph embedding $\mathbf{g}$, while using \texttt{None} corresponds to using some baseline embedding $\mathbf{b}$ that does not contain any information from $\mathcal{G}$.
$\mathbf{b}$ could be a zero vector, random vector, \textit{etc.}
Our experiments in Sec. \ref{sec:experiments_ablation} --- with $\mathbf{b}$ as a zero vector and Grad/Occl as saliency methods --- show that this approach does not yield good empirical results.
We believe the issue is that $\mathbf{b}$ does not contain any \texttt{None}-specific information.
Recall that the ensemble-based \textsc{SalKG}'s prediction is a weighted sum of $\mathcal{F}_{\text{KG}}$'s and $\mathcal{F}_{\text{No-KG}}$'s predictions, which means we interpolate between $\mathcal{F}_{\text{KG}}$'s and $\mathcal{F}_{\text{No-KG}}$'s predictions.
Here, $\mathcal{F}_{\text{No-KG}}$'s prediction actually contains meaningful information about $\mathcal{F}_{\text{No-KG}}$.
On the other hand, it does not make sense to interpolate between $\mathbf{g}$ and $\mathbf{b}$, since $\mathbf{b}$ does not have any meaningful information.
We also considered learning $\mathbf{b}$ when training the KG model, but this would require a complicated multitask learning setup where the KG and No-KG models are jointly trained using $\mathbf{g}$ and $\mathbf{b}$, respectively.

\subsection{Implementation Details for Grad-Based Fine Saliency Explanations}
\label{sec:app_fine}

% Given unit $u$ and its embedding $\mathbf{u} \in \R^d$ in $\mathcal{G}'$, the general form of the \textit{gradient$\times$input} (Grad) \cite{denil2014extraction} fine saliency score is:
% \begin{equation*}
%     s_{\text{f}}(u) = \sum_{j=1}^d \mathbf{u}_j \frac{\partial F_{\text{KG}}(x, \mathcal{G}')}{\partial \mathbf{u}_j}
% \end{equation*}

% For multi-choice QA, $u$ is a more salient unit in $\mathcal{G}_i'$ if it increases $p_{\text{KG}}(q, a_i)$ for $a_i = a^*$ and decreases $p_{\text{KG}}(q, a_i)$ for $a_i \neq a^*$. Thus, we compute $s_{\text{f}}(\mathbf{u})$ in multi-choice QA as follows:

In Sec. \ref{sec:explanations_fine}, we discussed the \textit{gradient$\times$input} (Grad) \cite{denil2014extraction} method for computing raw fine saliency scores $\phi$.
For multi-choice QA, assume we are given text statement $x_i = q \oplus a_i$ (formed from question $q$ and answer choice $a_i$), KG $\mathcal{G}_i$, unit $u_{ij}$, and $u_{ij}$'s embedding $\mathbf{u}_{ij} \in \R^d$ in $\mathcal{G}_i$.  
Also, let $u_{ij}^{(\ell)}$ be the $\ell$-th element of $u_{ij}$.
Then, $\phi$ is computed as follows: 
\begin{equation}
    \phi(u_{ij}; x_i, \mathcal{G}_i) = \numberthis \label{eq:3}
    \begin{cases}
        \sum_{\ell=1}^d \mathbf{u}_{ij}^{(\ell)} \frac{\partial p_{\text{KG}}(x_i, \mathcal{G}_i)}{\partial \mathbf{u}_{ij}^{(\ell)}}, \hspace{-1.5mm} &a_i = a^* \\
        -\sum_{\ell=1}^d \mathbf{u}_{ij}^{(\ell)} \frac{\partial p_{\text{KG}}(x_i, \mathcal{G}_i)}{\partial \mathbf{u}_{ij}^{(\ell)}}, \hspace{-1.5mm} &a_i \neq a^*
    \end{cases}
\end{equation}

Depending on the type of graph encoder used, a unit may or may not be given to the model as a single embedding. While node-based graph encoders take node embeddings as input, path-based graph encoders do not take path embeddings as input. Instead path-based graph encoders take node and relation embeddings as input, then form path embeddings from these node and relation embeddings.

As a result, for Grad, the computation of $\phi$ is slightly different between node-based and path-based graph encoders. 
For node-based encoders, unit embedding $\mathbf{u}_{ij}$ is just a node embedding. Thus, a node's $\phi$ score is computed directly using Eq. \ref{eq:3}. 
For path-based encoders, given a path, we first use Eq. \ref{eq:3} to compute a separate $\phi$ score for each node embedding and relation embedding in the path. Then, we compute the path's $\phi$ score as the sum of the $\phi$ scores of its constituent nodes and relations.

\subsection{Evaluation Protocol}
\label{sec:app_eval_protocol}
We present a more detailed description of the evaluation protocol used to obtain the results in Sec. \ref{sec:exp}.
First, define non-explanation models (No-KG, KG, and No-KG + KG) as models that are not regularized with any kind of explanation, and define explanation models (\textsc{Random}, \textsc{Heuristic}, \textsc{SalKG}) as models that are regularized with some kind of explanation.
Second, each non-explanation model’s performance is reported as the average over three seeds, which we denote as the non-explanation seeds.
Also, recall that each explanation model is built from No-KG and/or KG models.
Third, for each of the three non-explanation seeds, we train the explanation model on three more seeds, which we call the explanation seeds.
After that, we compute the explanation model performance by averaging over [three non-explanation seeds] $\times$ [three explanation seeds] = [nine total seeds].

We summarize the evaluation protocol below:
\begin{itemize}
    \item Non-explanation seeds: 1, 2, 3
    \item Explanation seeds: A, B, C
    \item Non-explanation performance: \textit{average}(1, 2, 3)
    \item Explanation performance: \textit{average}(1A, 1B, 1C, 2A, 2B, 2C, 3A, 3B, 3C)
\end{itemize}

\subsection{Dataset Details}
\label{sec:app_datasets}
Below are more detailed descriptions of the three datasets used for the experiments in Sec. \ref{sec:exp}.
All datasets and resources used in this paper are publicly available and free for any researcher to use.

\textbf{CommonsenseQA (CSQA)} \cite{talmor2019commonsenseqa} is a multi-choice QA dataset whose questions require commonsense reasoning to solve.
Questions and answer choices in CSQA are derived from ConceptNet \cite{speer2017conceptnet}.
The official (OF) data split has 9741/1221/1140 questions for OFtrain/OFdev/OFtest.
Since the labels for OFtest are not publicly available, we use the in-house (IH) data split introduced in \cite{lin2019kagnet} and used in many subsequent works \cite{feng2020scalable, wang2020connecting, yan2020learning}.
The in-house data split has 8500/1221/1241 questions for IHtrain/IHdev/IHtest, where the IHtrain and IHtest are obtained by partitioning OFtrain.

\textbf{OpenbookQA (OBQA)} \cite{mihaylov2018can} is a multi-choice QA dataset which aims to simulate open-book science exams.
OBQA has 4957/500/500 elementary-school-level science questions for train/dev/test, but also provides a supplementary ``open book'' resource containing 1326 core science facts.
To solve questions from OBQA, the model needs to reason over both information from the open book and commonsense knowledge from the KG (\textit{i.e.}, ConceptNet).

\textbf{CODAH} \cite{chen2019codah} is a multi-choice QA dataset which augments the SWAG \cite{zellers2018swag} sentence completion dataset with more difficult, adversarially-created questions.
Similar to SWAG, CODAH's questions are designed to require commonsense reasoning to solve.
CODAH contains 2801 questions, and its official split specifies five folds, which balance the distribution of question categories per fold.
Thus, by default, performance is evaluated by averaging over the five folds.
However, due to computational constraints, we only evaluate on the first fold and compare to the baselines presented in Sec. \ref{sec:oracle_eval} and Sec. \ref{sec:exp}, rather than to previously published methods.

\subsection{Threshold Tuning for Creating Explanations}
\label{sec:app_threshold}

\paragraph{Tuning $T$ Threshold for Coarse Explanations}
Recall that coarse explanations are binarized via threshold $T$ (Sec. \ref{sec:explanations_coarse}).
To set $T$, we manually tune $T$ to maximize \textsc{Oracle}-Coarse's dev accuracy.
This can be done efficiently, since \textsc{Oracle}-Coarse does not require any training.
We use a sweep of $T = [0.01, 0.02, 0.03, 0.04, 0.05]$ and find that $T = 0.01$ yields best performance overall.

\paragraph{Tuning top-$k$\% Threshold for Fine Explanations}
Recall that fine explanations are binarized via threshold $k$, used to set the top-$k$\% of units as positive (Sec. \ref{sec:explanations_fine}).
To set $k$, we manually tune $k$ to maximize \textsc{SalKG}-Coarse's dev accuracy.
Table \ref{tab:topk} shows the performance of RoBERTa+MHGRN and RoBERTa+PathGen on CSQA and OBQA, across different values of $k$.
Due to computational constraints, we report the average performance across [best non-explanation seed] $\times$ [three explanation seeds] = [three total seeds], as opposed to the default [three non-explanation seed] $\times$ [three explanation seeds] = [nine total seeds] (Sec. \ref{sec:app_eval_protocol}).
We use a sweep of $k = [5, 10, 30, 50]$ and find that $k = 5$ yields best performance overall, although there is not a clear trend that smaller $k$ is better.
In this paper, we used $k = 10$ for all experiments, so it may be promising to further explore tuning $k$ in the future.

% For \textsc{SalKG}-Fine, we use $k = 10$ for top-$k$ selection. This threshold was chosen by manually inspecting the saliency of 50 randomly sampled answer choice KGs in CSQA.

\begin{table*}[t]
\centering
\scalebox{0.70}{
\begin{tabular}{lcccc}
    \toprule &
    \multicolumn{2}{c}{\textbf{CSQA Test Accuracy (\%)}} & \multicolumn{2}{c}{ \textbf{OBQA Test Accuracy (\%)}}\\
    \cmidrule(lr){2-3} \cmidrule(lr){4-5}
    \textbf{Top-k\%} & \textbf{MHGRN} & \textbf{PathGen} & \textbf{MHGRN} & \textbf{PathGen} \\
    \midrule
    2       & 72.66 ($\pm$1.52)   & 69.86 ($\pm$1.11)  & 66.47 ($\pm$1.27) & 61.33 ($\pm$2.69)    \\
    5       & 72.58 ($\pm$0.74) & \textbf{71.64} ($\pm$3.17)    & \textbf{69.13} ($\pm$0.81)    & \textbf{64.80} ($\pm$1.40)    \\
    10      & \textbf{73.65} ($\pm$0.21) & 71.39 ($\pm$1.54)    & 65.07 ($\pm$1.70)    & 51.60 ($\pm$1.13)    \\
    30      & 71.98 ($\pm$0.47) & 69.76 ($\pm$0.44)    & 63.47 ($\pm$1.14)    & 61.87 ($\pm$4.61)    \\
    50      & 72.93 ($\pm$0.84) & 71.04 ($\pm$0.05)    & 63.27 ($\pm$3.00)    & 63.60 ($\pm$1.71)    \\
    70      & 72.04 ($\pm$1.05)   & 70.13 ($\pm$0.66)  & 65.80 ($\pm$1.91) & 64.40 ($\pm$0.40)    \\
    \bottomrule
\end{tabular}
}
\caption{\small \textbf{\textsc{SalKG}-Fine Performance for Different top-k\% Thresholds.} We report performance for RoBERTa+MHGRN and RoBERTa+PathGen on CSQA and OBQA. Best model is shown in \textbf{bold}.
}
\label{tab:topk}
\vspace{-0.3cm}
\end{table*}

% \begin{table*}[t]
% \centering
% \scalebox{0.70}{
% \begin{tabular}{lcccc}
%     \toprule &
%     \multicolumn{2}{c}{\textbf{CSQA Dev Accuracy (\%)}} & \multicolumn{2}{c}{ \textbf{OBQA Dev Accuracy (\%)}}\\
%     \cmidrule(lr){2-3} \cmidrule(lr){4-5}
%     \textbf{Top-k\%} & \textbf{MHGRN} & \textbf{PathGen} & \textbf{MHGRN} & \textbf{PathGen} \\
%     \midrule
%     5       & 76.88 ($\pm$0.62)    & 75.18 ($\pm$0.70)    & 69.27 ($\pm$0.64)    & 69.07 ($\pm$0.31)    \\
%     10      \\
%     30      & 76.82 ($\pm$0.57)    & 76.19 ($\pm$0.47)    & 65.33 ($\pm$3.20)    & 66.00 ($\pm$1.97)    \\
%     50      & 76.96 ($\pm$0.25)    & 76.60 ($\pm$0.47)    & 68.33 ($\pm$1.70)    & 67.20 ($\pm$1.91)    \\
%     \bottomrule
% \end{tabular}
% }
% \caption{\small \textbf{\textsc{SalKG}-Fine performance with varying top-k\% on CSQA and OBQA for RoBERTa models.
% }
% }
% \label{tab:topk2}
% \end{table*}

\subsection{Additional Details about \textsc{Oracle} Models}
\label{sec:app_oracle}
We provide more details about \textsc{Oracle}-Coarse and \textsc{Oracle}-Fine.
Given the coarse saliency explanations, \textsc{Oracle}-Coarse simply involves choosing the ``correct'' prediction --- between $\mathcal{F}_{\text{KG}}$'s and $\mathcal{F}_{\text{No-KG}}$'s predictions --- for each answer choice.
Given that $\mathcal{F}_{\text{KG}}$'s and $\mathcal{F}_{\text{No-KG}}$'s predictions are simply loaded from disk, this process runs very quickly, since it does not require additional training.
On the other hand, \textsc{Oracle}-Fine involves training the KG-augmented model while applying the fine saliency explanations as a binary mask to the graph encoder's attention weights.

\subsection{Additional \textsc{SalKG} Results on CODAH}
\label{sec:app_codah}
In this section, we present additional \textsc{SalKG} results on CODAH.
These additional results consist of RoBERTa+RN, BERT+MHGRN, BERT+PathGen, and BERT+RN, all using threshold top-$10$\%.
Also, across all settings, we report both Grad and Occl results for \textsc{SalKG}-Fine and \textsc{SalKG}-Hybrid.
Due to computational constraints, we report the average performance across [best non-explanation seed] $\times$ [three explanation seeds] = [three total seeds], as opposed to the default [three non-explanation seed] $\times$ [three explanation seeds] = [nine total seeds] (Sec. \ref{sec:app_eval_protocol}).
These results are shown in Table \ref{tab:app_codah}, along with the RoBERTa+MHGRN and RoBERTa+PathGen results from Table \ref{tab:obqa_codah}.

First, we see that \textsc{SalKG}-Hybrid (either Grad or Occl) performs the best on all settings except RoBERTa+PathGen.
For RoBERTa+PathGen, \textsc{Random}-Coarse and \textsc{Random}-Hybrid perform the best, although some \textsc{SalKG} models perform almost as well.
\textsc{Random}'s strong performance is likely due to us reporting performance for the best non-explanation seed, rather than averaging over three non-explanation seeds.
Second, for \textsc{SalKG}-Fine, Occl beats Grad on all settings except RoBERTa+PathGen.
Third, for \textsc{SalKG}-Hybrid, Occl beats Grad on BERT+MHGRN, BERT+PathGen, and BERT+RN, while Grad beats Occl on RoBERTa+MHGRN and RoBERTa+PathGen.

\begin{table*}[t]
% \vspace{-1.1cm}
\centering
\vspace{-0.2cm}
\scalebox{0.70}{
\begin{tabular}{lcccccc}
    \toprule &
    \multicolumn{6}{c}{ \textbf{CODAH Test Accuracy (\%)}}\\
    \cmidrule(lr){2-7}
    & \multicolumn{2}{c}{ \textbf{MHGRN}} & \multicolumn{2}{c}{ \textbf{PathGen}} & \multicolumn{2}{c}{ \textbf{RN}} \\
    \cmidrule(lr){2-3}
    \cmidrule(lr){4-5}
    \cmidrule(lr){6-7}
    \textbf{Model} & BERT & RoBERTa & BERT & RoBERTa & BERT & RoBERTa \\
    \midrule

    {No-KG} & 60.96 ($\pm$1.27) & 83.96 ($\pm$0.79) & 60.96 ($\pm$1.27) & 83.96 ($\pm$0.79) & 60.96 ($\pm$1.27) & 83.96 ($\pm$0.79) \\
    {KG} & 58.68 ($\pm$1.63) & 84.02 ($\pm$1.27) & 58.80 ($\pm$2.01) & 84.02 ($\pm$1.62) & 55.92 ($\pm$1.04) & 82.64 ($\pm$0.85) \\
    {No-KG + KG} & 60.60 ($\pm$1.30) & 84.08 ($\pm$1.46) & 60.42 ($\pm$1.14) & 84.69 ($\pm$1.48) & 58.62 ($\pm$1.53) & 84.08 ($\pm$0.55) \\

    \midrule
    {\textsc{Random}-Coarse} & 60.78 ($\pm$0.38) & 84.62 ($\pm$0.55) & 61.74 ($\pm$0.28) & \textbf{86.07} ($\pm$0.89) & 57.84 ($\pm$0.83) & 84.14 ($\pm$0.65) \\
    {\textsc{Random}-Fine} & 58.50 ($\pm$0.91) & 84.02 ($\pm$0.89) & 54.47 ($\pm$1.55) & 75.74 ($\pm$4.71) & 54.53 ($\pm$1.40) & 76.10 ($\pm$4.16) \\
    {\textsc{Random}-Hybrid} & 62.16 ($\pm$0.00) & 84.80 ($\pm$0.10) & 61.74 ($\pm$0.55) & 84.68 ($\pm$0.18) & 62.40 ($\pm$0.10) & 84.14 ($\pm$0.65)  \\

    \midrule

    {\textsc{Heuristic}-Coarse} & 58.38 ($\pm$0.00) & 85.11 ($\pm$0.10) & 61.08 ($\pm$0.00) & {85.59 ($\pm$0.00)} & 59.70 ($\pm$0.10) & 83.60 ($\pm$0.00) \\
    {\textsc{Heuristic}-Fine} & 60.18 ($\pm$1.36) & 83.72 ($\pm$0.92) & 55.98 ($\pm$0.28) & 82.64 ($\pm$2.61) & 54.71 ($\pm$3.07) & 81.80 ($\pm$2.77) \\
    {\textsc{Heuristic}-Hybrid} & 62.16 ($\pm$0.00) & 84.80 ($\pm$0.10) & 61.98 ($\pm$0.31) & 85.23 ($\pm$0.00) & 62.28 ($\pm$0.10) & 85.35 ($\pm$0.10)\\
    \midrule

    {\textsc{SalKG}-Coarse} & 61.02 ($\pm$0.10) & 85.41 ($\pm$0.18) & 61.20 ($\pm$0.28) & 85.95 ($\pm$0.18) & 61.74 ($\pm$0.21) & 84.98 ($\pm$0.42) \\
    {\textsc{SalKG}-Fine (Occl Top-10\%)} & 60.00 ($\pm$1.26) & 84.08 ($\pm$1.14) & 57.72 ($\pm$1.09) & 83.36 ($\pm$0.81) & 59.16 ($\pm$2.15) & 83.78 ($\pm$1.41) \\
     {\textsc{SalKG}-Fine (Grad Top-10\%)} & 59.16 ($\pm$0.38) & 84.20 ($\pm$1.17) & 57.36 ($\pm$0.75) & 83.00 ($\pm$1.51) & 55.86 ($\pm$0.79) & 83.66	($\pm$0.89) \\

    {\textsc{SalKG}-Hybrid (Occl Top-10\%)} & \textbf{62.28} ($\pm$0.10) & 85.71 ($\pm$0.10) & \textbf{62.04} ($\pm$0.45) & 84.44 ($\pm$0.63) & \textbf{62.58} ($\pm$0.10) & \textbf{85.11} ($\pm$0.28) \\
    {\textsc{SalKG}-Hybrid (Grad Top-10\%)} & 60.48 ($\pm$0.21) & \textbf{88.17} ($\pm$0.10) & 61.02 ($\pm$0.10) & 85.17  ($\pm$0.28) & 61.38 ($\pm$0.68) & \textbf{85.11} ($\pm$0.55) \\

    \bottomrule 
\end{tabular}
}
% \vspace{-1mm}
\caption{\small \textbf{\textsc{SalKG} Performance on CODAH for Additional Settings.}
Building upon the CODAH results in Table \ref{tab:obqa_codah} (RoBERTa+MHGRN and RoBERTa+PathGen), we additionally report results for RoBERTa+RN, BERT+MHGRN, BERT+PathGen, and BERT+RN, all using threshold top-$10$\%. We also report both Grad and Occl results for \textsc{SalKG}-Fine and \textsc{SalKG}-Hybrid. Best model is shown in \textbf{bold}.
}
\label{tab:app_codah}
\vspace{-0.3cm}
\end{table*}

\begin{table*}[b]
% \vspace{-1.1cm}
\centering
\vspace{-0.2cm}
\scalebox{0.70}{
\begin{tabular}{lcccccc}
    \toprule &
    \multicolumn{6}{c}{ \textbf{CSQA Test Accuracy (\%)}}\\
    \cmidrule(lr){2-7}
    & \multicolumn{2}{c}{ \textbf{MHGRN}} & \multicolumn{2}{c}{ \textbf{PathGen}} & \multicolumn{2}{c}{ \textbf{RN}} \\
    \cmidrule(lr){2-3}
    \cmidrule(lr){4-5}
    \cmidrule(lr){6-7}
    \textbf{Model} & BERT & RoBERTa & BERT & RoBERTa & BERT & RoBERTa \\
    \midrule
    % {No-KG} & 53.13 ($\pm$2.34) & 69.65 ($\pm$1.06) & 53.13 ($\pm$2.34) & 69.65 ($\pm$1.06) & 53.13 ($\pm$2.34) & 69.65 ($\pm$1.06) \\
    % {KG} & 57.48 ($\pm$0.89) & 73.14 ($\pm$0.78) & 56.54 ($\pm$0.73) & 72.58 ($\pm$0.57) & 56.46 ($\pm$1.22) & 71.37 ($\pm$1.20) \\
    % {No-KG + KG} & 56.14 ($\pm$2.29) & 72.14 ($\pm$0.67) & 57.29 ($\pm$1.30) & 71.44 ($\pm$0.71) & 56.86 ($\pm$1.10) & 70.53 ($\pm$1.82) \\
    % \midrule
    % {\textsc{SalKG}-Coarse} & 57.78 ($\pm$0.35) & 74.05 ($\pm$0.14) & 58.23 ($\pm$0.37) & 72.79 ($\pm$0.12) & 57.59 ($\pm$0.33) & 72.44 ($\pm$0.16) \\
    % \midrule
    {\textsc{SalKG}-Fine (Grad)} & 55.44 ($\pm$1.22) & 72.95 ($\pm$1.44) & 57.10 ($\pm$0.81) & 70.10 ($\pm$0.28) & 56.14 ($\pm$1.97) & \textbf{72.12} ($\pm$0.14) \\
    {\textsc{SalKG}-Fine (Occl)} & \textbf{56.78} ($\pm$2.14) & \textbf{73.65} ($\pm$0.21) & \textbf{57.64} ($\pm$2.12) & \textbf{71.39} ($\pm$1.54) & \textbf{56.86} ($\pm$0.41) & 71.58 ($\pm$1.10) \\
    \midrule
    {\textsc{SalKG}-Hybrid (Grad)} & 59.07 ($\pm$0.56) & 72.79 ($\pm$0.20) & 57.53 ($\pm$0.43) & 71.39 ($\pm$0.14) & 57.29 ($\pm$0.29) & 71.98 ($\pm$ 0.28) \\
    {\textsc{SalKG}-Hybrid (Occl)} & \textbf{59.12} ($\pm$0.28) & \textbf{73.41} ($\pm$0.16) & \textbf{60.35} ($\pm$0.32) & \textbf{73.11} ($\pm$1.00) & \textbf{58.80} ($\pm$0.19) & \textbf{74.64} ($\pm$0.09) \\
    % \midrule
    % {Best \textsc{Oracle}} & 95.89 & 98.63 & 90.49 & 96.78 & 92.26 & 95.25 \\
    \bottomrule 
\end{tabular}
}
\vspace{-1mm}
\caption{\small \textbf{CSQA Performance Comparison for \textsc{SalKG} Grad \textit{vs.} Occl Models.} Best model between Grad and Occl is shown in \textbf{bold}.}
\label{tab:csqa_grad}
\vspace{5mm}
\end{table*}

\begin{table*}[t]
% \vspace{-1.1cm}
\centering
% \vspace{-0.2cm}
\scalebox{0.70}{
\begin{tabular}{lcccccc}
    \toprule &
    \multicolumn{6}{c}{ \textbf{OBQA Test Accuracy (\%)}}\\
    \cmidrule(lr){2-7}
    & \multicolumn{2}{c}{ \textbf{MHGRN}} & \multicolumn{2}{c}{ \textbf{PathGen}} & \multicolumn{2}{c}{ \textbf{RN}} \\
    \cmidrule(lr){2-3}
    \cmidrule(lr){4-5}
    \cmidrule(lr){6-7}
    \textbf{Model} & BERT & RoBERTa & BERT & RoBERTa & BERT & RoBERTa \\
    \midrule
    % {No-KG} & 53.73 ($\pm$0.42) & 68.73 ($\pm$0.31) & 53.73 ($\pm$0.42) & 68.73 ($\pm$0.31) & 55.60 ($\pm$0.42) & 68.73 ($\pm$0.31) \\
    % {KG} & 55.60 ($\pm$2.12) & 68.87 ($\pm$2.16) & 53.33 ($\pm$1.70) & 68.40 ($\pm$1.59) & 56.87 ($\pm$1.70) & 66.80 ($\pm$4.73) \\

    % {No-KG + KG} & 55.20 ($\pm$1.44) & 68.53 ($\pm$0.95) & 54.93 ($\pm$1.47) & 69.67 ($\pm$1.45) & 54.67 ($\pm$1.55) & 69.40 ($\pm$0.35) \\
    % \midrule
    % {\textsc{SalKG}-Coarse} & 56.60 ($\pm$0.00) & 69.93 ($\pm$0.31) & 55.27 ($\pm$0.58) &70.00 ($\pm$0.35) &57.93 ($\pm$0.50) & 71.47 ($\pm$0.58) \\
    % \midrule
    {\textsc{SalKG}-Fine (Grad)} & 53.40 ($\pm$0.69) & 58.80 ($\pm$8.66) & 55.33 ($\pm$0.31) & \textbf{67.87} ($\pm$1.81) & \textbf{56.53} ($\pm$0.31) & \textbf{68.87} ($\pm$1.67) \\
    {\textsc{SalKG}-Fine (Occl)} & \textbf{53.93} ($\pm$1.01) & \textbf{65.07} ($\pm$1.70) & \textbf{55.40} ($\pm$0.53) & 51.60 ($\pm$1.13) & 55.67 ($\pm$0.90) & 62.33 ($\pm$0.90) \\
    \midrule
    {\textsc{SalKG}-Hybrid (Grad)} & 53.80 ($\pm$0.20) & 69.47 ($\pm$0.31) & \textbf{55.67} ($\pm$0.64) & 69.93 ($\pm$0.61) & 53.20 ($\pm$0.72) & 69.40 ($\pm$0.20) \\
    {\textsc{SalKG}-Hybrid (Occl)} & \textbf{56.20} ($\pm$0.20) & \textbf{70.73} ($\pm$0.12) & 55.33 ($\pm$0.23) & \textbf{70.07} ($\pm$0.12) & \textbf{53.93} ($\pm$0.42) & \textbf{70.80} ($\pm$0.00) \\
    % {\textsc{SalKG}-Hybrid (Grad)} & - & - & - & - & - & - \\
    % {\textsc{SalKG}-Hybrid (Occl)} & - & - & - & - & - & - \\
    % \midrule
    % {\textsc{SalKG}-Coarse-Target} & - & - & - & - & - & - \\
    % {\textsc{SalKG}-Fine-Target} & - & - & - & - & - & - \\
    % {\textsc{SalKG}-Hybrid-Target} & - & - & - & - & - & - \\
    % {Best \textsc{Oracle}} & 87.00 & 89.60 & 92.80 & 92.80 & 85.40 & 86.80 \\
    
    % {\textsc{SalKG}-Coarse (Ensemble)} & \textbf{55.87} ($\pm$1.10) & \textbf{69.93} ($\pm$0.42) & \textbf{58.07} ($\pm$0.92) & \textbf{69.20} ($\pm$0.40) & 57.93 ($\pm$0.50) & \textbf{71.13} ($\pm$0.50) \\
    % {\textsc{SalKG}-Coarse (Graph Emb)} & - & - & - & - & - & - \\
    % {\textsc{SalKG}-Fine (Unnormalized top-10)} & 55.07 ($\pm$1.27) & 60.20 ($\pm$7.00) & 51.53 ($\pm$1.36) & 67.47 ($\pm$1.14) & 51.20 ($\pm$1.40) & 67.20 ($\pm$1.71) \\
    % {\textsc{SalKG}-Fine (Normalized top-10\%)} & - & - & - & - & - & - \\
    % {\textsc{SalKG}-Hybrid} & - & - & - & - & - & - \\
    % \midrule
    % {\textsc{SalKG}-Coarse-Target (Ensemble)} & 70.60 & 79.40 & 65.00 & 76.60 & 69.00 & 79.00 \\
    % {\textsc{SalKG}-Coarse-Target (Graph Emb)} & - & - & - & - & - & - \\
    % {\textsc{SalKG}-Fine-Target (Unnormalized top-10)} & 72.27 ($\pm$7.60) & 78.40 ($\pm$7.79) & 60.33 ($\pm$0.58) & 71.27 ($\pm$0.50) & 66.53 ($\pm$3.40)  & 52.07 ($\pm$3.56) \\
    % {\textsc{SalKG}-Fine-Target (Normalized top-10\%)} & - & - & - & - & - & - \\
    % {\textsc{SalKG}-Fine-Target} & 80.07^* ($\pm$1.42) & 82.27^* ($\pm$4.69) & - & - & - & - \\
    % {\textsc{SalKG}-Hybrid-Target} & - & - & - & - & - & - \\
    \bottomrule 
\end{tabular}
}
% \vspace{-1mm}
\caption{\small \textbf{OBQA Performance Comparison for \textsc{SalKG} Grad \textit{vs.} Occl Models.} Best model between Grad and Occl is shown in \textbf{bold}.} 
\label{tab:obqa_grad}
\vspace{0.1cm}
\end{table*}

\subsection{Additional \textsc{SalKG} Results for Grad \textit{vs.} Occl}
\label{sec:app_grad_vs_occl}
In Tables \ref{tab:csqa_grad}-\ref{tab:obqa_grad}, we compare Grad \textit{vs.} Occl on CSQA and OBQA, respectively.
Due to computational constraints, we report the average test accuracy across [best non-explanation seed] $\times$ [three explanation seeds] = [three total seeds], as opposed to the default [three non-explanation seed] $\times$ [three explanation seeds] = [nine total seeds] (Sec. \ref{sec:app_eval_protocol}).
For \textsc{SalKG}-Fine and \textsc{SalKG}-Hybrid on CSQA, we find that Occl beats Grad on all settings, except \textsc{SalKG}-Fine on RoBERTa+RN.
However, for \textsc{SalKG}-Fine on OBQA, Grad beats Occl on RoBERTa+PathGen, BERT+RN, and RoBERTa+RN, while Occl beats Grad on BERT+MHGRN, RoBERTa+MHGRN, and BERT+PathGen. 
Meanwhile, for \textsc{SalKG}-Hybrid on OBQA, Occl beats Grad on all settings except BERT+PathGen.
Thus, we see that Occl generally outperforms Grad, although Grad can beat Occl on certain settings.

\subsection{Comparison to Published OBQA Baselines}
\label{sec:app_obqa_pub}
To further demonstrate that \textsc{SalKG} models perform competitively, we also compare \textsc{SalKG} to the many KG-augmented model baseline results published in \cite{feng2020scalable, wang2020connecting, yan2020learning}, for OBQA.
The baselines we consider are RN, RN + Link Prediction, RGCN, GconAttn, MHGRN, and PathGen.
For the non-\textsc{SalKG} versions of MHGRN, PathGen, and RN, we quote the published results.
Since these published results average over four seeds (instead of three), we report \textsc{SalKG} results over four seeds in Table \ref{tab:obqa_pub}.
For OBQA, we find that vanilla PathGen (quoted from published results) performs the best, while \textsc{SalKG}-Hybrid (MHGRN) and \textsc{SalKG}-Hybrid (PathGen) are almost as good.
These OBQA results indicate that our reproduction of vanilla PathGen may not have been optimally tuned, thus limiting the performance of the \textsc{SalKG} models built upon PathGen.
We plan to investigate this issue in future work.

\begin{table}[t]
% \vspace{-1.1cm}
\centering
% \vspace{-0.2cm}

\scalebox{0.70}{
\begin{tabular}{lc}
    \toprule
    \textbf{Model (RoBERTa)} & \textbf{OBQA Test Accuracy (\%)} \\
   
    \midrule
    
    {RN \cite{santoro2017simple}} & 65.20 ($\pm$1.18) \\
    {RN + Link Prediction \cite{wang2020connecting}} & 66.30 ($\pm$0.48) \\
    {RGCN \cite{schlichtkrull2018modeling}} & 62.45 ($\pm$1.57) \\
    {GconAttn \cite{wang2019improving}} & 64.75 ($\pm$1.48) \\
    {MHGRN \cite{feng2020scalable}} & 66.85 ($\pm$1.19) \\
    {PathGen \cite{wang2020connecting}} & \textbf{71.20} ($\pm$0.96) \\

    \midrule

    {\textsc{SalKG}-Coarse (MHGRN)} & 69.85 ($\pm$0.30) \\
    
    {\textsc{SalKG}-Fine (MHGRN)} & 64.65 ($\pm$1.62) \\

    {\textsc{SalKG}-Hybrid (MHGRN)} & 70.75 ($\pm$0.10) \\
    
    {\textsc{SalKG}-Coarse (PathGen)} & 69.70 ($\pm$0.93)  \\
    
    {\textsc{SalKG}-Fine (PathGen)} & 54.30  ($\pm$5.84) \\

    {\textsc{SalKG}-Hybrid (PathGen)} & 70.00  ($\pm$0.16) \\
    
    \bottomrule 
\end{tabular}
}

\vspace{0.2cm}
\caption{\small \textbf{Comparison of \textsc{SalKG} to Published OBQA Results.
} Best model is shown in \textbf{bold}.
}
\label{tab:obqa_pub}
% \vspace{-0.3cm}
\end{table}

\subsection{Low-Resource Learning}
\begin{figure}[tb!]
\begin{center}
    \centering
% \vspace{-0.6cm}
    \includegraphics[width=0.9\textwidth]{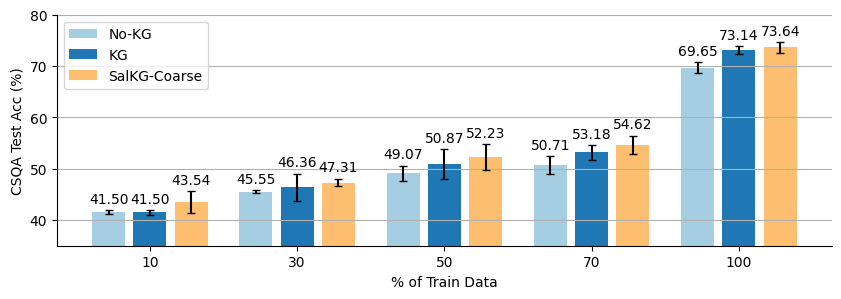}
    % \vspace{-0.6cm}
    \caption{\small \textbf{Low-Resource Learning.} CSQA test accuracy for No-KG, KG, and \textsc{SalKG}-Coarse, when using varying amounts of training data.}
    \label{fig:low_resource}
% \vspace{-0.5cm}
\end{center}
\end{figure}

In Fig. \ref{fig:low_resource}, we show CSQA performance for different models in low-resource settings. Specifically, we experiment with low-resource learning by training the model on 10\%, 30\%, 50\%, or 70\% of the training data. For reference, we also include CSQA performance when using 100\% of the training data. Here, we consider No-KG (RoBERTa), KG (MHGRN), and \textsc{SalKG}-Coarse (RoBERTa+MHGRN). Across all settings, we find that \textsc{SalKG}-Coarse outperforms both No-KG and KG, suggesting that regularizing the model with coarse explanations can provide a helpful inductive bias for generalizing from limited training data.

\subsection{Analyzing the Impact of Coarse Explanations}
\label{sec:app_case_coarse}

\textsc{SalKG}-Coarse is based on the insight that KG information may help the model on some instances but hurt on others. 
% That is, the No-KG and KG base models have complementary predictions.
% , which means there are some instances in the dataset that one of the two models predicts correctly. 
Thus, even if KG outperforms No-KG on average, No-KG may still correctly predict some instances that KG got wrong.
\textsc{SalKG}-Coarse takes advantage of such complementary predictions between No-KG and KG, in order to achieve performance higher than $\max(\text{No-KG}, \text{KG})$.
As shown by RoBERTa+PathGen and RoBERTa+RN on OBQA (Table \ref{tab:obqa_codah}), \textsc{SalKG}-Coarse can still beat $\max(\text{No-KG}, \text{KG}, \text{No-KG} + \text{KG})$ even when No-KG outperforms KG.

In Table \ref{tab:coarse_impact}, we analyze the performance of BERT (\textit{i.e.}, No-KG), PathGen (\textit{i.e.}, KG), \textsc{SalKG}-Coarse (BERT+PathGen), and \textsc{Oracle}-Coarse (BERT+PathGen) on various sets of questions in CSQA. Due to computational constraints, each model's performance here is reported for one seed (instead of using the protocol described in Sec. \ref{sec:app_eval_protocol}), so these results are not directly comparable to those in Table \ref{tab:csqa}.
% In the top section, we report the percentage of questions for which: No-KG is correct, KG is correct, only No-KG is correct, only KG is correct, both are correct, both are incorrect, and at least one is correct.
% In the bottom section, we report the percentage of questions for which: \textsc{SalKG}-Coarse is correct and \textsc{SalKG}-Coarse-Target is correct.
Through this performance breakdown, we can isolate the potential improvement contributed by each base model to \textsc{SalKG}-Coarse. 
We begin by looking at the questions for which \textsc{SalKG}-Coarse has no influence. These are the 46.01\% of questions correctly answered by both models and the 33.92\% of questions incorrectly answered by both models. Since \textsc{SalKG}-Coarse is trained to choose between the two models' predictions, \textsc{SalKG}-Coarse's output is fixed if both models make the same prediction.
This leaves 20.07\% of questions that were correctly answered by exactly one of the two models: 9.43\% were from No-KG, while the other 10.64\% were from KG. This 20.07\% of constitutes the complementary predictions leveraged by \textsc{SalKG}-Coarse.

Based on this question-level analysis, we would estimate the \textsc{Oracle}-Coarse accuracy to be 66.08\%, the percentage of questions that at least one model answered correctly.
However, as stated in Sec. \ref{sec:explanations_coarse}, coarse saliency targets are created at the answer choice level (not question level), which offers us more flexibility to choose between No-KG and KG. As a result, \textsc{Oracle}-Coarse's accuracy is actually 68.57\%.
This leaves \textsc{SalKG}-Coarse (56.65\%) significant room for improvement, perhaps through better model architecture and training.

% Instead of always using the KG for making predictions, the \textsc{SalKG}-Coarse model only uses the KG for a given task instance if it determines that the KG provides salient information for this instance. By doing this, the model is able to take advantage of the fact that the No-KG and KG models produce complementary predictions. Sometimes, only the text input is relevant, so the No-KG model is better. In other cases, the KG input provides relevant information, so the KG model is better.

\begin{table}[t]
\centering
\scalebox{0.70}{
\begin{tabular}{lc}
    \toprule  
    {\textbf{Question Set}} & \textbf{Question Percentage (\%)} \\
    \midrule
    {No-KG Correct} & 55.44 \\
    {KG Correct} & 56.65 \\
    {Only No-KG Correct} & 9.43 \\
    {Only KG Correct} & 10.64 \\
    {Both Correct} & 46.01 \\
    {Both Incorrect} & 33.92 \\
    {At Least One Incorrect} & 66.08 \\
    \midrule
    {\textsc{SalKG}-Coarse Correct} & 56.65 \\
    {\textsc{Oracle}-Coarse Correct} & 68.57 \\
    \bottomrule
\end{tabular}
}
\vspace{0.2cm}
\caption{\small \textbf{Impact of Coarse Explanations.} Using BERT+PathGen on CSQA, we present a performance breakdown for various question sets, in order to analyze why \textsc{SalKG}-Coarse is able to beat No-KG and KG.}
\label{tab:coarse_impact}
\vspace{-0.4cm}
\end{table}

\subsection{Comparing Salient and Non-Salient KG Units}
\label{sec:app_sal_vs_nonsal}

This paper explores learning from explanations of KG units' saliency (\textit{i.e.}, usefulness).
Overall, our focus is on how using salient KG units can yield improve model performance.
In this subsection, we also analyze whether salient and non-salient KG units, as determined by our coarse/fine explanation methods, can differ in other ways that are not directly related to performance (Table \ref{tab:sal_vs_nonsal}).
For both coarse and fine explanations, we use the BERT+MHGRN model on CSQA, where MHGRN is a node-based graph encoder (Sec. \ref{sec:oracle_eval}).
Recall that Q nodes and A nodes are nodes (\textit{i.e.}, concepts) mentioned in the given question and answer choice, respectively (Sec. \ref{sec:exp_eval}).

For coarse explanations, we use the ensemble-based explanations introduced in Sec. \ref{sec:explanations_coarse}.
We compare salient and non-salient KGs with respect to the number of nodes in the KG (\# nodes), percentage of Q nodes in the KG (\% Q nodes), percentage of A nodes in the KG (\% A nodes), clustering coefficient (cluster coeff.), and average node degree (degree).
These results are shown in Table \ref{tab:sal_vs_nonsal}a.
We see that these metrics are not very discriminative, as salient and non-salient KGs perform similarly on all of these metrics.

For fine explanations, we use the Grad-based explanations described in Sec. \ref{sec:explanations_fine} and Sec. \ref{sec:app_fine}.
We compare salient and non-salient nodes with respect to the percentage of Q nodes among salient/non-salient nodes in the KG (\% Q nodes), percentage of A nodes among salient/non-salient nodes in the KG (\% A nodes), and node degree (degree).
These results are shown in Table \ref{tab:sal_vs_nonsal}b.
Here, we see that \%Q nodes and \%A nodes are actually quite discriminative metrics between salient and non-salient nodes.
On average, the percentage of Q nodes among salient nodes (16.84\%) is 56.07\% greater than the percentage of Q nodes among non-salient nodes (10.79\%).
Similarly, on average, the percentage of A nodes among salient nodes (10.00\%) is 65.02\% greater than the percentage of Q nodes among non-salient nodes (6.06\%).
However, compared to \%Q nodes and \%A nodes, degree is not as discriminative.
This indicates that the difference between salient and non-salient nodes may be more semantic than structural.

\begin{table}[t]
\centering
\subfloat[\small Salient \textit{vs.} Non-Salient KGs.]{
\scalebox{0.75}{
    \begin{tabular}{lcc}
        \toprule
        \textbf{Metric} & \textbf{Salient} & \textbf{Non-Salient} \\
        \midrule
        % {\# nodes} & 125.877 & 120.568 \\
        % {\% Q nodes} & 9.085 & 9.171 \\
        % {\% A nodes} & 2.943 & 3.124 \\
        {\# nodes} & 125.88 & 120.57 \\
        {\% Q nodes} & 9.09 & 9.17 \\
        {\% A nodes} & 2.94 & 3.12 \\
        % {\% QA paths} & - & - \\
        {cluster coeff.} & 4.26E-1 & 4.25E-1 \\
        {degree} & 9.89 & 9.78 \\
        \bottomrule
    \end{tabular}
}
    
}
\subfloat[\small Salient \textit{vs.} Non-Salient Nodes.]{
\scalebox{0.75}{
    \begin{tabular}{lcc}
        \toprule
        \textbf{Metric} & \textbf{Salient} & \textbf{Non-Salient} \\
        \midrule
        % {\% QA units} & - & - \\
        {\% Q nodes} & 16.84 & 10.79 \\
        {\% A nodes} & 10.00 & 6.06 \\
        % {\% QA paths} & - & - \\
        % {\% QA neighbors} & - & - \\
        {degree} & 15.41 & 13.11 \\
        \bottomrule
    \end{tabular}
}
}
    
\vspace{0.2cm}
\caption{\small \textbf{Salient \textit{vs.} Non-Salient KG Units.} Using BERT+MHGRN on CSQA, we compare salient and non-salient KG units. In (a), we compare salient and non-salient KGs, as determined by coarse explanations. In (b), we compare salient and non-salient nodes, as determined by fine explanations.}
\label{tab:sal_vs_nonsal}
% \vspace{-0.3cm}
\end{table}

\subsection{Robustness to KG Perturbation}
Table \ref{tab:csqa_kg_perturb} shows the CSQA performance of KG and \textsc{SalKG} models subjected to different forms of KG perturbation. Relation perturbation (Relation) permutes the relation labels of all edges in the KG, while node perturbation (Node) permutes the node labels of all nodes in the KG. These perturbation methods are designed to alter the semantics of the KG. For relation perturbation and node perturbation, \textsc{SalKG}-Coarse (Node) performs best on almost all settings, with KG (Node) barely beating \textsc{SalKG}-Coarse for node perturbation on BERT+PathGen. However, with KG perturbation, \textsc{SalKG}-Hybrid does not perform as well, sometimes even worse than KG and \textsc{SalKG}-Fine. This may be because \textsc{SalKG}-Hybrid relies most heavily on fine explanations, making it especially sensitive to KG perturbation.   

We also compare these KG-perturbed models to models without any KG perturbation. As expected, across all settings, the KG-perturbed models outperform the non-KG-perturbed models. Interestingly, we find that \textsc{SalKG}-Coarse is most robust to KG perturbation. For BERT+RN and RoBERTa+RN, \textsc{SalKG}-Coarse (Relation) is less than 1\% worse than \textsc{SalKG}-Coarse. This makes sense, since \textsc{SalKG}-Coarse relies least on the KG. For a given instance, \textsc{SalKG}-Coarse has the option to completely ignore KG information when making its prediction. When the KG is perturbed, it would be advantageous for \textsc{SalKG}-Coarse to focus only on the text input.

\begin{table*}[t]
% \vspace{-1.1cm}
\centering
\vspace{-0.2cm}
\scalebox{0.70}{
\begin{tabular}{lcccccc}
    \toprule &
    \multicolumn{6}{c}{ \textbf{CSQA Test Accuracy (\%)}}\\
    \cmidrule(lr){2-7}
    & \multicolumn{2}{c}{ \textbf{MHGRN}} & \multicolumn{2}{c}{ \textbf{PathGen}} & \multicolumn{2}{c}{ \textbf{RN}} \\
    \cmidrule(lr){2-3}
    \cmidrule(lr){4-5}
    \cmidrule(lr){6-7}
    \textbf{Model} & BERT & RoBERTa & BERT & RoBERTa & BERT & RoBERTa \\
    \midrule

    {KG (Relation)} & 52.89 ($\pm$0.73) & 67.41 ($\pm$0.84) & 52.35 ($\pm$0.60) & 70.08 ($\pm$0.38) & 54.15 ($\pm$0.40) & 68.95 ($\pm$1.58) \\
    {\textsc{SalKG}-Coarse (Relation)} & \textbf{55.86} ($\pm$0.48) & \textbf{72.53} ($\pm$0.50) & \textbf{56.07} ($\pm$0.44) & \textbf{71.55} ($\pm$0.85) & \textbf{56.93} ($\pm$0.51) & \textbf{72.43} ($\pm$0.96) \\
    {\textsc{SalKG}-Fine (Relation)} & 52.58 ($\pm$0.70) & 68.84 ($\pm$0.67) & 53.32 ($\pm$0.61) & 71.23 ($\pm$1.21) & 53.94 ($\pm$0.63) & 69.80 ($\pm$0.64) \\
    {\textsc{SalKG}-Hybrid (Relation)} & 51.28 ($\pm$0.70) & 69.84 ($\pm$0.57) & 53.33 ($\pm$0.55) & 70.34 ($\pm$1.03) & 52.41 ($\pm$1.11) & 68.77 ($\pm$0.80) \\
    
    \midrule
    {KG (Node)} & 53.63 ($\pm$0.70) & 67.35 ($\pm$0.41) & \textbf{55.60} ($\pm$0.16) & 70.51 ($\pm$1.69) & 54.15 ($\pm$2.27) & 70.48 ($\pm$1.71) \\
    {\textsc{SalKG}-Coarse (Node)} & \textbf{55.75} ($\pm$0.60) & \textbf{71.83} ($\pm$0.60) & 55.43 ($\pm$0.55) & \textbf{71.36} ($\pm$0.81) & \textbf{56.14} ($\pm$0.73) & \textbf{71.20} ($\pm$0.72) \\
    {\textsc{SalKG}-Fine (Node)} & 53.60 ($\pm$0.83) & 66.81 ($\pm$1.09) & 53.13 ($\pm$0.99) & 70.80 ($\pm$1.55) & 54.02 ($\pm$0.84) & 71.08 ($\pm$1.02) \\
    {\textsc{SalKG}-Hybrid (Node)} & 51.14 ($\pm$1.03) & 69.58 ($\pm$0.77) & 50.80 ($\pm$0.83) & 69.85 ($\pm$0.72) & 53.24 ($\pm$0.72) & 69.57 ($\pm$1.14) \\

    \midrule
    {KG} & 57.48 ($\pm$0.89) & 73.14 ($\pm$0.78) & 56.54 ($\pm$0.73) & 72.58 ($\pm$0.57) & 56.46 ($\pm$1.22) & 71.37 ($\pm$1.20) \\
    {\textsc{SalKG}-Coarse} & {57.98 ($\pm$0.90)} & {\textbf{73.64} ($\pm$1.05)} & {57.75 ($\pm$0.77)} & \textbf{73.07} ($\pm$0.25) & 57.50 ($\pm$1.25) & 73.11 ($\pm$1.13) \\
    {\textsc{SalKG}-Fine} & 54.36 ($\pm$2.34) & 70.00 ($\pm$0.81) & 54.39 ($\pm$2.03) & 72.12 ($\pm$0.91) & 54.30 ($\pm$1.41) & 71.64 ($\pm$1.51) \\
    {\textsc{SalKG}-Hybrid} & \textbf{58.70} ($\pm$0.65) & 73.37 ($\pm$0.12) & \textbf{59.87} ($\pm$0.42) & 72.67 ($\pm$0.65) & \textbf{58.78} ($\pm$0.14) & \textbf{74.13} ($\pm$0.71) \\

    \bottomrule 
\end{tabular}
}
% \vspace{-1mm}
\caption{\small \textbf{\textsc{SalKG} Performance Comparison on CSQA with Perturbed KGs.} Best performance in \textbf{bold}.
}
\label{tab:csqa_kg_perturb}
\vspace{-0.2cm}
\end{table*}

\begin{table*}[b]
\centering
\vspace{-0.2cm}
\scalebox{0.70}{
\begin{tabular}{lcccccc}
    \toprule &
    \multicolumn{6}{c}{ \textbf{CSQA p-values }}\\
    \cmidrule(lr){2-7}
    & \multicolumn{2}{c}{ \textbf{MHGRN}} & \multicolumn{2}{c}{ \textbf{PathGen}} & \multicolumn{2}{c}{ \textbf{RN}} \\
    \cmidrule(lr){2-3}
    \cmidrule(lr){4-5}
    \cmidrule(lr){6-7}
    \textbf{Model} & BERT & RoBERTa & BERT & RoBERTa & BERT & RoBERTa \\
    \midrule
    {Best \textsc{SalKG} Model \textit{vs.} Best Non-\textsc{SalKG} Model} & 0.1235 & 0.4238 & 0.0701 & 0.2690 & 0.1336 & 0.0441 \\
    \bottomrule 
\end{tabular}
}
\caption{\small \textbf{\textsc{SalKG} T-Test Results on CSQA.} For each setting in Table \ref{tab:csqa}, we perform the T-test between the best \textsc{SalKG} model and the best non-\textsc{SalKG} model.
}
\label{tab:ttest_csqa}
\end{table*}

\begin{table*}[t]
\centering
\scalebox{0.70}{
\begin{tabular}{lccccc}
    \toprule &
    \multicolumn{3}{c}{ \textbf{OBQA p-values }} & \multicolumn{2}{c}{ \textbf{CODAH p-values }}\\
    \cmidrule(lr){2-4} \cmidrule(lr){5-6}
    \textbf{Model (RoBERTa)} & \textbf{MHGRN} & \textbf{PathGen} & \textbf{RN} & \textbf{MHGRN} & \textbf{PathGen} \\
    \midrule
    {Best \textsc{SalKG} Model \textit{vs.} Best Non-\textsc{SalKG} Model} & 0.2909 & 0.8890 & 0.0005 & 0.1223 & 0.2823 \\

    \bottomrule 
\end{tabular}
}
\caption{\small \textbf{\textsc{SalKG} T-Test Results on OBQA and CODAH.} For each setting in Table \ref{tab:obqa_codah}, we perform the T-test between the best \textsc{SalKG} model and the best non-\textsc{SalKG} model.
}
\label{tab:ttest_obqa_codah}
% \vspace{-0.3cm}
\end{table*}

\subsection{Statistical Significance of Main Results}
\label{sec:app_ttest}
In this section, we verify the statistical significance of our results in Sec. \ref{sec:exp_main}.
For each setting in Tables \ref{tab:csqa}-\ref{tab:obqa_codah} (except RoBERTa+PathGen on CODAH), we perform the two-sided unpaired T-test with unequal variance between the best \textsc{SalKG} model and the best non-\textsc{SalKG} model.
The $p$-values are shown in Tables \ref{tab:ttest_csqa}-\ref{tab:ttest_obqa_codah}.

If we use threshold $\alpha=0.1$ (\textit{i.e.}, $p < 0.1$), then we find that SalKG yields statistically significant improvements on CSQA BERT+PathGen, CSQA RN+RoBERTa, and OBQA RN+RoBERTa. If we use threshold $\alpha=0.05$ (\textit{i.e.}, $p < 0.05$), then we find that SalKG yields statistically significant improvements on CSQA RN+RoBERTa and OBQA RN+RoBERTa. In particular, the improvement on OBQA RN+RoBERTa is very statistically significant, with $p=0.0005$.
Our T-test results show that SalKG can produce significant performance gains on a number of model-dataset settings, while yielding competitive performance in other settings.

\begin{figure}[tb!]
\begin{center}
    \centering
% \vspace{-0.4cm}
    \includegraphics[width=\textwidth]{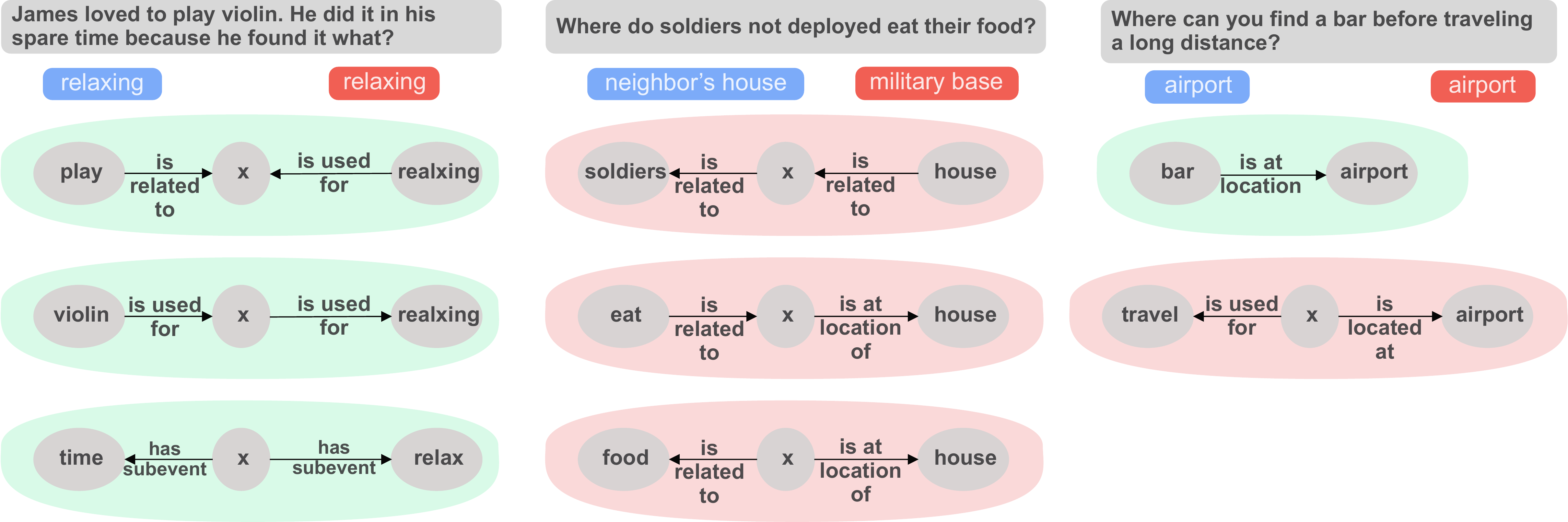}
    % \vspace{-0.6cm}
    \caption{\textbf{\small Examples of coarse/fine saliency explanations.} Illustration of examples presented in Sec. \ref{sec:case_studies}. Blue denotes given answer choice, while red denotes target answer.}
    \label{fig:case_study_1}
% \vspace{-0.5cm}
\end{center}
\end{figure}

\begin{figure}[tb!]
\begin{center}
    \centering
% \vspace{-0.4cm}
    \includegraphics[width=\textwidth]{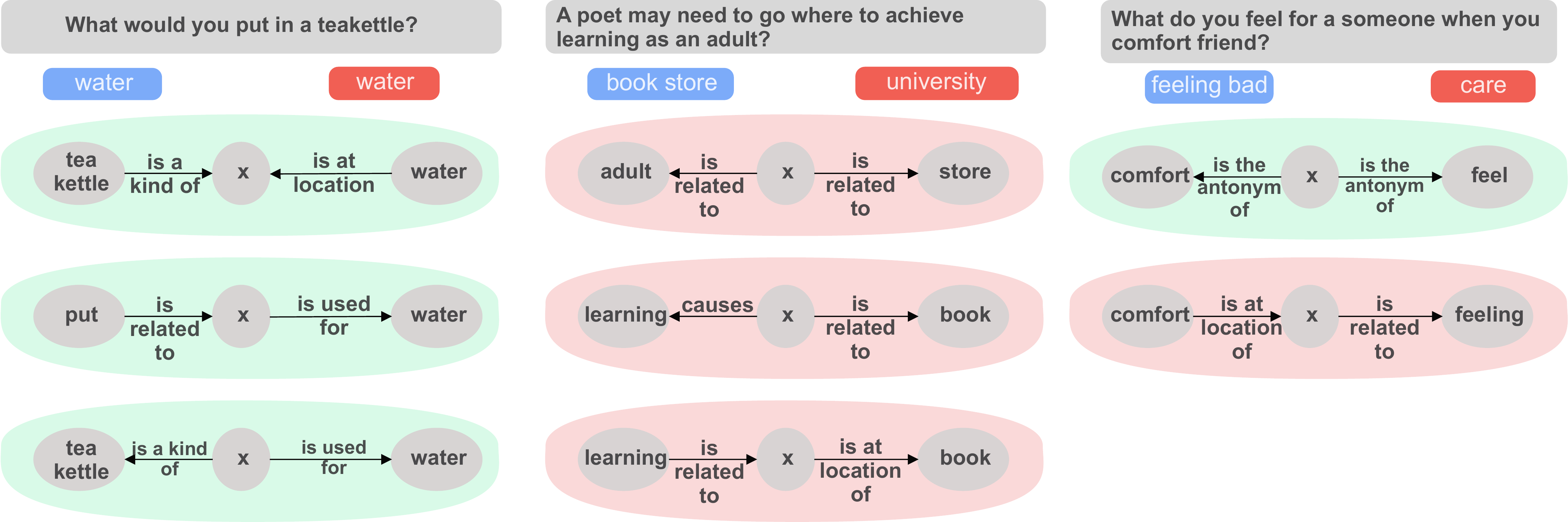}
    \caption{\textbf{\small More examples of coarse/fine saliency explanations.} Illustration of examples presented in Sec. \ref{sec:app_case_qual}. Blue denotes given answer choice, while red denotes target answer.}
    % \vspace{-0.6cm}
    \label{fig:case_study_2}
\vspace{-0.3cm}
\end{center}
\end{figure}

\subsection{Case Studies: Qualitative Analysis of KG Saliency Explanations}
\label{sec:app_case_qual}
In this section, we build upon Sec. \ref{sec:case_studies} and illustrate more examples of coarse/fine explanations created from BERT+PathGen on CSQA, with 1-hop or 2-hop paths as fine units.
Notice that 2-hop paths consist of two nodes and two relations, with the intermediate node replaced with a placeholder node \texttt{x}, following \cite{feng2020scalable}.
By constructing 2-hop paths this way, the model is able to learn from more general 2-hop paths.

First, for coarse explanations, we provide more examples of positive (\textit{i.e.}, useful) and negative KGs.

\begin{itemize}
    \item 
    For the positive KG example, the question is \textit{What would you put in a teakettle?}, the answer choice is \textit{water}, and the target answer is \textit{water}.
    Its paths are: \textbf{(1)} \begin{small}\texttt{teakettle} --[is a kind of]--> \texttt{x} <--[is at location]-- \texttt{water} \end{small}, \textbf{(2)} \begin{small}\texttt{put} --[is related to]--> \texttt{x} --[is used for]--> \texttt{water} \end{small}, and \textbf{(3)} \begin{small}\texttt{teakettle} --[is a kind of]--> \texttt{x} --[is used for]--> \texttt{water} \end{small}.
    
    \item
    For the negative KG example, the question is \textit{A poet may need to go where to achieve learning as an adult?}, the answer choice is \textit{book store}, and the target answer is \textit{university}.
    Its paths are: \textbf{(1)} \begin{small}\texttt{adult} <--[is related to]-- \texttt{x} --[is related to]--> \texttt{store} \end{small}, \textbf{(2)} \begin{small}\texttt{learning} <--[causes]-- \texttt{x} <--[is related to]-- \texttt{book} \end{small}, and \textbf{(3)} \begin{small}\texttt{learning} --[is related to]--> \texttt{x} --[is at location of]--> \texttt{book} \end{small}.
    
\end{itemize}

Second, we provide more examples of fine explanations.
Here, the question is \textit{What do you feel for a someone when you comfort friend?}, the answer choice is \textit{feeling bad}, and target answer is \textit{care}.
The positive path is: \begin{small}\texttt{comfort} <--[is the antonym of]-- \texttt{x} --[is the antonym of]--> \texttt{feel} \end{small}.
The negative path is: \begin{small}\texttt{comfort} --[is at location of]--> \texttt{x} --[is related to]--> \texttt{feeling} \end{small}.

The examples from Sec. \ref{sec:case_studies} are shown in Fig. \ref{fig:case_study_1}.
The examples introduced in this subsection (Sec. \ref{sec:app_case_qual}) are shown in Fig. \ref{fig:case_study_2}.
Again, in the coarse/fine explanations, we can roughly see that the positive KGs/paths tend to be useful for predicting the correct answer, and vice versa.
However, note that the model's judgment of KG/path usefulness may not necessarily align with human judgment \cite{raman2020learning}.

\subsection{User Studies: Quantitative Analysis of KG Saliency Explanations}
To better understand the role and limitations of KG saliency explanations, we quantitatively analyze KG saliency explanations in the context of two user studies.
In both user studies, the goal is to measure KG saliency explanations' plausibility, \textit{i.e.}, how closely the explanations align with human judgment.

Note that explanation plausibility is orthogonal to our paper's main claims, since we argue that KG saliency explanations can be used as additional supervision for improving performance, not that the explanations are plausible.
Nonetheless, these user studies may still provide some useful insights about KG saliency explanations.

\subsubsection{User Study 1: Coarse Saliency Explanations}
The first user study measures how well the coarse (graph-level) explanations align with human judgment of usefulness. Given a RoBERTa+PathGen model, we begin by uniformly sampling 25 high-saliency (positive) KGs and 25 low-saliency (negative) KGs from the CSQA training set. Recall that whether a KG is high-saliency or low-saliency was determined by coarse explanations (Sec. \ref{sec:explanations_coarse}) generated with respect to the given model.

\begin{wraptable}{R}{0.45\textwidth}
\vspace{-0.5cm}
\centering
\scalebox{0.80}{
    \begin{tabular}{lc}
        \toprule
        \textbf{Graph Type} & \textbf{Usefulness Score} \\
        \midrule
        High-Saliency Graph & 0.929 ± 0.734  \\
        Low-Saliency Graph   & 0.935 ± 0.764 \\
        \bottomrule
    \end{tabular}
}
\caption{\small \textbf{Human Evaluation of Coarse Saliency Explanations.} Human-annotated usefulness scores for high- (positive) and low- (negative) saliency graphs.}
\label{tab:graph_type_usefulness}
\vspace{-0.4cm}
\end{wraptable}
% % \vspace{-1mm}
% \label{tab:csqa_graph_type}
% % \vspace{-0.3cm}
% \end{table*}
% Graph Type	Usefulness Score
% High-Saliency Graph	0.929 ± 0.734
% Low-Saliency Graph	0.935 ± 0.764

Note that each KG corresponds to one answer choice of a question, so each question in CSQA has up to five corresponding KGs. To ensure that none of the KGs in our sample come from the same question, we ended up pruning two high-saliency and two low-saliency KGs, yielding a final sample of 23 high-saliency and 23 low-saliency KGs.

Since a KG can contain hundreds of paths, it is not feasible to ask humans to evaluate the entire KG's usefulness. Thus, as a very rough representation of the KG, we uniformly sampled three paths from the KG. Then, for each KG, we asked ten human annotators to score each of the three paths' usefulness for predicting the same answer choice predicted by the RoBERTa+PathGen model. To score the paths, all annotators were also given the question, correct answer, and model's predicted answer. The paths were scored on the following 0-2 scale:
\begin{itemize}
    \item \textbf{0} = definitely not useful (\textit{i.e.}, this path is either irrelevant or would cause someone to NOT select the model's predicted answer)
    \item \textbf{1} = possibly useful (\textit{i.e.}, this path provides some support for selecting the model's predicted answer)
    \item \textbf{2} = definitely useful (\textit{i.e.}, this path provides strong support for selecting the model's predicted answer)
\end{itemize}

Finally, each KG's score is computed as the mean of its three constituent path scores. Below, we show the mean and standard deviation scores for high-saliency and low-saliency graphs. We find that the two graph types have similar mean usefulness scores, while also having relatively large standard deviations. This suggests that coarse saliency explanations do not align strongly with human judgment. One key limitation of this study is that the three sampled paths may not be representative of the entire KG. In the future, we plan to redesign the user study to provide annotators a more comprehensive representation of the KG to evaluate.

\subsubsection{User Study 2: Fine Saliency Explanations}
The second user study measures how well the fine (path-level) explanations align with human judgment of usefulness. Given a RoBERTa+PathGen model trained on CSQA, we begin by uniformly sampling 25 correctly answered questions and 25 incorrectly answered questions from the CSQA training set. For each question, we take the model's predicted answer choice and the KG corresponding to the predicted answer choice, then select: \textbf{(1)} the path with the highest fine saliency score, \textbf{(2)} the path with median fine saliency score, and \textbf{(3)} the path with the lowest saliency score. To get finer-grained saliency signal in this study, we consider the raw fine saliency scores, instead of the binarized fine explanations actually used to regularize the model. Recall that a path's fine saliency score (Sec. \ref{sec:explanations_fine}) is calculated with respect to the given model.

Next, we asked ten human annotators to score each path's usefulness for predicting the same answer choice predicted by the RoBERTa+PathGen model. Like before, to score the paths, all annotators were also given the question, correct answer, and model's predicted answer. Again, the paths were scored on the following 0-2 scale:
\begin{itemize}
    \item \textbf{0} = definitely not useful (\textit{i.e.}, this path is either irrelevant or would cause someone to NOT select the model's predicted answer)
    \item \textbf{1} = possibly useful (\textit{i.e.}, this path provides some support for selecting the model's predicted answer)
    \item \textbf{2} = definitely useful (\textit{i.e.}, this path provides strong support for selecting the model's predicted answer)
\end{itemize}

Below, we show the mean scores for high-saliency, median-saliency, and low-saliency paths. We display these scores for paths from all predictions, correct predictions, and incorrect predictions. Overall, we find that the three path types have similar mean usefulness scores, although the mean score for median-saliency paths is somewhat higher than the other two path types'. Still, the standard deviations for all scores are relatively large, so this trend may not be meaningful. These results suggest that fine saliency explanations do not strongly align with human judgment. Additionally, we find that the path usefulness scores for correct predictions tend to be higher than those from incorrect predictions. This makes sense, since, intuitively, a model is more likely to predict the correct answer if it is using more useful knowledge as context.

\begin{table}[t]
\centering
\scalebox{0.73}{
    \begin{tabular}{lccc}
        \toprule
        \textbf{Path Type} & \textbf{Usefulness Score (All Preds)} & \textbf{Usefulness Score (Correct Preds)} & \textbf{Usefulness Score (Incorrect Preds)} \\
        \midrule
        High-Saliency Path & 1.091 ± 0.805 & 1.298 ± 0.782 & 0.884 ± 0.776  \\
        Med-Saliency Path & 1.222 ± 0.769 & 1.320 ± 0.729 & 1.124 ± 0.798  \\
        Low-Saliency Path & 1.060 ± 0.733 & 1.182 ± 0.730 & 0.938 ± 0.717 \\
        \bottomrule
    \end{tabular}
}
\vspace{0.2cm}
\caption{\small \textbf{Human Evaluation of Fine Saliency Explanations.} Human-annotated usefulness scores for high-, median-, and low-saliency paths. We display the usefulness scores for paths from all predictions, correct predictions, and incorrect predictions.}
\label{tab:path_type_usefulness}
\vspace{-0.5cm}
\end{table}
% Path Type	Usefulness Score (All Preds)	Usefulness Score (Correct Preds)	Usefulness Score (Incorrect Preds)
% High-Saliency Path	1.091 ± 0.805	1.298 ± 0.782	0.884 ± 0.776
% Med-Saliency Path	1.222 ± 0.769	1.320 ± 0.729	1.124 ± 0.798
% Low-Saliency Path	1.060 ± 0.733	1.182 ± 0.730	0.938 ± 0.717

\begin{wraptable}{R}{0.45\textwidth}
\vspace{-1.2cm}
\centering
\scalebox{0.80}{
    \begin{tabular}{lc}
        \toprule
        \textbf{User Study} & \textbf{Fleiss' Kappa} \\
        \midrule
        Coarse Explanations & 0.2089 \\
        Fine Explanations & 0.1296 \\
        \bottomrule
    \end{tabular}
}
\caption{\small \textbf{Inter-Annotator Agreement for Explanation User Studies.} Using Fleiss' kappa, we measure the inter-annotator agreement for the human evaluation of coarse and fine saliency explanations. In both settings, the inter-annotator agreement is relatively low.}
\label{tab:fleiss_kappa}
\vspace{-0.5cm}
\end{wraptable}
% User Study	Fleiss' Kappa
% Coarse Explanations	0.2089
% Fine Explanations	0.1296

\subsubsection{Inter-Annotator Agreement}
Here, we measure inter-annotator agreement for both user studies, using Fleiss' kappa. For the user study of coarse explanations, the kappa score is 0.2089, which is on the borderline of slight agreement and fair agreement. For the user study of fine explanations, the kappa score is 0.1296, which indicates slight agreement.

These low kappa scores show that even humans can hardly agree on whether the coarse/fine explanations are useful. Therefore, it may not always be beneficial to measure explanation quality in terms of alignment with human judgment. Moreover, this shows that weak alignment with human judgment does not necessarily imply poor explanation quality.

\subsubsection{Analysis}
In our user studies, we did not find strong evidence that coarse/fine saliency explanations align well with human judgment. 
However, we also found that human annotators had very low agreement about the usefulness of the explanations, which suggests that alignment with human judgment may not be the best measure of explanation quality.

In light of this, we emphasize that the user study results do not contradict our paper's conclusions, as our work does not claim that the generated saliency explanations are plausible. Rather, we merely claim that using KG-based saliency explanations as additional supervision to regularize KG-augmented models can yield higher performance. 

Our work appeals to the view that an explanation's quality should be measured by how well it distills knowledge for improving performance on some task \cite{pruthi2020evaluating}. Furthermore, the results of our user studies are actually in line with the conclusions from \cite{raman2020learning}, which found that KG-augmented models can effectively leverage KG information to improve performance, but in a manner that may not make sense to humans.

\subsection{Training Hyperparameters}
\label{sec:app_hparams}
Since we consider a very large number of models and settings in our experiments, we only describe the core hyperparameters here.
Let bsz denote batch size, let $\text{lr}_{\text{text}}$ denote text encoder learning rate, let $\text{lr}_{\text{graph}}$ denote graph encoder learning rate, and let $\text{lr}_{\text{task}}$ denote task predictor learning rate.
Across all models (both baselines and \textsc{SalKG}), we generally used the following hyperparameter sweeps: $\text{bsz} = [8, 16, 32, 64]$, $\text{lr}_{\text{text}} = [1\mathrm{e}{-5}, 2\mathrm{e}{-5}, 3\mathrm{e}{-5}, 5\mathrm{e}{-5}]$, $\text{lr}_{\text{graph}} = [1\mathrm{e}{-4}, 2\mathrm{e}{-4}, 3\mathrm{e}{-4}, 5\mathrm{e}{-4}]$, and $\text{lr}_{\text{task}} = [1\mathrm{e}{-4}, 2\mathrm{e}{-4}, 3\mathrm{e}{-4}, 5\mathrm{e}{-4}]$.
For CSQA and OBQA, we set the maximum number of epochs to 100. For CODAH, we set the maximum number of epochs to 30. For all three datasets, we used early stopping with a patience of 5 epochs.
For more details about hyperparameters, please refer to our \href{https://github.com/INK-USC/SalKG}{code repository}.

\subsection{Computational Costs and Resources}
\label{sec:app_compute}

Since the \textsc{SalKG} pipeline (as well as \textsc{Oracle}, \textsc{Random}, and \textsc{Heuristic}) involves training models across multiple stages, its computational costs are considerably greater than those from just training a No-KG or KG model individually.
Specifically, the pipeline involves: \textbf{(1)} training the No-KG and KG models; \textbf{(2)} creating coarse/fine explanations from the No-KG and KG models; \textbf{(3)} training the \textsc{SalKG}-Coarse model; \textbf{(4)} training the \textsc{SalKG}-Fine model; and \textbf{(5)} training the \textsc{SalKG}-Hybrid model.
In particular, using the Occl method to create fine explanations can be especially costly since it requires $n+1$ KG model forward passes per KG, where $n$ is the number of units in the given KG.
Also, if we tune the $T$ or $k$ thresholds comprehensively, then the total training time further increases.
For reference, each of our experiments was run on one NVIDIA Quadro RTX 8000 GPU.

Nonetheless, since we are the first to propose regularizing KG-augmented models with saliency explanations, it is expected that not all components of our method will already be fully optimized. That is, the goal of our work is simply to introduce a new paradigm for training KG-augmented models and demonstrate its potential by showing that it can yield improved performance. Certainly, there are various parts of the SalKG pipeline whose efficiency can be improved. For example, we could explore faster explanation generation via some KG-specific heuristic/approximation, training SalKG-Hybrid with coarse/fine explanations in a single step (instead of Steps 3-5 above), or generating explanations that can cover multiple instances at a time. Such potential improvements could be interesting directions for future work.

\subsection{Related Work (Extended)}
\label{sec:app_related_work}

% \subsubsection{Creating Model Explanations}
\paragraph{Text-Based Explanations}
Many works have been proposed for explaining the predictions of language models, especially PLMs.
Although some of these works focus on abstractive (free-text) explanations \cite{rajani2019explain, strout2019human, zhao2020lirex}, most aim to provide extractive explanations which highlight salient tokens in the model's text input.
Such extractive explanations typically use either gradient-based \cite{sundararajan2017axiomatic, li2015visualizing, denil2014extraction}, attention-based \cite{mohankumar2020towards, tutek2020staying, ghaeini2018interpreting, lee2017interactive}, and occlusion-based \cite{deyoung2019eraser, poerner2018evaluating, kadar2017representation, li2016understanding} feature attribution methods.
How feature attribution methods should be chosen remains an open question and the subject of much recent debate \cite{bastings2020elephant, wiegreffe2019attention, serrano2019attention, jain2019attention}.
% In this work, we propose \textsc{SalKG} for explaining KG-augmented models, which take both text and graph inputs. 
While \textsc{SalKG} also uses feature attribution methods (e.g., G$\times$I) to create extractive explanations, our study is limited to explanations regarding KG-augmented models' graph inputs.

\paragraph{Graph-Based Explanations}
There are also methods proposing extractive explanations for graph encoders, especially GNNs.
Such explanations are designed to point out components in the graph input that contribute most to the model's prediction. 
Some GNNs use attention for pooling, which naturally highlights nodes with higher attention weights \cite{lee2019self, lee2018graph}.
More sophisticated approaches use post-hoc optimization to identify salient nodes \cite{huang2020graphlime, ying2019gnnexplainer} or subgraphs \cite{ying2019gnnexplainer}.

% Many GNNs use attention for pooling and/or message passing, which can highlight nodes \cite{lee2019self, lee2018graph} or edges \cite{velivckovic2017graph, zhang2018gaan}. However, message passing attention \cite{velivckovic2017graph, zhang2018gaan} is less helpful for explainability, since layer-wise attention cannot globally explain the model's prediction. Saliency is also used for post-hoc GNN explanations, either with respect to node features \cite{huang2020graphlime, ying2019gnnexplainer} or subgraphs \cite{ying2019gnnexplainer}.

Unlike individual PLMs and graph encoders, KG-augmented models take both text and graph inputs.
The KG-augmented model's graph encoder usually computes graph embeddings via attention pooling of nodes/paths, and the attention weights can be used to explain which nodes/paths in the input KG are salient \cite{lin2019kagnet, feng2020scalable, liu2020commonsense, wang2020connecting, yan2020learning}. These KG explanations can be interpreted as identifying knowledge in the KG that is complementary to the knowledge encoded in the PLM. However, there is little work on how such KG explanations should be used. \textsc{SalKG} considers graph-based extractive explanations of KG-augmented models, but focuses more on how explanations are used rather than created.
% There exist relatively few efforts in explaining KG-augmented models
% KG-augmented model explanations generally focus on identifying KG's influence with respect to the language model. There exist relatively few efforts in this direction, via attention 
% , which may not be faithful. However, \citet{liu2020commonsense} also considers abstractive explanations, which require explanation labels and are even less likely to be faithful.

\paragraph{Learning From Model Explanations}

% Generally, model explanations are created with the goal of aiding humans in decision-making or model design \cite{lipton2018mythos, doshi2017towards}.
% Given this human-in-the-loop setting, it may not be straightforward to define concrete procedures for using these explanations.
% Nonetheless, despite the extensive literature on creating model explanations, there is relatively little concrete procedures for using these explanations. 
To improve the model's learning, explanations can be used in a diverse range of ways, including as extra supervision or regularization \cite{pruthi2020evaluating, hase2020leakage, narang2020wt5, andreas2017learning}, pruned inputs \cite{jain2020learning, bastings2019interpretable, lei2016rationalizing}, additional inputs \cite{hase2021can, co2018guiding}, and intermediate variables \cite{wiegreffe2020measuring, zhou2020towards, rajani2019explain}.
The most similar work to ours is \cite{pruthi2020evaluating}, which proposed training a student model to mimic a teacher model's predictions by regularizing the student model's attention via text explanations created from the teacher model.
However, \cite{pruthi2020evaluating} aims to evaluate explanations, while our goal is to improve performance via explanations. 
Still, methods for learning from explanations have largely focused on domains like text and images, as opposed to graphs. To the best of our knowledge, \textsc{SalKG} is the first work to train KG-augmented models using KG explanations as supervision.

\subsection{Societal Impact}
Our proposed \textsc{SalKG} approach for learning from KG explanations can be applied to any KG-augmented model and can be adapted from any off-the-shelf saliency method.
This enables KG-augmented models to improve generalization ability and learn more efficiently from data, thus yielding better performance while requiring less labeled data.
However, in the present version of \textsc{SalKG}, this generalization ability and data efficiency comes with increased computational costs, as described in Sec. \ref{sec:app_compute}.
In the future, we plan to explore methods for improving generalization and data efficiency while minimizing computational costs.

\end{document}